\documentclass[preprint,12pt]{elsarticle}

\usepackage[T1]{fontenc}

\usepackage{amssymb}
\usepackage{amsmath}
\usepackage{graphicx}
\usepackage{algorithm}
\usepackage{algpseudocode}
\usepackage{diagbox}
\usepackage{subcaption}
\usepackage{ragged2e}
\usepackage[export]{adjustbox}
\usepackage{multirow}
\usepackage{booktabs}
\usepackage{tabularx}
\usepackage{xcolor}
\usepackage{array}
\usepackage{hyperref}
\usepackage{rotating}

\newcolumntype{P}[1]{>{\centering\arraybackslash}m{#1}}

\newcommand{\revise}[1]{\textcolor{black}{#1}}

\newcommand{\revisenew}[1]{\textcolor{black}{#1}}

\newcommand{\revisefinal}[1]{\textcolor{black}{#1}}

\DeclareMathOperator{\mean}{mean}

\journal{Robotics and Autonomous Systems}

\begin{document}

\begin{frontmatter}



\title{Vision-driven River Following of UAV via Safe Reinforcement Learning using Semantic Dynamics Model}


\author{Zihan Wang and Nina Mahmoudian} %% Author name

\affiliation{organization={School of Mechanical Engineering, Purdue University},%Department and Organization
            addressline={1500 Kepner Dr.}, 
            city={West Lafayette},
            postcode={47906}, 
            state={IN},
            country={USA}}

%% Abstract
\begin{abstract}
Vision-driven autonomous river following by Unmanned Aerial Vehicles is critical for applications such as rescue, surveillance, and environmental monitoring, particularly in dense riverine environments where GPS signals are unreliable.
\revisefinal{These safety-critical navigation tasks must satisfy hard safety constraints while optimizing performance.
Moreover, the reward in river following is inherently history-dependent (non-Markovian) by which river segment has already been visited, making it challenging for standard safe Reinforcement Learning (SafeRL).
}% We \revisefinal{address these gaps by framing} river following as a coverage control problem in which the reward function is submodular—yielding diminishing returns as more unique river segments are visited—thereby framing the task as a Submodular Markov Decision Process.
\revisefinal{To address these gaps, we cast river following as a coverage control problem with a submodular reward that exhibits diminishing returns as more river segments are visited, framing the task as a Submodular Markov Decision Process.}
Building on the SafeRL paradigm and the First Order Constrained Optimization in Policy Space (FOCOPS) algorithm, we propose three contributions. 

First, we introduce the Marginal Gain Advantage Estimation (MGAE), which refines the reward advantage function using a sliding-window baseline calculated from historical episodic returns, aligning the advantage estimate with non-Markovian dynamics. 
Second, we develop a Semantic Dynamics Model (SDM) based on patchified water semantic masks, offering more interpretable and data-efficient short-term prediction of future observations compared to latent vision dynamics models. 
Third, we present the Constrained Actor Dynamics Estimator (CADE) architecture, which integrates the actor, cost estimator, and SDM for cost advantage estimation to form a model-based SafeRL framework capable of solving partially observable Constrained Submodular Markov Decision Processes.

The simulation results demonstrate that MGAE achieves faster convergence and superior performance compared to critic-based methods like Generalized Advantage Estimation.
SDM provides more accurate short-term state predictions, enabling the cost estimator to better predict potential violations. 
Overall, CADE effectively integrates safety regulation into model-based RL, with the Lagrangian approach providing a "soft" balance between reward and safety during training, while the safety layer enhances inference by imposing a "hard" action overlay.
Our code is publicly available on Github (https://github.com/EdisonPricehan/omnisafe-cade/tree/cacd). 
\end{abstract}


\begin{highlights}
\item Propose a marginal gain advantage method for non-Markovian rewards in coverage tasks.
\item Develop a homography-based semantic dynamics model for vision-driven navigation.
\item Introduce a model-based safe RL framework using learned reward and cost estimators.
\item Marginal gain advantage enables faster learning and surpasses critic-based methods.
\item Compare Lagrangian training with the cost-planning safety layer for safe policy execution.

\end{highlights}

%% Keywords

\end{frontmatter}

\section{INTRODUCTION}\label{sec:intro}
Autonomous navigation in riverine environments is a crucial task for Unmanned Aerial Vehicles (UAVs) in applications such as environmental monitoring, surveillance, and rescue missions. 
Although previous work has focused on vision-driven navigation using segmentation masks for water detection \cite{taufik2015multi}, these approaches often rely on manual tuning and simplistic environmental models.
\revisefinal{Manually tuned controller can work in controlled settings, but as environmental variability (trees, bridges, and river geometries) increases it becomes brittle to distribution shift and costly, subjective, and time-consuming to retune for each site \cite{lambert2022rosebud,wang2023aerial}.}
% making them unsuitable for complex, real-world scenarios involving obstacles like trees, bridges, and varying river geometries \cite{lambert2022rosebud,wang2023aerial}. 
\revisefinal{In contrast, a common alternative is GPS-based waypoint navigation, which tracks preplanned river centerlines based on prior maps.
However, seasonal changes in channel shape and canopy-induced GPS outages degrade these plans, undermining the reliability of safe and precise river following.}
% Moreover, conventional GPS-based waypoint navigation suffers from inaccuracies due to seasonal variations in river shape and can be further compromised in dense jungle riverine environments where GPS signals are unreliable or unavailable, making these methods unsuitable for safe and precise river following missions. 
Thus, vision-based navigation becomes essential in such environments \cite{wang2024synergistic}.

% Describe the problem (CSMDP) and its key features/challenges
Vision-driven river following, as a specific form of coverage control problem, poses significant challenges to Reinforcement Learning (RL) due to its \textit{partial observability}, \textit{submodular reward structure}, and \textit{non-Markovian nature}. 
The partial observability arises from the agent's reliance on vision-based observations, which provide only a limited first-person view of the environment, making it difficult to perceive the entire river layout \cite{wang2024synergistic}.
Unlike standard Markov Decision Processes (MDPs, \cite{sutton2018reinforcement}), where the reward depends solely on the current state and action, river-following rewards are non-Markovian, as they depend on the agent's historical trajectory and are given only when new unvisited segments of the river are explored \cite{wang2024synergistic}.
\revisefinal{
History-dependent rewards also arise in information gathering problems, where rewards encode reductions in uncertainty, making the return depend on the observation history rather than only the current state \cite{lauri2019information}.
}
% Feature: submodularity
%Additionally, this reward function is submodular \cite{prajapat2023submodular}, as visiting new segments provides diminishing returns over time. This submodularity requires a more sophisticated advantage estimation method that can leverage the unique structure of the reward.
\revise{Furthermore, this reward function is submodular, meaning that the marginal gain from visiting river segments diminishes as more segments are explored, akin to a diminishing returns property in optimization. 
This submodular structure, combined with the fact that the rewards depend on the agent’s past trajectory rather than just the current state, makes the task a Submodular Markov Decision Process (SMDP, \cite{prajapat2023submodular}). 
As a result, a more sophisticated advantage estimation method is required to effectively leverage the submodular nature of the reward and the dependence on the trajectory.}
Furthermore, safety is a critical concern in such environments, necessitating the use of Constrained Markov Decision Processes (CMDPs, \cite{altman2021constrained}) that introduce a cost function $C$ to penalize unsafe actions, extending MDPs with the necessary safety constraints. 
% Conclude to CSMDP
These challenges require more sophisticated reward estimation and policy optimization methods that can effectively capture temporal dependencies, leverage the submodular reward structure, and address safety requirements, leading to the formulation of a Constrained Submodular Markov Decision Process (CSMDP).

% Propose MGAE
To address the submodular and non-Markovian nature of the reward function in river following tasks, we propose a Marginal Gain Advantage Estimation (MGAE) method. 
Unlike traditional advantage estimators such as Generalized Advantage Estimation (GAE, \cite{schulman2015high}), which look forward along the trajectory and rely on a state-dependent value function as the baseline, MGAE combines both forward and backward perspectives. 
It looks forward to accumulate actual rewards while looking backward into the agent's trajectory to estimate cumulative marginal gains. 
This approach captures the historical dependencies essential in environments where rewards are given only for visiting new, unvisited river segments, resulting in a less biased estimation of the agent's true progress. 
By accounting for the submodular reward structure and using the average episodic return as a baseline, MGAE is particularly well-suited for tasks where the reward depends on the agent's exploration history, rather than solely on the current state or observation.

% Propose SDM
In addition to improving the reward advantage estimation in SMDP, we introduce a Semantic Dynamics Model (SDM) to predict future states based on the agent's visual input.
In the context of visual semantic navigation \cite{yang2018visual, mayo2021visual, santos2022deep, lyu2023double}, some model-based safe reinforcement learning approaches leverage semantic information to guide decision-making without explicit knowledge of spatial context. SDM complements this by exploiting the dynamics of visual observations in both geometric and semantic manners, enhancing the agent's understanding of the environment.
SDM explicitly incorporates homography transformations between observations in first-person view, preserving both geometric and semantic information. Unlike state-of-the-art latent dynamics models \cite{hafner2019dream, hafner2023mastering, hafner2019learning, hafner2020mastering, huang2023safe}, which abstract raw pixel data into latent vectors, losing essential spatial and semantic relationships, our approach offers greater interpretability by modeling state transitions based on the actual geometric structure of the riverine environment. 
This makes our model particularly suitable for real-world applications, such as outdoor navigation, where spatial reasoning and explainability are critical.
SDM helps the agent anticipate potential hazards to safely navigate through complex riverine environments. 
By integrating this model with our advantage estimation method, the agent can make more informed decisions that account for both future rewards and potential safety violations, leading to safer exploration and better task performance.

Our architecture, Constrained Actor Dynamics Estimator (CADE), brings together MGAE and SDM within the framework of safe reinforcement learning. 
The CADE architecture is designed to optimize a policy that balances reward maximization with safety constraints, ensuring that the UAV navigates effectively while avoiding collisions or other hazardous behaviors. 
Inspired by recent advances in constrained policy optimization methods \cite{achiam2017constrained,zhang2020first}, our approach applies the Lagrangian relaxation technique to handle safety constraints during training, making it suitable for real-world deployment in safety-critical environments.
This work builds upon and extends our prior research in vision-based navigation \cite{wang2024synergistic} using safe reinforcement learning \cite{wang2024vision}, addressing the unique challenges of autonomous river following in complex environments with only visual inputs.

% Contributions
In summary, the main contributions of this work are as follows:
\begin{itemize}
    \item We propose the Marginal Gain Advantage Estimation (MGAE) method to address non-Markovian submodular reward functions in coverage control problems like track/river following tasks.
    \item We develop the Semantic Dynamics Model (SDM) based on homography transformation between adjacent semantic observations to predict future observations and help improve safer decision-making for vision-driven navigation.
    \item We introduce the Constrained Actor Dynamics Estimator (CADE) architecture, integrating MGAE for reward advantage estimation, SDM and cost estimator for cost advantage estimation, forming a model-based SafeRL framework to solve partially observable CSMDPs.
    %vision-driven UAV navigation in constrained environments, achieving both high task performance and low constraint violations.
\end{itemize}

% Outline (Optional)
This paper is organized as follows:
Section \ref{sec:related} introduces related work in terms of river following and vision-driven model-based safe reinforcement learning that solves this coverage control problem. 
Section \ref{sec:methodologies} explains our contributions in marginal gain advantage estimation for reward, the semantic dynamics model, the cost advantage of the cost estimator, and the combined architecture of the constrained actor dynamics estimator architecture.
Section \ref{sec:experiments} shows our experiment design, and Section \ref{sec:results} shows the comparison results and analyses.
Section \ref{sec:discussion} discusses several design choices and their implications for future work.
Finally, Section \ref{sec:conclusions} concludes our paper with key contributions and conclusions.

\section{RELATED WORK}\label{sec:related}
This section introduces the related work from the viewpoints of several perspectives for the vision-driven safe river following task: existing approaches and environments for autonomous \revisenew{semantic perception driven} river following (Section \ref{sec:related-river}), reward advantage estimation methods in usual Markov Decision Processes (Section \ref{sec:related-advantage}), vision representation learning and vision-based dynamics model construction (Section \ref{sec:related-dynamics}), and methods using the dynamics model in safe reinforcement learning (Section \ref{sec:related-mbsrl}). 

\subsection{River Navigation}\label{sec:related-river}
Autonomous UAV navigation along curved river segments has previously been implemented using threshold-based HSV (hue, saturation, and value) water segmentation and a proportional heading controller \cite{taufik2015multi}. 
However, the semantic mask extraction in this approach requires a precise manual tuning, which can limit its adaptability to varying river types and lighting conditions. 
Furthermore, the proportional controller struggles with navigation in rivers with tributaries, while a fixed-altitude flight path can pose challenges when encountering bridges.
\revisenew{
Recent advances in image segmentation, particularly deep convolutional neural networks, have enabled robust and generalizable semantic segmentation for robotics and UAV navigation tasks, far surpassing classic HSV thresholding methods that require extensive manual tuning and are sensitive to environmental variations.
Modern approaches leverage semantic segmentation masks as input to control policies, facilitating robust navigation and obstacle avoidance in complex environments \cite{teso2020semantic, miyamoto2019vision, hong2021semantically, hong2018virtual, osinski2020simulation, guan2022ga}.
For example, \cite{guan2022ga} shows that a segmentation-based preprocessing step (GA-Nav) classifies terrain navigability in unstructured outdoor environments, leading to a $10\%$ increase in navigation success and a $4.6 \%$ to $13.9\%$ decrease in trajectory roughness compared to RGB-only policies on a ground vehicle.
In general, semantic masks serve as domain-invariant, high-level inputs that boost sim-to-real transferability and reduce sample complexity compared to raw RGB or autoencoder embeddings. 
Our approach builds on this trend by employing patchified semantic water masks as the observation space (Section \ref{sec:methodologies-sdm}), which enhances sample efficiency and the robustness to visual noise.
}
%To address these limitations, modern image embedding and semantic segmentation methods utilizing deep convolutional neural networks offer reduced manual tuning and improved generalization. 
%Leveraging these encoded visual data, reinforcement learning methods can be applied in photorealistic simulation environments to train autonomous navigation policies, allowing for obstacle avoidance and robustness testing before real-world deployment.

The researchers used imitation learning to develop a river navigation policy based on vision input in an environment based on Unreal Engine \cite{liang2021vision, wei2022vision}. 
They showed that separate training of the image encoder (variational autoencoder in their experiments) and the control policy is superior to an end-to-end neural network controller in terms of navigation distance and intervention rate.
However, their method has several limitations that impact its applicability. 
The intervention-based DAgger algorithm \cite{ross2011reduction} depends on manual guidance and reset criteria during training to ensure safety, which is \revisenew{human involved for safety feedback}. 
In addition, it restricts the agent’s actions to a 2D plane, lacking the full 3D maneuverability required for UAV tasks. 
The lack of public access to the environment further limits reproducibility and broader research collaboration. 
These constraints make the method highly dependent on human involvement during training, for both task completion and obstacle avoidance. 

In comparison, our Unity-based Safe Riverine Environment (SRE) \cite{wang2024vision} provides immediate cost feedback, enabling the reinforcement learning agent to learn a safe river-following policy autonomously. 
The UAV agent in SRE has 4 degrees of freedom (translation and heading) in 3D space, along with obstacle detection for bridges, making it more realistic for complex riverine environments. 
SRE includes three difficulty levels and is publicly available as an OpenAI Gym environment, built using the Unity ML-Agents Toolkit \cite{juliani2020}. 
During policy testing, a semantic segmentation network trained on the Aerial Fluvial Image Dataset (AFID) \cite{PURR4105, wang2023aerial} generates the water mask. 
To mitigate the effects of simulation-to-reality differences on policy training, we replace RGB encoding with a patchified, coarser representation of the water mask (Figure \ref{fig:patchified}).

\subsection{Advantage Estimation}\label{sec:related-advantage}
Advantage estimation is a critical component in reinforcement learning, as it enables the agent to evaluate the benefit of actions relative to its expected returns.
\revisenew{
A fundamental approach involves Temporal Difference (TD, \cite{williams1991function}) Advantage Estimation. This method directly estimates the advantage at each time step by using a simple TD error between the current value estimate and the immediate reward plus the value at the next state. Although computationally simple, it can suffer from high variance, particularly in environments with sparse or delayed rewards.
}

\revisenew{
One of the most widely adopted methods is Generalized Advantage Estimation (GAE, \cite{schulman2015high}) in the Proximal Policy Optimization (PPO) framework. GAE calculates the advantage function by blending TD errors in multiple time steps with a smoothing parameter $\lambda$, which provides a trade-off between bias and variance. 
Using this temporal smoothing, GAE is particularly effective in scenarios where accurate, low-variance advantage estimates are essential for stable policy learning.
Since GAE downgrades to TD when $\lambda = 0$, we set $\lambda = 0.95$ for all GAE-related experiments.
}

For partially observable or non-Markovian reward environments, V-trace is an advantage estimation method used within the IMPALA architecture \cite{espeholt2018impala}. V-trace corrects off-policy trajectories by scaling the advantage estimates according to importance sampling weights, reducing variance introduced by sampling errors, and ensuring stable advantage estimation in large-scale distributed setups.

Although these methods have shown success across various domains, they rely on state-wise advantage estimation, focusing primarily on short-term TD signals. However, this can be limiting in environments with non-Markovian and submodular reward structures, where the immediate reward signal is insufficient to accurately reflect long-term progress. Such settings, including river or track-following tasks, require the agent to maintain awareness of visited regions to avoid revisiting and to maximize task-specific rewards. Traditional state-wise approaches often fail to capture this trajectory-level context, as they overlook the cumulative effect of prior actions on the current reward.

To address these limitations, we propose the Marginal Gain Advantage Estimation (MGAE), which incorporates the cumulative marginal gain over trajectories to reflect the agent’s progress in visiting new states. Unlike GAE and other state-wise methods, MGAE tracks the historical dependencies of rewards by estimating advantage based on the accumulated marginal gains, making it better suited for non-Markovian tasks with history-dependent rewards. This approach provides a less biased and more informative signal for policy optimization in complex navigation tasks, where maintaining trajectory-level awareness is critical for effective and efficient exploration.

\subsection{Vision Dynamics Model}\label{sec:related-dynamics}
% Latent dynamics
In recent years, model-based reinforcement learning approaches, such as PlaNet \cite{hafner2019learning} and the Dreamer series \cite{hafner2019dream,hafner2020mastering,hafner2023mastering}, have demonstrated the potential of learning latent dynamics models directly from high-dimensional visual inputs. 
PlaNet introduced a recurrent state-space model (RSSM) for planning in latent space. DreamerV1 \cite{hafner2019dream} extended this by using a variational autoencoder (VAE) \cite{kingma2013auto} to encode raw RGB observations into a compact latent space, enabling efficient simulation and decision-making. DreamerV2 \cite{hafner2020mastering} improved DreamerV1 with better stability and scalability, supporting longer training horizons and continuous control tasks. DreamerV3 \cite{hafner2023mastering} further refined the framework by improving generalization and planning efficiency in complex environments, optimizing both reward prediction and policy learning within the latent dynamics model.

% Drawbacks of LDM
\revisefinal{
Different levels of abstraction in vision dynamics models depend on the task and environment \cite{ai2025review}.
}
While PlaNet and Dreamer models have advanced the learning of compact latent dynamics, they lack interpretability, overlook explicit geometrical information, and do not emphasize task-relevant semantics, potentially leading to \revisefinal{efficiency, object permanence and temporal consistency issues in riverine} environments. 
Additionally, if an agent needs to learn a Markovian cost function based on vision input, a latent vector input may obscure dependencies and make learning less intuitive. 
In contrast, the proposed semantic dynamics model (SDM) integrates interpretable geometrically structured representations and semantically meaningful features, enhancing the agent’s capacity for efficient learning, spatial coherence and decision-making in complex task-oriented environments.

% DVF
Deep Visual Foresight \cite{finn2017deep} provides a flexible framework for robot motion planning through pixel-based self-supervised prediction, enabling online adaptation without semantic input. 
While highly adaptable, this approach can face challenges with substantial lighting changes or unfamiliar terrain and incurs significant computational cost due to raw pixel processing. 
The SDM, in contrast, reduces visual dimensionality through a patchified semantic mask, yielding a compact, interpretable representation with essential environmental context preserved. 
Although SDM uses 2D homography rather than full 3D geometry, it captures spatial transformations efficiently, offering a structured and computationally efficient model for geometrically constrained ego-centric vision-driven tasks.

\subsection{Model-based Safe Reinforcement Learning}\label{sec:related-mbsrl}
% General background intro
In safe reinforcement learning (SafeRL), the objective is to optimize a policy while keeping the cumulative cost below a certain threshold \cite{altman2021constrained}.
Decoupling task-specific rewards from environment-driven costs ensures that safety remains an independent priority rather than being tied to task performance.
%A general solution to SafeRL is the Lagrangian method, introducing a lagrange multiplier that penalizes the agent whenever it violates the cost constraint.
Many model-free RL algorithms \cite{achiam2017constrained,zhang2020first,stooke2020responsive,ma2022joint,dai2023augmented,hogewind2022safe} try to solve the safe optimal policy in CMDP using the Lagrangian method.
However, in model-free Lagrangian SafeRL, suboptimal performance under low-cost thresholds arises because the critic’s estimation inaccuracies lead to frequent constraint violations, as the agent lacks precise cost information for its actions \cite{huang2023safe}.
In addition, safe model-free algorithms still require extensive interactions with the environment to converge, while model-based methods can significantly improve sample efficiency by also learning from imagined trajectories \cite{as2022constrained}.
\revisenew{
Moreover, as pointed out in \cite{brunke2022safe, roza2023towards}, system dynamics knowledge, either learned from data or acquired from prior modeling, is necessary for stronger safety guarantees. 
}

% Existing Safe RL algorithms that use model for planning
\revisenew{
Building on the better data efficiency and constraint satisfaction of model-based approaches, several safety regulation paradigms have incorporated dynamics models to enhance SafeRL.
First, Control Barrier Functions (CBF) enforce one-step safety by solving a quadratic program that constrains the Lie derivative of a smooth barrier function, thus guaranteeing forward invariance of a safe set \cite{harms2024neural, so2024train, zhang2023exact, liu2023safe, ma2021model, marley2021maneuvering}. 
However, synthesizing this function requires domain knowledge and does not scale to high-dimensional visual observations \cite{guerrier2024learning}. 
Shielding methods extended this idea by precomputing backup policies or reachable safe sets that logically block unsafe actions at runtime \cite{yang2023safe, bloem2015shield, alshiekh2018safe, mazzi2023risk}, yet they depend on formal specifications and generally do not feed back into policy learning.
Third, model-based Lagrangian policy optimization techniques \cite{as2022constrained, huang2023safe, stooke2020responsive, wen2018constrained} embed cost constraints directly into the policy gradient objective via dual variables, allowing the agent to learn trade-offs between cumulative reward and cost.
}
%For example, the LAMBDA \cite{as2022constrained} algorithm combines a Bayesian latent dynamics model with an augmented Lagrangian framework to improve constraint satisfaction and policy optimization efficiency. Similarly, SafeDreamer \cite{huang2023safe} incorporates a latent dynamics model for foresight into safety-reward planning, relying on a PID-Lagrangian approach \cite{stooke2020responsive} to balance trade-offs between safety and reward during trajectory planning using the Constrained Cross-Entropy Method (CCEM, \cite{wen2018constrained}).
\revisenew{
However, Lagrangian-based methods ensure constraint satisfaction in expectation during training, but do not enforce per‐step safety once the policy is deployed.
}
%However, Lagrangian-based methods, while effective in balancing objectives, often fall short of ensuring strict safety guarantees, particularly in high-stakes environments where constraints must be rigorously enforced and the dynamics of the environment are not known a priori.
\revisenew{
Fourth, proactive model predictive shielding (MPS) approaches, including Statistical MPS \cite{bastani2021safe} and Dynamic MPS \cite{banerjee2024dynamic, goodall2023approximate}, either uses probabilistic check to ensure safety with high probability, or rank candidate actions using finite-horizon rollouts through a known or learned dynamics model to dynamically synthesize shielding action at each step, co-training the planner with the policy to improve both safety and performance.
}

\revisenew{
In comparison to classic CBF or shielding methods that project or reject unsafe actions as post-hoc safety filters without altering the underlying policy, we focus on using a model predictive safety shield that regulates policy during training like Lagragian-based methods, but also can be activated during inference like MPS approaches.  
}
\revisenew{While existing Lagrangian-based or MPS methods} incorporate dynamics models for safety-aware planning, they primarily rely on cost critics to estimate long-term safety violations.
\revisenew{
This is not the case in our UAV river/track waypoint navigation setting, where the vehicle is treated as point mass that can simply reverse course upon sensing danger.
Thus, the genuine safety failures only occur the moment of, for example collision, not from drifting too far into the future \cite{konighofer2020safe, konighofer2023online}. 
%In contrast to reward critics, which estimate long-term state values to guide task completion over many steps, safety violations often occur within just a few steps. 
Consequently, immediate Markovian costs, captured by a cost estimator, are preferable to provide timely danger signals for policy updates, without the complexity or potential inaccuracy of long‐horizon cost value functions, for agents with ego-centric observations.
}
However, leveraging predicted immediate costs to form a cost advantage for SafeRL, or incorporating this signal into a safety layer during training, remains underexplored. 

\revise{
To address this gap, our CADE framework integrates a cost estimator with a semantic dynamics model, enabling two distinct approaches to enforce safety.
The Lagrangian-based method, used in the nominal CADE framework, leverages the short-horizon cost advantage for policy updates, ensuring the agent gradually responds to potential hazards without relying on long-term bootstrapping. 
As a separate method, the cost-planning safety \revisenew{filter} can be enabled during training or inference, performing real-time action overlays by predicting and assessing short-horizon safety violations to enforce safe behavior when necessary.
}

%“SafeDreamer” \cite{huang2023safe} proposes a framework for safe reinforcement learning by integrating Dreamer’s dynamics model with a Lagrangian-based safety constraint, achieving near-zero safety violations in both low-dimensional and vision-based tasks, particularly outperforming previous safe RL approaches in the Safety-Gymnasium benchmark \cite{ji2023safety}.

\section{METHODOLOGIES}\label{sec:methodologies}
In this section, we introduce the Constrained Submodular Markov Decision Process (CSMDP), which extends the traditional MDP framework by incorporating both non-Markovian reward and Markovian cost (Section \ref{sec:methodologies-csmdp}). 
This dual feedback mechanism allows for more nuanced control over the learning process, balancing task performance with safety constraints \cite{hogewind2022safe}.
To learn a performant policy in such CSMDPs, we subsequently detail the proposed Marginal Gain Advantage Estimation (MGAE) method for non-Markovian reward in Section \ref{sec:methodologies-mgae}. 
Additionally, we explain the Semantic Dynamics Model (SDM), which plays a crucial role in enabling vision-driven reinforcement learning by capturing the temporal and spatial relationships between consecutive observations (Section \ref{sec:methodologies-sdm}). 
Finally, we detail how the SDM is integrated with a cost advantage estimator to form the Constrained Actor Dynamics Estimator (CADE, Section \ref{sec:methodologies-cade}) architecture, ensuring that safety considerations are efficiently embedded into the learning process.

\subsection{Constrained Submodular Markov Decision Process}\label{sec:methodologies-csmdp}
Safe RL tries to find an optimal policy that maximizes episodic rewards while satisfying safety constraints in a Constrained Markov Decision Process (CMDP) \cite{altman2021constrained}. 
CMDP is a tuple formed by $(S, A, R, C, P, \mu)$, where $S$ is the set of states, $A$ is the set of actions, $R: S \times A \rightarrow \mathbb{R}_{\geq 0}$ is the reward function, $C: S \times A \rightarrow \mathbb{R}_{\geq 0}$ is the cost function, $P: S \times A \times S \rightarrow [0, 1]$ is the state transition probability function, and $\mu: S \rightarrow [0, 1]$ is the initial state distribution. 
Compared to Markov Decision Process (MDP) for regular RL, CMDP for safe RL incorporates cost set function $C$, penalizing undesirable states or actions to enforce safety constraints by keeping cumulative costs within a threshold.

Formally, the optimal feasible policy is a policy $\pi : S \rightarrow A$ within a feasible policy space $\pi \in \Pi_{C}$,
\begin{equation}\label{eqn:cmdp}
    \pi^{\ast} = \arg \max _{\pi \in \Pi_{C}} J^R(\pi).
\end{equation}
A policy's optimality is determined by its performance measure \revisenew{$J^R(\pi) \in \mathbb{R}_{\geq 0}$}, which is usually the expected discounted cumulative rewards over the infinite horizon: $J^R(\pi) \doteq \revisenew{E_{\pi}} [\sum_{t=0}^{\infty} \gamma^t R(s_t, a_t)]$. 
A policy's feasibility is determined by its safety measure \revisenew{$J^C(\pi) \in \mathbb{R}_{\geq 0}$}, which is similar as the performance measure in definition: $J^C(\pi) \doteq \revisenew{E_{\pi}} [\sum_{t=0}^{\infty} \gamma^t C(s_t, a_t)]$. Thus $\Pi_{C} \doteq \{\pi \mid J^C(\pi) \leq d \}$ is the feasible policy set, where $d$ is a cost budget hyperparameter.
In this work we only focus on single cost constraint, but CMDP does not have restrictions on the number of cost constraints.
In CMDP, the reward and cost advantage functions are both defined as the difference between the state action value function $Q_{\pi} (s, a) \doteq E_{\tau \sim \pi} [\sum_{t=0}^{\infty} \gamma^t R(s_t, a_t) \mid s_0 = s, a_0 = a]$ and the state value function $V_{\pi} (s) \doteq E_{\tau \sim \pi} [\sum_{t=0}^{\infty} \gamma^t R(s_t, a_t) \mid s_0 = s]$, i.e., $A_{\pi} (s, a) = Q_{\pi} (s, a) - V_{\pi} (s)$\revisefinal{, where $\tau = (s_0, a_0, s_1, a_1, ...)$ is the trajectory of state-action pairs induced by policy $\pi$}. 
However, in environments where the reward function is submodular and non-Markovian, the traditional state-wise advantage estimation may fail to capture the trajectory-dependent nature of rewards, leading to suboptimal policy updates.
Therefore, a trajectory-wise advantage estimation that considers the cumulative submodular rewards is necessary for more effective learning in these complex environments.

A Constrained Submodular Markov Decision Process (CSMDP) extends the CMDP by recognizing the submodularity of the reward function, meaning the reward in CSMDP is non-additive and non-Markovian (history-dependent), and its value decreases in light of similar states visited previously, \cite{prajapat2023submodular}. 
Formally, the set function $R$ in CSMDP can be denoted as $R_{sub}$, which is submodular. \revisenew{For two subsets $S_1, S_2$ of set $S$ that meet $\forall S_1 \subseteq S_2 \subseteq S, s \in S \setminus S_2$}, we have
\begin{equation}\label{eqn:submodularity}
    \revisefinal{R_{sub}(S_1 \cup \{s\}) - R_{sub}(S_1) \geq R_{sub}(S_2 \cup \{s\}) - R_{sub}(S_2).}
\end{equation}
The marginal gain of state $s$ is defined as
\begin{equation}\label{eqn:marginal_gain}
    \revisefinal{\Delta(s | S_1) := R_{sub}(S_1 \cup \{s\}) - R_{sub}(S_1).}
\end{equation}
\revisefinal{Given a partial trajectory $\tau_{0:j} = (s_0, a_0, ..., s_j)$, let $S_j = \{s_0, ..., s_j\}$ denote the visited set. We write $\Delta (s_{j+1} \mid \tau_{0:j})$ as shorthand for $\Delta(s_{j+1} \mid S_j) = R_{sub} (S_j \cup \{s_{j+1}\}) - R_{sub} (S_j)$.}
$R_{sub}$ is a monotone submodular function if \revisenew{$R_{sub}(S_1) \leq R_{sub}(S_2),  \forall S_1 \subseteq S_2 \subseteq S$}.
\revisefinal{Here, $R_{sub} (S)$ denotes the submodular, set-based reward in the Submodular MDP, the cumulative reward accrued over the visited set $S$. 
The symbol $R(s,a)$ appears only in the preliminaries as standard MDP notation and is not used subsequently.}
In plain words, in CSMDP, adding an element $s$ to the \revisefinal{sub-subset $S_1$} will help at least as much as adding it to the \revisefinal{subset $S_2$}, in terms of rewards gaining.
Besides, the marginal gain \revisefinal{$\Delta(s | S_1)$} will be non-negative when the reward function $R_{sub}$ is monotone submodular.
This is exactly the case in track/river following tasks in environments like CliffCircular \cite{wang2024synergistic} and Safe Riverine Environment \cite{wang2024vision}, where repeated visiting of the visited states will not bring neither positive nor negative rewards.
\revisefinal{Safety is modeled separately via the cost signal $C(s,a)$.
In the constrained formulation, the goal is to exploit the history-dependent marginal-gain structure of reward while ensuring the cost constraint is satisfied, Equation \ref{eqn:cmdp}.
}
% The punishments to suboptimal policies will be reflected by the cost function $C(s,a)$.
% It is critical for safe RL agents in CSMDP to understand the submodular and non-Markovian nature of the reward function, leverage this knowledge, and optimize a policy that effectively balances reward maximization with cost minimization.

\subsection{Marginal Gain Advantage Estimator}\label{sec:methodologies-mgae}
MGAE builds on top of the algorithm First Order Constrained Optimization in Policy Space (FOCOPS, \cite{zhang2020first}), that solves the CMDP by first solving a constrained optimization problem in the non-parameterized policy space, then projecting the update policy back into the parameterized policy space.
FOCOPS solves the reward advantage optimization problem with a cost advantage constraint and a trust region constraint,
\begin{equation}\label{eqn:focops}
\begin{split}
    \pi_{k+1} &= \arg \max_{\pi \in \Pi} \underset{\substack{s \sim d_{\pi_k} \\ a \sim \pi}}{E} [A^R_{\pi_k} (s, a)] \\
    s.t. \hspace{5pt} J^{C_i}_{\pi_k} &\leq d_i - \frac{1}{1 - \gamma} \underset{\substack{s \sim d_{\pi_k} \\ a \sim \pi}}{E} [A^{C_i}_{\pi_k} (s, a)] \hspace{5pt} \forall i \\ 
    & \Bar{D}_{KL} (\pi || \pi_k) \leq \delta
\end{split}
\end{equation}
where $\Bar{D}_{KL} (\pi || \pi_k) = E_{s \sim \pi_k} [\Bar{D}_{KL} (\pi || \pi_k) [s]]$, $D_{KL} (\pi || \pi_k)$ is the Kullback–Leibler divergence between the target unparameterized policy $\pi$ and the current parameterized policy $\pi_k$, \revisefinal{$d_{\pi_k}$ is the state distribution induced by the policy $\pi_k$}, $\delta$ is the target KL divergence, $\gamma$ is the discount factor, $i$ is the cost constraint index number ($i \in \{1\}$ in our experiments, single cost is considered).

\revise{
Since the problem in Equation \ref{eqn:focops} is convex w.r.t the unparameterized policy $\pi$, therefore strong duality holds and the Lagrangian can be defined (Equation \ref{eqn:lagrangian}) and solved in its dual form (Equation \ref{eqn:duality}). 
}
% \revise{
% \begin{equation}\label{eqn:lagrangian}
%     \begin{split}
%     & L(\pi, \lambda, \nu) = 
%     \lambda \delta +
%     \nu \tilde{b} + \\ &
%     \underset{s \sim d_{\pi_k}}{E} [\underset{a \sim \pi}{E} [A^R_{\pi_k}] - \nu \underset{a \sim \pi}{E} [A^C_{\pi_k}] - \lambda D_{KL} (\pi || \pi_k)]
%     \end{split}
% \end{equation}
% }
\revise{
\begin{equation}\label{eqn:lagrangian}
    \revisenew{L(\pi, \alpha, \beta) = 
    \alpha \delta +
    \beta \tilde{b} + } 
    \underset{s \sim d_{\pi_k}}{E} [\underset{a \sim \pi}{E} [A^R_{\pi_k}] - \revisenew{\beta} \underset{a \sim \pi}{E} [A^C_{\pi_k}] - \revisenew{\alpha} D_{KL} (\pi || \pi_k)]
\end{equation}
}
\revise{
\begin{equation}\label{eqn:duality}
    \revisenew{\underset{\pi \in \Pi}{\max} \underset{\alpha, \beta \geq 0}{\min} L(\pi, \alpha, \beta) = 
    \underset{\alpha, \beta \geq 0}{\min}
    \underset{\pi \in \Pi}{\max} 
    L(\pi, \alpha, \beta)}
\end{equation}
}
\revise{
where $\tilde{b}=(1 - \gamma)(d - J^C_{\pi_k})$, \revisenew{$\alpha$} is the first Lagrange multiplier that controls the greediness of the algorithm, \revisenew{$\beta$} is the second Lagrange multiplier that acts as a cost penalty term.
}
\revise{
Solving the inner maximization in the dual problem in Equation \ref{eqn:duality} results in the optimal unparameterized policy,
}
\revise{
\begin{equation}\label{eqn:focops-optimal}
    \pi^*(a \mid s) = \frac{\pi_{\theta_k} (a \mid s)}{\revisenew{Z_{\alpha, \beta} (s)}} 
    \exp(\revisenew{\frac{1}{\alpha}} (A^R_{\pi_{\theta_k}}(s,a) - \revisenew{\beta} A^C_{\pi_{\theta_k}}(s,a)))
\end{equation}
}
\revise{
where \revisenew{$Z_{\alpha, \beta} (s)$} is the partition function that ensures Equation \ref{eqn:focops-optimal} is a valid probability distribution.
Substituting $\pi^*$ into Equation \ref{eqn:duality} results in the minimization problem to solve the optimal values for $\alpha$ and $\beta$.
}

To project the optimal update policy $\pi^*$ to the parameterized policy $\pi_\theta$, 
% let $\tau = (s_0, a_0, s_1, a_1, ..., s_{T-1}, a_{T-1})$ represent a trajectory of length $T$, then 
\revisefinal{minimize the loss function}
\begin{equation} \label{eqn:focops-loss}
    \mathcal{L}(\theta)=\underset{\revisefinal{s \sim d_{\pi_{\theta_k}}}}{\mathbb{E}}\left[D_{K L}\left(\pi_\theta \| \pi^*\right)[\revisefinal{s}]\right].
\end{equation}
The policy gradient is 
\begin{align} \label{eqn:focops-grad}
    & \nabla_\theta \mathcal{L}(\theta) = 
    \underset{\revisefinal{s \sim d_{\pi_{\theta_k}}}}{\mathbb{E}}\left[\nabla_\theta D_{K L}\left(\pi_\theta \| \pi^*\right)[\revisefinal{s}]\right] & \nonumber \\ 
    & = \nabla_\theta D_{K L}\left(\pi_\theta \| \pi_{\theta_k}\right)[\revisefinal{s}] - & \nonumber \\
    & \revise{\frac{1}{\alpha}} \underset{\revisefinal{s \sim d_{\pi_{\theta_k}}}}{\mathbb{E}} 
    \left[ \sum_{i=0}^{T-1} \frac{\nabla_\theta \pi_\theta(a \mid s)}{\pi_{\theta_k}(a \mid s)}
    \left(
    A_R^{\pi_{\theta_k}}(s_i, a_i)
    - \revisenew{\beta} A_C^{\pi_{\theta_k}}(s_i, a_i)\right)\right]. &
\end{align}
%% also explain other symbols
%where $\lambda$ is the first Lagrangian multiplier that controls the greediness of the algorithm, $\nu$ is the second Lagrangian multiplier that acts as a cost penalty term. 
\revise{
We refer readers to \revisefinal{Appendix B} of \cite{zhang2020first} for detailed proof.
}
\revisefinal{
In implementation we estimate the expectation in Equation \ref{eqn:focops-loss} and Equation \ref{eqn:focops-grad} by averaging the per-step loss along the most recent on-policy rollout rather than sampling isolated states from a buffer.
Partial observability and the history-dependent submodular reward require computing the recurrent hidden state (more in Section \ref{sec:methodologies-cade}) along a contiguous trajectory, and reconstructing them for random samples would require re-encoding full episodes after each policy update.
This full trajectory "sampling" does not violate the validity of policy loss of FOCOPS.
}

\begin{figure}[h]
    \centering
    \includegraphics[width=0.8\linewidth]{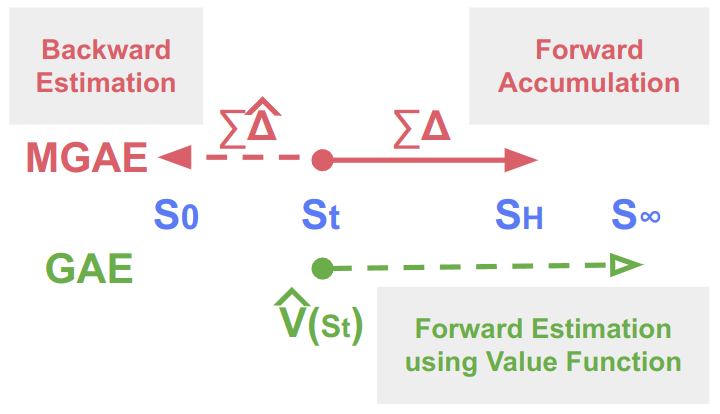}
    \caption{Comparison of Marginal Gain Advantage Estimation (MGAE) and Generalized Advantage Estimation (GAE). MGAE uses backward-looking estimation to calculate cumulative marginal gains, while GAE uses forward-looking value function estimation.}
    \label{fig:mgae-gae-comp}
\end{figure}

\revisefinal{
Let $\tau = (s_0, a_0, s_1, a_1, ..., s_{T-1}, a_{T-1})$ represent a trajectory of length $T$,
}
the reward advantage $A_R^{\pi_{\theta_k}}(s_i, a_i)$ is defined in a submodularity-inspired trajectory-wise way, named Marginal Gain Advantage Estimation (MGAE) for reward,
\begin{equation}\label{eqn:sub_advantage}
    A_R^{\pi_{\theta_k}}(s_i, a_i) = 
    \sum_{j=i}^{T-1} \Delta (s_{j+1} \mid \tau_{0:j}) - 
    b(\tau_{0:i}),
\end{equation}
\revisefinal{where $\Delta (s_{j+1} \mid \tau_{0:j}) \equiv \Delta (s_{j+1} \mid S_j)$ represents the true marginal gain of the state $s_{j+1}$ given the trajectory $\tau_{0:j}$ up to horizon $j$ (Equation \ref{eqn:marginal_gain})}, and $b(\tau_{0:i})$ is the baseline function dependent on the trajectory up to horizon $i$ that helps to reduce the variance of reward advantage estimation \cite{greensmith2004variance}. 
The important part here is how to choose the baseline function. It should represent a reasonable anticipation of the future cumulative rewards that can guide the policy improvement by comparing with the actual partial return to form the reward advantage. 
As proved in \cite{prajapat2023submodular}, baseline can be any history-dependent function.  
We choose this baseline function to be the agent's current best guess of the remaining rewards along the current trajectory to further reduce the variance,
%% Note the symbol j needs to align with above i
\begin{equation}\label{eqn:sub_baseline}
    b(\tau_{0:i}) = \underset{\tau \sim \pi_{\theta}}{\mean} 
( {\sum_{i=0}^{T-1} \Delta(s_{i+1} \mid \tau_{0:i})} ) - \sum_{i=0}^{j}  \hat{\Delta} (s_{i+1} \mid \tau_{0:i}).
\end{equation}
The left part in Equation \ref{eqn:sub_baseline} is the mean trajectory return across a sliding window of trajectories up to the baseline calculation ($\tau \sim \pi_{\theta}$), whereas the right part is the sum of estimated marginal gains up to step $j$ along the current trajectory. $\hat{\Delta} (s_{i+1} \mid \tau_{0:i})$ is the estimated marginal gain of the state $s_{i+1}$ given the trajectory $\tau_{0:i}$ up to horizon $i$.
\revisefinal{
MGAE uses a history-aware, action-independent baseline aligned with the marginal-gain structure. 
It is one practical choice among many unbiased baselines. 
We select it for simplicity and for direct correspondence to the submodular objective.
Exploring alternative baselines is promising in future work.
}
% The illustrative comparison between marginal gain-based MGAE and the value critic based GAE \cite{schulman2015high} is visually presented in Figure \ref{fig:mgae-gae-comp}.
\revisefinal{
Figure \ref{fig:mgae-gae-comp} contrasts MGAE, which leverages marginal gains, with GAE \cite{schulman2015high}, which relies on a value critic.
}

% Figure showing difference between MGAE and GAE

% Relationship with GAE (TD-based methods)
MGAE is designed to handle non-Markovian rewards in SMDPs, where future rewards depend on historical trajectory rather than just the current state. 
Unlike GAE, which bootstraps future rewards using a value critic, MGAE looks backward by leveraging a recurrent reward estimator to approximate immediate rewards. 
This avoids bias from value function errors and ensures advantage estimation aligns with the cumulative nature of submodular rewards. 
By focusing on historical gains rather than future predictions, MGAE provides a more stable and accurate policy update in environments where rewards depend on past exploration.
\revisenew{
On the other hand, by conditioning the baseline on the evolving reward estimator (rather than on static Monte Carlo returns), we ensure that the baseline itself benefits from further data, much like bootstrapped value functions undergo continual refinement. 
This design also facilitates online learning or continual learning during deployment, where the learned reward estimator can still update the baseline of MGAE for policy update.
}

% Relationship with REINFORCE
MGAE shares similarities with REINFORCE \cite{williams1992simple} in that it avoids bootstrapping from a learned value function and instead relies on cumulative returns to estimate advantage. 
However, unlike standard REINFORCE, which directly uses \revisenew{actually observed} rewards, MGAE incorporates an estimated reward function, making advantage computation more adaptive to the agent’s evolving perception of \revisenew{marginal gain in SMDPs}. 
Compared to REINFORCE with baselines \cite{sutton1999policy, keane2022variance}, which subtracts a state-dependent value function to reduce variance, MGAE instead employs a Monte Carlo baseline based on the agent’s historical episodic returns. This baseline dynamically adjusts to the agent’s recent performance trends, maintaining a \revise{less biased and responsive} advantage estimation in SMDPs.

\subsection{Semantic Dynamics Model}\label{sec:methodologies-sdm}
\begin{figure}[h]
    \centering
    \includegraphics[width=0.98\linewidth]{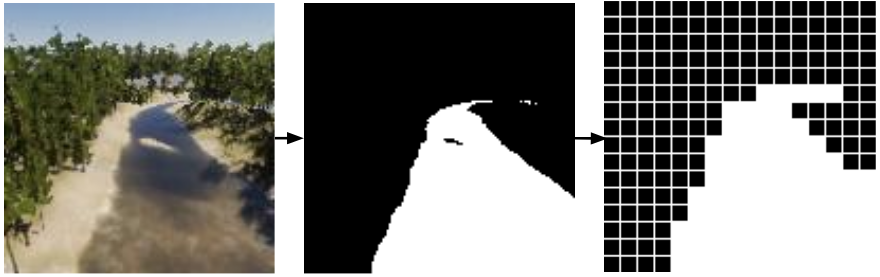}
    \caption{Observation images in Safe Riverine Environment. From left to right: RGB image, water semantic mask, patchified water semantic mask as RL observation.}
    \label{fig:patchified}
\end{figure}

% Key aspects of dynamics model in vision-driven safe rl:
% 0. Motivate why dim reduction is necessary
% 1. How to Form the Compact Visual Representation
% 2. How to Build the Dynamics of this Visual Representation
% 3. How to Use the Dynamics for MBRL to Learn Performant and Safe Policies
% 4. Exploration vs. Exploitation
% 5. Multi-step Dynamics for long-term predictions
% 6. Generalization Across Environments
When processing high-dimensional visual inputs, reducing the complexity to a more compact and meaningful representation is essential for efficient decision making in reinforcement learning.
In traditional approaches, encoding RGB vision inputs using techniques like Variational Autoencoders (VAE) \cite{kingma2013auto} or other similar methods often results in embeddings that, while compact, lack interpretability and structured geometric information. 
These embeddings may obscure critical spatial relationships necessary for precise decision-making in complex environments. 
To address this, we preprocess the RGB images via semantic segmentation to extract a water mask, which focuses the model’s attention on the most relevant environmental semantic features. 
Subsequently, we apply patchification to the water mask according to some threshold criterion like grid whose portion of water pixels exceeds $50\%$ is deemed as a water patch, reducing it to a coarser, yet spatially meaningful, representation (Figure \ref{fig:patchified}). 
This approach ensures that the downstream RL model receives input that retains the essential geometric and contextual structure of the scene, improving both the interpretability and the efficiency of the learning process. 

To further improve spatial consistency, our method employs a geometrically constrained semantic dynamics model (SDM), where a neural network estimates the homography between consecutive semantic observations.
This approach assumes that a significant portion of pixels in the image represents a flat surface (e.g., water) in the 3D environment, allowing the homography to approximate the geometric transformation between frames effectively.
Additionally, the model assumes that most objects in the scene are static, which helps simplify the estimation process.
By preserving the spatial transformations under these assumptions, our method enables the RL model to better capture short-term dynamic changes in the environment. 
%By integrating this structured approach, we provide a more semantically rich and geometrically coherent input to the RL model, which enhances not only the effectiveness of learning and planning but also contributes to safer decision-making by improving the predictability of future states and ensuring consistency in sequential observations.

The transition of first person view (semantic) images can be modeled as a $3 \times 3$ homography transformation matrix $H$ that linearly maps one homogeneous pixel coordinate $x = \left(u, v, 1 \right)^T$ to another $x' = \left(u', v', 1 \right)$,
\begin{equation}
    \left[\begin{array}{l}
        u^{\prime} \\
        v^{\prime} \\
        1
        \end{array}\right] =
    \left[\begin{array}{lll}
        h_{11} & h_{12} & h_{13} \\
        h_{21} & h_{22} & h_{23} \\
        h_{31} & h_{32} & h_{33}
    \end{array}\right]
    \left[\begin{array}{l}
        u \\
        v \\
        1
    \end{array}\right]
\end{equation}
Since only ratios of a homography matrix's elements matter, $H$ only has 8 degrees of freedom. Due to the high variance in the magnitude of elements in $H$, it would be difficult to enforce $H$ to be non-singular \cite{nguyen2018unsupervised}. Thus we adopted 4-point homography parameterization $H_{4pt}$ \cite{baker2006parameterizing}, which is a $4 \times 2$ matrix that contains the four corner coordinates' variations. 
Given current patchified semantic state $s_t$, current discrete action $a_t$ and next patchified semantic state $s_{t+1}$, the multi-layer perceptron $f(s, a) = \Delta x^C_k = (\Delta u^C_k, \Delta v^C_k), k \in \{ 1,2,3,4 \}$ in SDM outputs the 4 corners offsets, which can be used to build the homography $H$ considering the point correspondence between $x$ and $x'$,
\begin{equation}\label{eqn:corner-correspondence}
    u^{\prime}=\frac{h_{11} u+h_{12} v+h_{13}}{h_{31} u+h_{32} v+h_{33}} ; 
    v^{\prime}=\frac{h_{21} u+h_{22} v+h_{23}}{h_{31} u+h_{32} v+h_{33}}.
\end{equation}
The homography $H$ is then applied to $s_t$, warping the current state $s_t$ to get the predicted next state $s'_{t+1}$, 
\begin{equation}\label{eqn:sdm}
    s'_{t+1} = \mathcal{W}(s_t, \mathcal{P}(x^{CF}, x^{CF} + f(s_t, a_t))).
\end{equation}
$x^{CF}$ is the coordinates of 4 fixed corner pixels in the semantic state with known rows $r$ and columns $c$,
\begin{equation}
    x^{CF} = 
    \begin{bmatrix}
        0 & r-1 & r-1 & 0 \\
        0 & 0 & c-1 & c-1 
    \end{bmatrix}^{T},
\end{equation}
and the predicted corner coordinates of the next semantic state $s'_{t+1}$ is
\begin{equation}
    x^{CF'} = x^{CF} + f(s_t, a_t).
\end{equation}
$\mathcal{P}$ is the perspective transform function that calculates the homography $H$ given 2 pairs of quadrangles ($x^{CF}$ and $x^{CF'}$) that have corner correspondence,
\begin{equation}\label{eqn:perspective}
    H = \mathcal{P} (x^{CF}, x^{CF'}) = A^{-1} b,
\end{equation}
where $A$ is an $8 \times 8$ matrix formed by the coordinates of $x^{CF}$ and $x^{CF'}$ by the relation in Equation \ref{eqn:corner-correspondence}, $b$ is the $8 \times 1$ vector of flattened non-homogeneous coordinates of $x^{CF'}$.
$\mathcal{W}$ is the warping function that perspectively transforms every pixel in the current semantic state $s_t$ by the estimated homography $H$,
\begin{equation}
    s'_{t+1} = \mathcal{W} (s_t, H).
\end{equation}
Both $\mathcal{P}$ and $\mathcal{W}$ are differentiable functions in Kornia package \cite{riba2020kornia}, for effective learning of the geometrical constraint in the vision transition dynamics.

To train the semantic dynamics model, soft Intersection over Union (IoU) loss (Jaccard loss) between the homography-transformed state $s'_{t+1}$ and the ground truth next state $s_{t+1}$ is used,
\begin{equation}\label{eqn:jaccard}
    \mathcal{L}_{Jaccard} = 1 - \frac{\sum_{i=0}^{N-1} p_i g_i}{\sum_{i=0}^{N-1} p_i + \sum_{i=0}^{N-1} g_i - \sum_{i=0}^{N-1} p_i g_i},
\end{equation}
where $p_i$ and $g_i$ are the semantic patch values in the predicted next water mask $s'_{t+1}$ and the ground truth next water mask $s_{t+1}$, $N$ is the number of semantic patches in a state.
All vacant patches after transformation will be filled $0.5$ as a neutral uncertain semantic value, since in this work we only consider binary semantic observation. More specifically, $0$ is non-cliff and $1$ is cliff in CliffCircular environment, $0$ is non-water and $1$ is water in Safe Riverine Environment, Section \ref{sec:experiments-csmdp}.

% Assumptions
SDM is valid on the basis of several assumptions:
\begin{itemize}
    \item Flat Water Surface: A substantial portion of the scene consists of a flat water surface, which is a valid assumption in riverine environments. This flat surface allows the use of homography transformations to approximate the geometric relationships between consecutive observations. \footnote{However, in the CliffCircular environment, this homography assumption is not required since the agent's observation is the nadir view of a flat surface, thus the homography matrix (more specifically, the translational elements in it) can be perfectly learned.}  
    %\item Static Environment: The majority of objects in the scene are static, which is a reasonable assumption in riverine environments, where moving objects, such as boats or wildlife, are rare and typically have minimal impact on the overall scene dynamics.
    \revise{
    \item Static Environment: The majority of objects in the scene are assumed to be static, which is generally valid in riverine environments where moving objects, such as boats or wildlife, are infrequent. Small dynamic obstacles, if accurately segmented, are unlikely to significantly alter the patchified water mask due to their minimal impact on large-scale scene geometry. However, large or dense dynamic obstacles could disrupt SDM learning, as homography transformations may capture false correspondences caused by object motion rather than camera viewpoint changes, leading to erroneous dynamics predictions.
    }
    \item Reliable Semantic Segmentation: The semantic segmentation model provides sufficiently high-quality results, ensuring that any false positives or false negatives in predicted pixels are minimal and fall within an acceptable margin of error for robust dynamics modeling.
\end{itemize}

\subsection{Constrained Actor Dynamics Estimator Architecture}\label{sec:methodologies-cade}
\begin{figure}[h]
    \centering
    \includegraphics[width=\linewidth]{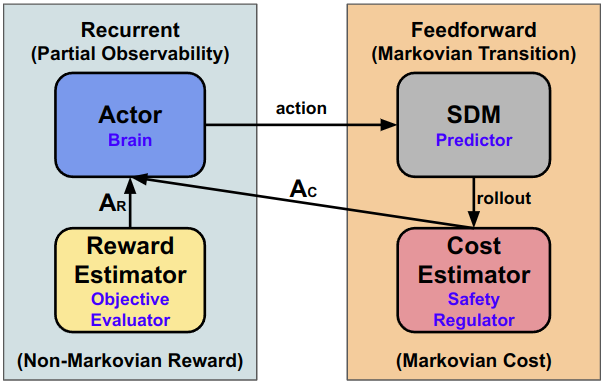}
    \caption{Constrained Actor Dynamics Estimator (CADE) architecture to solve CSMDP. \revisenew{The reason of choosing \textit{Recurrent} or \textit{Feedforward} network for each module is explicitly noted near the module block, and the character each module plays is noted inside the module block.}}
    \label{fig:cade}
\end{figure}

% CADE intro
CADE architecture consists of 4 major components: an actor as the brain, an immediate reward estimator as the objective evaluator, a semantic dynamics model as the predictor, and an immediate cost estimator as the safety regulator, Figure \ref{fig:cade}. 
The estimated immediate rewards by the reward estimator are used to calculate the reward advantage $A_R$ by MGAE (Section \ref{sec:methodologies-mgae}). 
The SDM (Section \ref{sec:methodologies-sdm}) is used to predict future observations given the current observation and subsequent actions.
The cost estimator estimates the immediate costs of the predicted observations to calculate the cost advantage $A_C$, regulating the surrogate advantage in policy gradient method (Equation \ref{eqn:focops-grad}).
% CADE structure
% Recurrent part
Instead of augmenting the state space $S$ with current horizon step as in \cite{prajapat2023submodular}, we use shared recurrent neural network (GRU \cite{chung2014empirical} in our experiments) for both the actor and the reward estimator, followed by separate Multi-Layer Perceptron (MLP) networks for each.
This design is for capturing the temporal dependencies in the agent's history due to the non-Markovian nature of the reward and partial observability of the first-person view observation.
Since incorporating the entire history into the state is infeasible, policy is commonly conditioned on a compact representation of past interactions \cite{hogewind2022safe}. 
The GRU layer acts as a memory of past observations and actions, providing the actor with a richer context when making decisions.
By sharing the GRU layer, both the actor and reward estimator can leverage a common representation of past information, allowing for more coherent learning and reduced redundancy.

% Feedforward part
For the cost estimator, we design its network as a feedforward MLP $\hat{C}$ since the cost function $C$ in our environments is Markovian, depending only on the current observation (details in Table \ref{tab:env_comparison}).
\revisenew{
This is based on the assumption that in a pure vision-driven task, the external immediate environmental hazards are encoded in the current perception.
On the contrary, internal variables like remaining battery or cumulative wear reflect the agent’s own evolving resource constraints, which introduce history-dependence because battery only decreases over time, rendering the costs non-Markovian.
Therefore, using a strictly Markovian cost function is a simplification of the learning problem considering safety, which does not require state augmentation. 
}
Hence, a MLP is sufficient for learning relationships that do not require memory of past events.
The SDM $\mathcal{M}$ also uses a MLP network since the state transition is Markovian, \revisenew{Figure \ref{fig:cade}}. 
\revisenew{
Techniques such as Monte Carlo Tree Search (MCTS \cite{coulom2006efficient}), which uses simulated rollouts to evaluate action sequences, Imagination-Augmented Agents (I2A \cite{racaniere2017imagination}), which learns a world model and perform imagined trajectories to inform policy decisions, Model-Based Value Expansion (MBVE \cite{feinberg2018model}), which controls for uncertainty in the model by only allowing imagination to fixed depth, and latent-space planners like PlaNet \cite{hafner2019learning}, which conduct fast online planning through a learned dynamics model in a compact embedding, all share the core idea of sampling imagined futures via a learned model rather than relying solely on real environment interactions to improve policy learning.
Inspired by the above works, and works using truncated n-step returns for bias-variance trade-off \cite{daley2024averaging, poiani2023truncating}, our method employs short horizon rollouts through learned vision dynamics and cost estimator networks, to construct the \textit{truncated Monte Carlo cost-to-go} in vanilla REINFORCE manner \cite{williams1992simple}, as the cost advantage for Lagrangian-based policy training. 
}
The cost advantage $\bar{A_C}$ is defined as the cumulative sum of the discounted predicted immediate costs over a prediction horizon $H$:
\begin{equation}\label{eqn:cost_adv_bar}
    \bar{A_C}(s_t, a_t) = \sum_{h=0}^{H-1} \gamma^h \hat{C}(s_{t+h}, a_{t+h}) = \sum_{h=0}^{H-1} \gamma^h \hat{C}(s'_{t+h+1}),
\end{equation}
where \revisefinal{horizon actions are induced by the current policy $a_{t+h} \sim \pi_{\theta}(\cdot | s_{t+h})$, $s'_{t+h+1} = \mathcal{M}(s_{t+h}, a_{t+h})$ is the predicted next state and $\hat{C} (s'_{t+h+1})$ is the predicted Markovian cost of this state}.
Unlike reward advantages that require a baseline to stabilize updates due to their long-term dependency, cost advantages in safety-constrained RL often prioritize short-horizon, immediate feedback to enforce constraints effectively \cite{yu2022towards}. 
Therefore, our formulation avoids bootstrapping by using the predicted cumulative cost itself as the cost advantage estimation.
%This aligns with the intuition that cost signals in safety-critical environments should provide immediate feedback rather than long-term estimates, mitigating potential inaccuracies from compounding model errors.
\revisefinal{We keep this short-horizon form across MDPs for consistency: when costs are terminal-only, the sum reduces to a single step; when costs are dense and state-dependent, using $H > 1$ captures near-term hazard accumulation more accurately and provides a better approximation to the cumulative cost-to-go, yielding a more reliable cost-advantage estimate.}

\revisefinal{To stabilize training and align magnitudes between the estimated reward and cost terms}, the cost advantage function is modulated using a sigmoid transformation\footnote{\revisefinal{We do not claim novelty for this transformation. It is a stabilization choice akin to widely used normalization/clipping heuristics.}}.
Specifically, a scaling factor 
$k$ and a cost baseline $c_b$ are applied as follows:
\begin{equation}\label{eqn:cost_adv}
    A_C(s_t, a_t) = \frac{1}{1 + e^{-k (\bar{A_C} - c_b)}}
\end{equation}
This transformation suppresses small predicted costs, preventing overly conservative behavior, while preserving the gradient signal for larger costs, ensuring that the agent remains sensitive to critical safety violations. 
% \revisenew{
% This non-linear transformation of cost advantage estimation is inspired by softmax temperature heuristics \cite{painter2023monte} in MCTS for exploration–exploitation trade-off, utility‐based risk transforms in risk‐sensitive RL \cite{garcia2015comprehensive, adjei2025prospect} and the bounded quadratic structure of Normalized Advantage Functions \cite{gu2016continuous} for stabilizing policy gradients and yielding bounded safety scores for interpretable thresholding.
% }
\revisefinal{
This mirrors common practice in deep RL of normalizing or clipping advantages/rewards to control gradient scales and improve robustness across batches and tasks (e.g., advantage clipping in PPO implementation in Stable-Baselines3 \cite{raffin2021stable}).
Besides, keeping cost- and reward-side signals on comparable scales makes the Lagrangian update less sensitive to batch-to-batch variation in raw magnitudes. 
The constraint mechanism ($\beta$ in Equation \ref{eqn:focops-optimal}) remains the primary tool for balancing performance and safety.
}
In our experiments, we select $k$ to be 8 and $c_b$ to be 0.5 in both environments.

\begin{figure}[h]
    \centering
    \includegraphics[width=\linewidth]{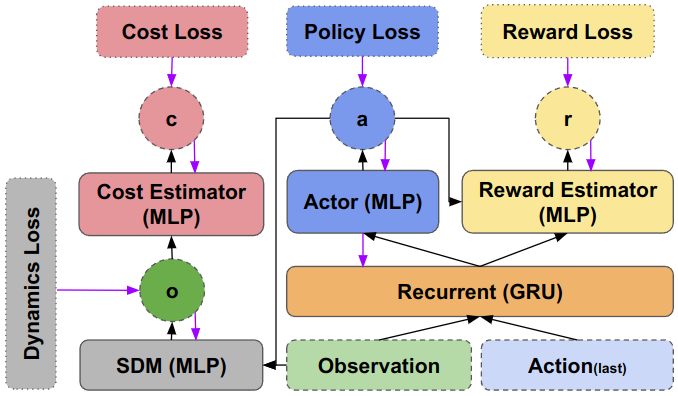}
    \caption{CADE computational graph. \textbf{a} is the current action to be executed, \textbf{r} is the estimated current reward, \textbf{c} is the estimated current cost and \textbf{o} is the predicted next observation. MLP is Multi-Layer Perceptron, GRU is Gated Recurrent Unit. Black arrows denote forward pass. Purple arrows represent the backpropagation pass. Best view in color.}
    \label{fig:cade-comp-graph}
\end{figure}

\revisenew{
The CADE computational graph is shown in Figure \ref{fig:cade-comp-graph}. Overall, CADE contains five trainable modules:
\begin{itemize}
    \item \textit{Recurrent}: a GRU network that accepts the current observation and the last action, then outputs the hidden state that captures the historical information of past observations and actions.
    \item \textit{Actor}: a MLP that accepts the hidden state from GRU as input, and outputs the action. It is trained using the policy loss in Equations \ref{eqn:focops-loss} and \ref{eqn:focops-grad}, which updates both the \textit{Actor} and the \textit{Recurrent} networks.
    \item \textit{Reward Estimator}: a MLP that accepts the hidden state from GRU and the current action from the \textit{Actor} as inputs, and outputs the estimated immediate reward. The reward loss is the Mean Squared Error (MSE) computed with the actual reward. This loss only updates the \textit{Reward Estimator}, instead of the \textit{Recurrent} network, which reduces the complexities of multi-objective learning of the shared recurrent layers. 
    \item \textit{SDM}: A MLP that accepts the current observation and the current action from the \textit{Actor} as inputs, and outputs the predicted next observation. The dynamics loss is calculated by the misalignment between the homography-transformed estimated and ground-truth observations following Equations \ref{eqn:sdm} and \ref{eqn:jaccard}.
    \item \textit{Cost Estimator}: A MLP whose input is the current observation (either predicted or actual), and outputs the estimated cost associated with that observation, with the assumption that the cost is Markovian. The cost loss is the MSE computed with the actual cost. 
\end{itemize} 
}

The CADE training procedure is shown in Algorithm \ref{alg:cade}. 
CADE framework incorporates several key design choices to ensure stable and effective policy learning while adhering to safety constraints.
First, the Lagrange multiplier is upper-bounded to prevent excessive conservatism, ensuring the policy remains sufficiently exploratory throughout training. 
Second, an on-policy buffer stores collected trajectories, \revisenew{but unlike standard experience replay through random sampling, network updates occur episode by episode \cite{lee2019sample, kapturowski2018recurrent}}. This design ensures the GRU hidden state maintains an accurate representation of historical information, though it introduces higher variance in advantage estimation. However, this variance is mitigated by the MGAE method, which leverages historical trends in episodic rewards for stabilization.

%A crucial aspect of CADE is the shared GRU network between the actor and the reward estimator, which introduces interdependencies in gradient updates. Empirical results indicate that a higher reward scale (e.g., max(r) = 1) improves policy learning, likely due to the stronger gradients influencing policy refinement, whereas smaller rewards result in weaker policy updates. 

\begin{algorithm}
    \caption{Constrained Actor Dynamics Estimator Outline}
    \label{alg:cade}
\begin{algorithmic}
    \State \textbf{Input:} Initial policy $\pi_\theta$, reward estimator $\hat{\Delta}$, semantic dynamics model $\mathcal{M}$, cost estimator $\hat{C}$
\State \textbf{Hyperparameters:} Learning rate 
% $\alpha_\theta$, $\alpha_{\hat{\Delta}}$, $\alpha_C$, $\alpha_{\mathcal{M}}$, 
\revisenew{$l_\beta$}, episodic cost budget $d$, \revisenew{KL divergence threshold $\Bar{\delta}$, Lagrange multiplier threshold $\Bar{\beta}$}
% \State Initialize parameters $\theta$, $\hat{\Delta}$, $\mathcal{M}$, $V_C$, $\lambda$
\While{maximum training steps not reached}
    \State Collect trajectory $\tau = \{s_0, a_0, r_0, c_0, s_1, \dots, s_T\}$ using policy $\pi_\theta$

    \State \textbf{Lagrange Multiplier Update:} Adjust \revisenew{$\beta$} based on constraint violation:
    \[
    \revisenew{\beta \leftarrow \min(\Bar{\beta} ,\max(0, \beta - l_{\beta} (d - J^C(\pi_{\theta}))))}
    \]

    \State \textbf{SDM Update:} Update the semantic dynamics model $\mathcal{M}$ using Equation \ref{eqn:sdm} by the Jaccard loss (Equation \ref{eqn:jaccard})

    \State \textbf{Cost Estimator Update:} Update $\hat{C}(s, a)$ using the collected costs by MSE loss $MSE(\hat{C}, C)$
    
    \State \textbf{Reward Advantage Update:} Calculate reward advantage using MGAE: Equations \ref{eqn:sub_advantage} and \ref{eqn:sub_baseline}

    \State \textbf{Cost Advantage Update:} Calculate cost advantage using Equation \ref{eqn:cost_adv}, with predicted rollout by SDM $\mathcal{M}$ and $\pi_\theta$

    \State \textbf{Reward Estimator Update:} Update $\hat{\Delta}(s, a)$ using the collected rewards by MSE loss $MSE(\hat{\Delta}, \Delta)$
        
    \State \textbf{Actor \& Reward Estimator Update:} Compute policy gradient using Equation \ref{eqn:focops-grad}, adjust parameters of both the policy and the shared GRU network
    % \[
    % \theta \leftarrow \theta - \alpha_\theta \nabla_\theta \mathcal{L}(\pi_\theta) 
    % \]
    
\EndWhile
\end{algorithmic}
\end{algorithm}

The reward and cost estimators, as well as the SDM, are trained as regression problems, facilitating efficient convergence. The reward estimator is conditioned on the GRU latent state and the agent’s action, regressing to the actual immediate reward received after the action. In contrast, the cost estimator depends only on the current observation, which encodes environmental hazards, and is trained to predict the cost at the next state rather than the current state, Equation \ref{eqn:cost_adv_bar}. 

To maintain trust region constraints, CADE employs the FOCOPS-style KL divergence masking, where policy updates are masked for steps exceeding a predefined divergence threshold. 
Furthermore, policy training is preempted when the KL divergence between the most recent and original policy exceeds a certain threshold, enforcing a strict trust region constraint. However, training of the reward estimator, SDM, and the cost estimator continues, as these components are not bound by the same trust region restrictions, ensuring their continual improvement even when policy updates are restricted.

\revise{
Additionally, reward advantage is normalized to ensure balanced policy gradient updates and prevent dominance by either component. 
\revisenew{Whereas reward advantage is not normalized when comparing different reward advantage estimation methods, Section \ref{sec:experiments-mgae}.}
\revisenew{
However, cost advantage is left unnormalized, since it has already been transformed to an absolute scale using sigmoid function, Equation \ref{eqn:cost_adv}.
This selective normalization is a heuristic that depends on the definitions of reward and cost in the target environment or task, thus may not generalize to different settings.  
}
%However, cost advantage is left unnormalized, as normalization was found to negatively impact the agent’s exploration ability by suppressing meaningful reward signals during training.
}

\section{EXPERIMENTS}\label{sec:experiments}
In this section, we detail the two vision-based track/river following simulation environments that are Constrained Submodular Markov Decision Processes in Section \ref{sec:experiments-csmdp}. 
In Section \ref{sec:experiments-mgae}, the detailed comparison between the proposed the marginal gain advantage estimation method and the other popular reward advantage estimation methods is illustrated. 
Section \ref{sec:experiments-sdm} introduces the implementation details of the proposed semantic dynamics model that learns the transitions of patchified semantic masks, as well as the other vision dynamics models that use latent representations of visual observations. 
In Section \ref{sec:experiments-cade}, we explain how online imaginary rollouts, based on the actor, the cost estimator and the semantic dynamics model, form the cost advantage for Lagrangian-based post-execution constrained policy update, and how they are used for direct on-execution action overlay using safety layer with cost planning, for both training-phase and evaluation-phase safety regulation in CSMDPs. 

\subsection{CSMDP Environments}\label{sec:experiments-csmdp}
In this work we experiment in two Constrained Submodular Markov Decision Process (CSMDP) gym environments: CliffCircular \cite{wang2024synergistic} and Safe Riverine Environment (SRE) \cite{wang2024vision} for the track/river following task. 
CliffCircular serves as a simplified, miniature version of the SRE, where the agent's objective is to navigate along the track (non-cliff grids) using local semantic observations, while avoiding stepping onto cliffs. 
Similarly, in SRE, the UAV agent's goal is to follow the river based on the first-person visual input from the simulated on-board camera. The agent operates within a safe 3D space defined by a Catmull-Rom spline \cite{catmull1974class}, which represents the river's central line, while also avoiding obstacles such as bridges and maintaining a heading reasonably aligned with the tangent of the river spline.
\revisenew{
The UAV agent in SRE has discrete actions (Table \ref{tab:env_comparison}), rendering the river following problem as waypoint control (in contrast to velocity control), with the assumption of the availability of a robust low-level controller that can accurately track the high-level waypoints or action sequences produced by the policy during deployment.
}
CliffCircular is created based on the CliffWalking environment in Gymnasium \cite{towers2024gymnasium}, SRE is built in Unity using the ML-Agents toolkit \cite{juliani2020}.

\begin{figure}[h]
    \centering
    \begin{subfigure}[b]{0.98\textwidth}
        \includegraphics[width=0.32\textwidth]{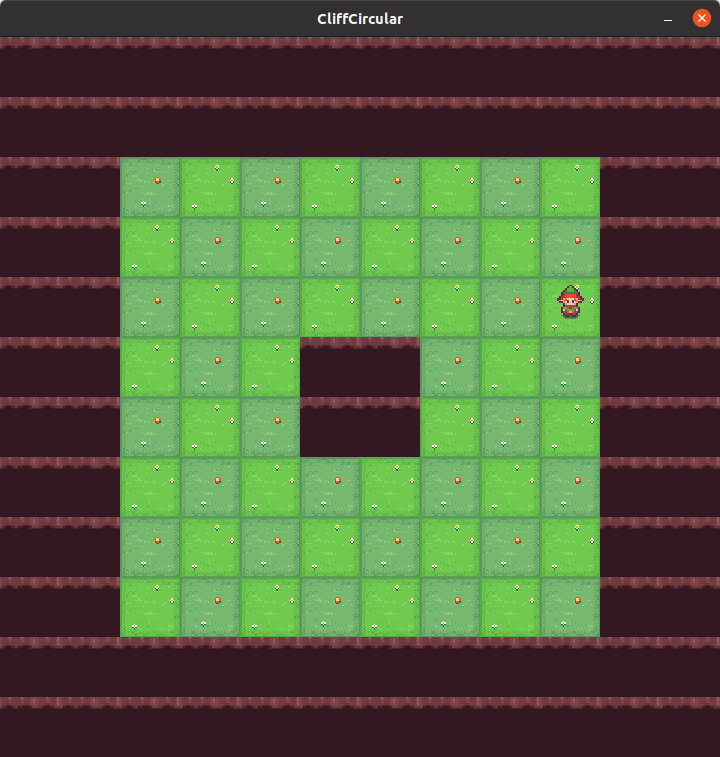}
        \includegraphics[width=0.32\textwidth]{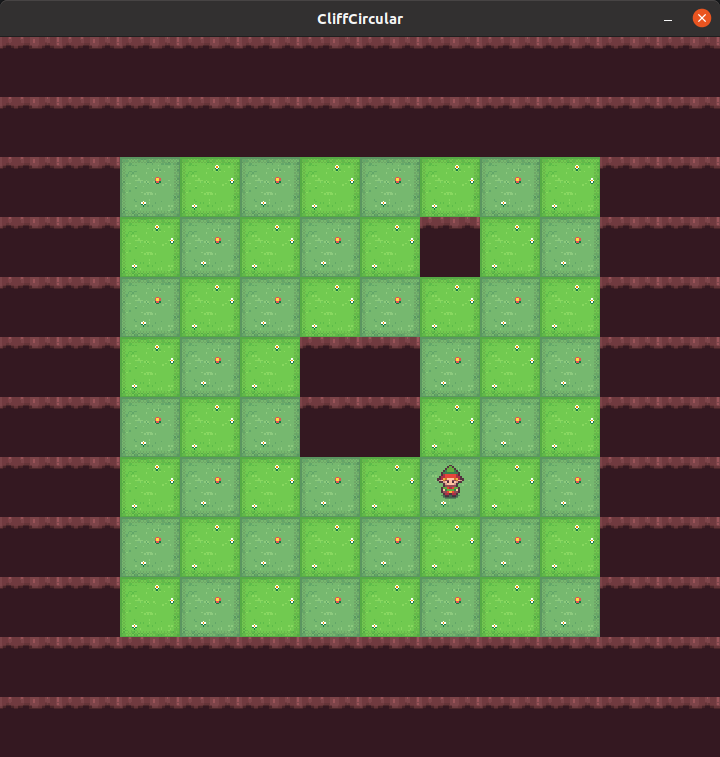}
        \includegraphics[width=0.32\textwidth]{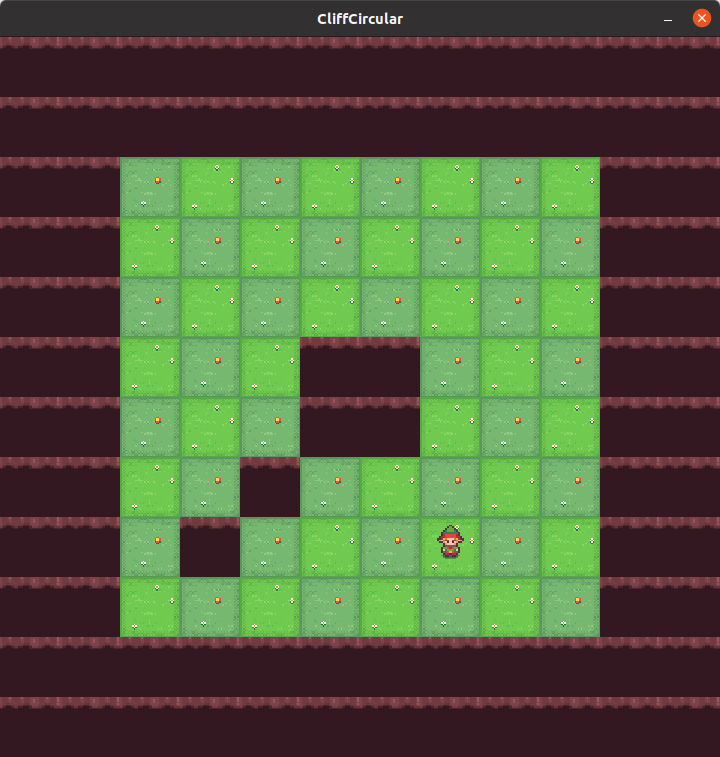}
        \caption{CliffCircular-v1 environment.}
        \label{fig:cliffcircular}
    \end{subfigure}
    
    \vspace{2pt}
    
    \begin{subfigure}[b]{0.98\textwidth}
        \includegraphics[width=0.32\textwidth]{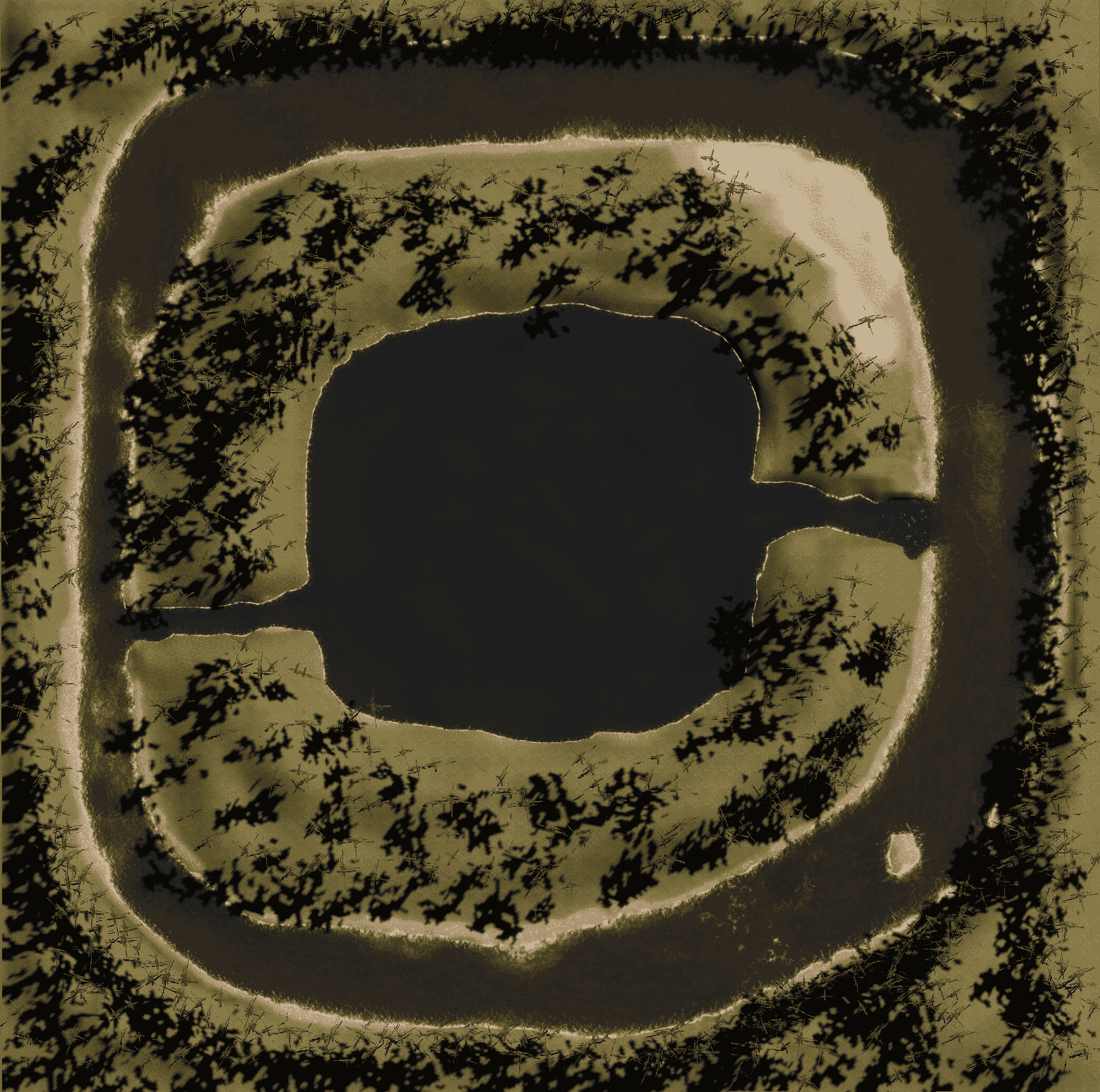}
        \includegraphics[width=0.32\textwidth]{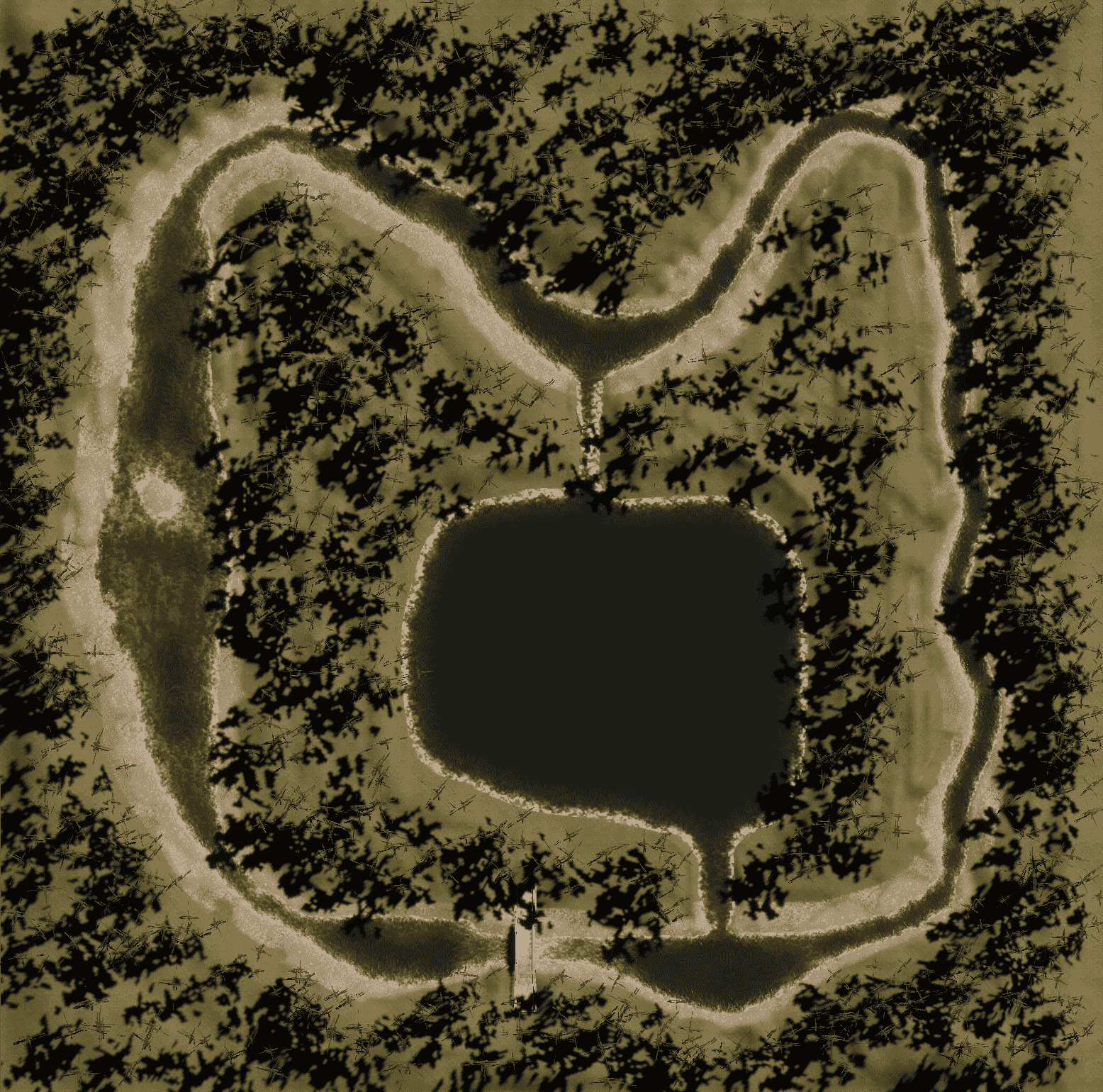}
        \includegraphics[width=0.32\textwidth]{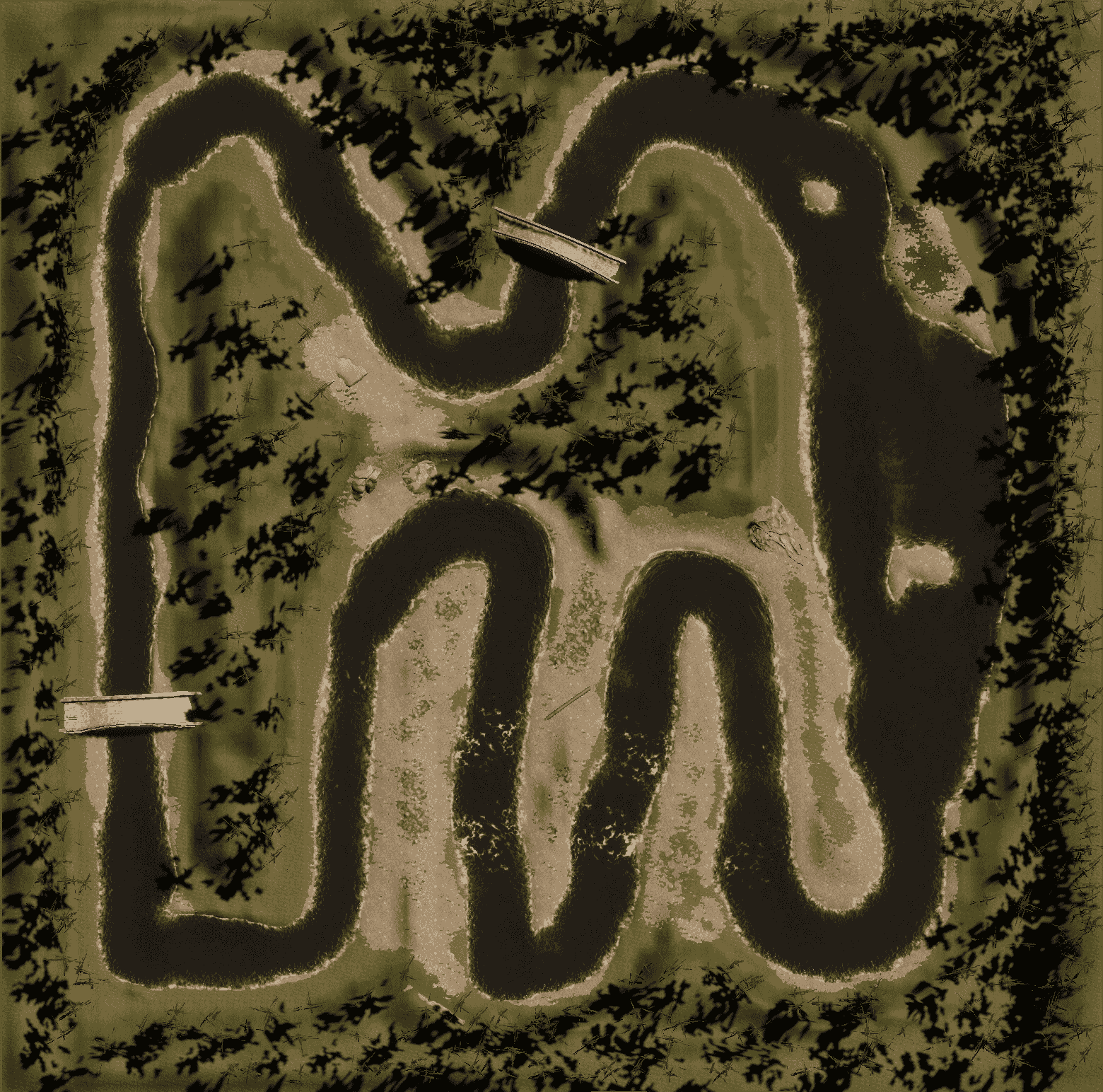}
        \caption{Orthographic view of Safe Riverine Environment (SRE).}
        \label{fig:SRE}
    \end{subfigure}
    
    \caption{Overview of two Constrained Submodular Markov Decision Process (CSMDP) environments, with increasing difficulty level \{easy, medium, hard\} from left to right.}
    \label{fig:env-overview}
\end{figure}

% Difference among 3 difficulty levels
Each environment set has three difficulty levels: easy, medium, and hard. In CliffCircular-v1, higher difficulty means more randomly spawned cliff grids at the episode's start. For SRE, increased difficulty adds more river turns and bridges.
All algorithms will be trained in the medium environment and tested across all three difficulty levels.

% Description of CliffCircular env
In the CliffCircular environment, the agent observes a $5 \times 5$ grid centered around itself, where each cell contains a binary semantic value indicating the presence of a cliff or non-cliff. The agent operates in a 5-dimensional action space: {no operation, up, right, down, left}. An invisible central track, consisting of 20 grids, serves as the reference for reward calculation. The agent earns a reward only when it visits an unvisited grid along this central track. The immediate cost is proportional to the percentage of cliff grids within a $3 \times 3$ area around the agent. A final cost of 1 is applied if the agent steps on a cliff grid, triggering an environment reset, where the agent is respawned on a safe, non-cliff grid.

% Description of SRE
Similarly, in the SRE, the agent receives a reward upon visiting unvisited segments of the river spline. The immediate cost depends on the portion and shape of water pixels in the agent’s patchified semantic observation. A large final cost is applied, with 0.5 for minor safety violations and 1 for severe violations, such as crashing into bridges or drifting off the river, resulting in the termination of the episode and the agent being randomly reset to a safe location.
% Assumption of low-level controller

% Summary of these two envs
Both environments feature non-Markovian rewards, as rewards depend on the agent’s historical observations and actions, and Markovian costs, as costs depend solely on the current observation. Additionally, with rewards and costs being decoupled, these environments guide the agent to balance task performance with safety considerations, framing them as Constrained Submodular Markov Decision Processes (CSMDP). Structured comparison of the two environments are in Table \ref{tab:env_comparison}.

\begin{sidewaystable}[htbp]
\centering
\caption{Thorough comparison of observation space, action space, non-Markovian reward functions, and Markovian cost functions for CliffCircular-v1 and SRE environments, with illustrative figures adapted from \cite{wang2024vision}.}
\label{tab:env_comparison}
\footnotesize
% \adjustbox{angle=90}{
\begin{tabular}{|P{2.5cm}|P{3.5cm}|P{2.8cm}|P{2.4cm}|P{2.6cm}|P{2.6cm}|}
% \begin{tabular}{|p{2cm}|p{3cm}|p{4cm}|p{3cm}|p{4cm}|p{4cm}|}
% \begin{tabular}{|c|c|c|c|c|c|}
\hline
\textbf{Environment} & \textbf{Observation Space} & \textbf{Action Space} & \textbf{Reward Function (Non-Markovian)} & \textbf{Cost Function (Markovian)} & \textbf{Scheme} \\ \hline

\textbf{CliffCircular} 
& \textbf{MultiBinary(25)}: \newline $5\times5$ semantic cliff mask around the agent. 
& {\textbf{Discrete(5)}: \newline \{no operation, up, right, down, left\}} 
& $\boldsymbol{1}$ for visiting a new central track grid; \newline $\boldsymbol{0}$ otherwise. 
& $\boldsymbol{1}$ for stepping on a cliff grid; \newline otherwise, proportional to the \textbf{percentage} of cliff grids in the $3\times3$ area around the agent. 
& \includegraphics[width=0.14\textwidth]{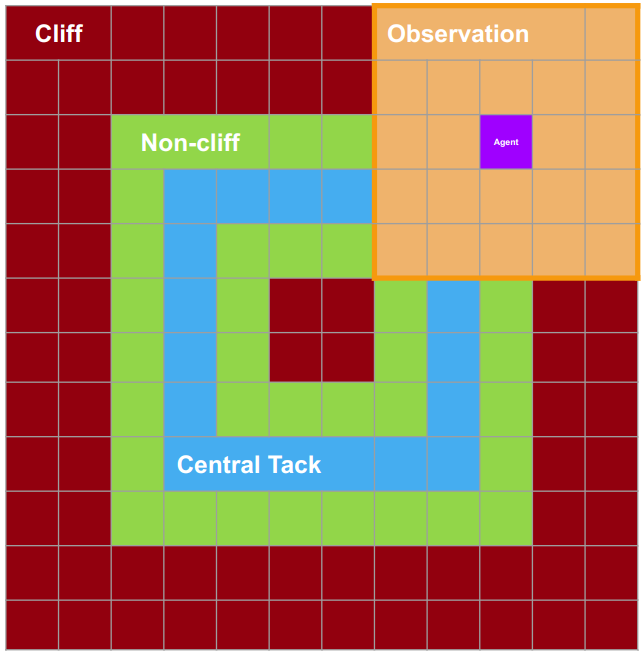} \\ \hline

\textbf{SRE} 
& \textbf{MultiBinary(256)}: \newline $16\times16$ patchified semantic water mask in the agent’s first-person view. 
& \textbf{MultiDiscrete ([3,3,3,3])}: \newline \{vertical translation, horizontal rotation, longitudinal translation, latitudinal translation\} 
& $\boldsymbol{1}$ for each newly visited line segment on the river spline; $\boldsymbol{0}$ otherwise. 
& $\boldsymbol{0.5}$ for minor reset conditions (e.g., excessive yaw deviation); $\boldsymbol{1}$ for severe resets (e.g., leaving the river spline volume). 
& \includegraphics[width=0.14\textwidth]{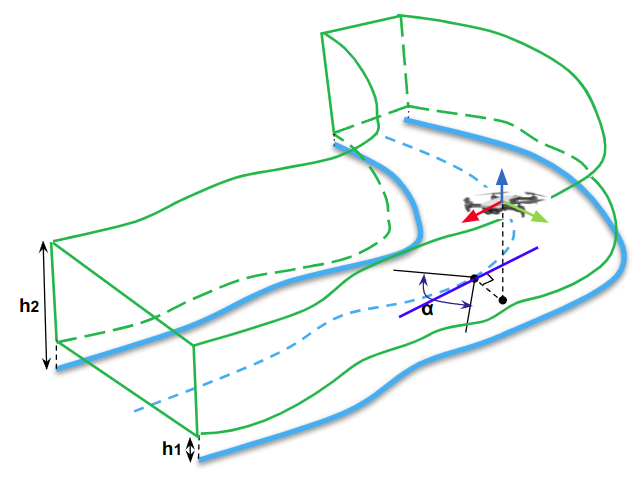} \\ \hline

\end{tabular}
% }
\end{sidewaystable}

\subsection{Marginal Gain Advantage Estimation Examination}\label{sec:experiments-mgae}
The proposed Marginal Gain Advantage Estimation (MGAE) method for reward (Equation \ref{eqn:sub_advantage} and Equation \ref{eqn:sub_baseline}) is compared with the plain Temporal Difference (TD, \cite{williams1991function}) advantage estimation method, Generalized Advantage Estimation (GAE, \cite{schulman2015high}), its variant with reward-to-go as target reward value (GAE-RTG, \cite{schulman2015high}), \revisenew{its special Monte Carlo form when $\lambda = 1$ (REINFORCE, \cite{williams1992simple})} and the V-trace \cite{espeholt2018impala} method.
The detailed comparison is shown in Table \ref{tab:advantage_comparison}.

% Table of advantage methods comparison
\begin{sidewaystable}[htbp]
\centering
\caption{Comparison of Marginal Gain Advantage Estimation (MGAE) with existing advantage estimation methods.}
\footnotesize
\begin{tabular}{|P{3cm}|P{5cm}|P{2.5cm}|P{1.5cm}|P{2.5cm}|P{5cm}|}

\hline

\textbf{Method} & \textbf{Advantage Estimation Equation} & \textbf{Markovian Reward Assumption} & \textbf{Critic Dependence} & \textbf{Temporal Scope} & \textbf{Suitability} \\ \hline
\textbf{MGAE (Proposed)} & 
$\displaystyle A_j = \sum_{k=j}^{T} r_k + \sum_{i=0}^{j} \hat{r}_i - \text{mean}(R)$ & 
\revise{\textbf{No}} &
%; considers non-Markovian rewards. & 
No &
\revise{\textbf{Full trajectory}} & 
Environments with \revise{\textbf{submodular, history-dependent rewards}}. \\ \hline

\textbf{TD \cite{williams1991function}} & 
$\displaystyle A_t = r_t + \gamma V(s_{t+1}) - V(s_t)$ & 
Yes &
%; relies on Markovian property. & 
Yes &
One-step lookahead & 
Simple, Markovian environments. \\ \hline

\textbf{GAE \cite{schulman2015high}} & 
$\displaystyle A_t = \sum_{k=0}^{n-1} (\lambda \gamma)^k \delta_{t+k}, \, \newline 
\delta_t = r_t + \gamma V(s_{t+1}) - V(s_t)$ & 
Yes &
%; assumes Markovian property. & 
Yes &
Multi-step lookahead & 
Markovian environments needing a bias-variance trade-off. \\ \hline

\textbf{\revisenew{REINFORCE \cite{williams1992simple}}} & 
\revisenew{$\displaystyle A_t = G_t - V(s_t)$} &
\revisenew{No} &
\revisenew{Yes} & 
\revisenew{Multi-step lookahead} & 
\revisenew{Episodic tasks with variance controlled by a learned critic baseline.} \\ \hline

\textbf{V-trace \cite{espeholt2018impala}} & 
$\displaystyle \delta_t = \rho_t (r_t + \gamma V(s_{t+1}) - V(s_t)), \newline \rho_t = \min(c, \frac{\pi(a_t|s_t)}{\mu(a_t|s_t)})$ & 
Yes &
%; designed for off-policy corrections in Markovian environments. & 
Yes &
Multi-step lookahead (off-policy corrected) & 
Environments where off-policy correction is necessary (e.g., distributed training). \\ \hline

\end{tabular}

\label{tab:advantage_comparison}
\end{sidewaystable}

\revise{In RL, advantage estimation methods vary in how they utilize trajectory information to evaluate the benefit of specific actions within given states. 
The MGAE method distinguishes itself by leveraging the entire trajectory, incorporating both \revisenew{past estimated and future observed rewards} to calculate the advantage at each state. 
This comprehensive approach is particularly beneficial in environments where the reward function is submodular, meaning that the marginal gain from visiting new segments diminishes as more segments are explored. 
In such settings, the agent's historical actions significantly influence future rewards, making a trajectory-wide perspective essential.}

\revise{In contrast, traditional methods like TD learning, GAE, \revisenew{REINFORCE} and V-Trace primarily focus on future rewards:}
\begin{itemize}
    \item \revise{\textbf{TD}: Utilizes one-step lookahead, updating value estimates based on the immediate reward and the estimated value of the subsequent state. This approach is myopic, considering only the next step without accounting for the broader trajectory context.}
    \item \revise{\textbf{GAE}: Employs a multi-step lookahead, averaging temporal difference errors over multiple future steps to estimate advantages. While this reduces variance compared to single-step methods, it still relies solely on information following the current state, neglecting past trajectory data.}
    \item \revisenew{\textbf{REINFORCE with a baseline}: Implements GAE with $\lambda = 1$, often referred to as Monte Carlo advantage, which subtracts the state value estimate $V(s)$
    from the observed return $G_t = \sum_{t=0}^{N} \gamma^t R(s_t, a_t)$ to reduce variance without introducing bias.}
    \item \revise{\textbf{V-Trace}: Similar to GAE, V-Trace incorporates multi-step returns and adjusts for off-policy corrections using importance sampling. It focuses on future rewards and does not integrate past trajectory information into advantage estimates.}
\end{itemize}

\revise{The distinction between MGAE and these traditional methods becomes more pronounced in the context of SMDPs. 
In standard MDPs, the reward structure often encourages the agent to remain within certain high-reward regions of the state space. 
For example, in the classic inverted pendulum problem, maintaining the pendulum in an upright position yields continuous rewards, making state-value critics effective for estimating long-term returns.
However, in SMDPs, the reward structure is inherently different due to its submodular nature. 
Here, the agent receives diminishing returns for visiting previously explored regions, incentivizing continuous exploration of new areas. 
In such scenarios, traditional state-value critics may not accurately capture the true advantage of actions, as they do not account for the diminishing marginal gains associated with revisiting states. 
MGAE addresses this limitation by utilizing an immediate reward estimator that focuses on the marginal gain of each action, considering the entire trajectory to effectively guide the agent's exploration strategy.}

All reward advantage estimation methods are based on the same CADE architecture shown in Figure \ref{fig:cade} for fair comparison, meaning that the reward estimator for critic-based methods outputs the state value, and shares the same GRU layers with the actor, while preserving the MLP layers that accept the latent output from the GRU network.
Differently, the reward estimator's MLP network outputs the predicted immediate reward for MGAE method.
%, but the return value (cumulative discounted rewards) for the other methods. 
\begin{equation} \label{eqn:focops-grad-mage}
    \nabla_\theta \mathcal{L}(\theta) =  
    \nabla_\theta D_{K L}\left(\pi_\theta \| \pi_{\theta_k}\right)[s] -
    \revisenew{\frac{1}{\alpha}} \underset{\tau \sim \pi_{\theta_k}}{\mathbb{E}} 
    \left[ 
    \sum_{i=0}^{T-1} \frac{\nabla_\theta \pi_\theta(a \mid s)}{\pi_{\theta_k}(a \mid s)}
    A_R^{\pi_{\theta_k}}(s_i, a_i)
    \right]
\end{equation}
The cost advantage in Equation \ref{eqn:focops-grad} is disabled when calculating the policy gradient (resulting in Equation \ref{eqn:focops-grad-mage}) to isolate the effectiveness of the reward advantage estimator in driving task-related performance. 
In this comparison, the policy update solely reflects task performance feedback, without being influenced by environmental dangers or cost signals. 
This allows for a more focused comparison of how different reward advantage estimators contribute to maximizing task-specific rewards.
\revisefinal{Window size of MGAE's baseline (Equation \ref{eqn:sub_baseline}) is chosen as $10$ in both environments.}
Different advantage estimation methods are compared in their episodic returns during training in the \textit{Medium} level environment for the same set of 3 seeds, and during testing in the \textit{Easy} and \textit{Hard} levels for 30 episodes.

\subsection{Semantic Dynamics Model Evaluation}\label{sec:experiments-sdm}
In this section, we compare the performance of the proposed Semantic Dynamics Model (SDM) in the Safe Riverine Environment \footnote{Observation in CliffCircular environment is both binary and small in dimension, thus is not applicable for latent dynamics models for comparison.} against four other vision dynamics models: Semantic Dynamics Model with Multi-Layer Perceptron (SDM-MLP), Latent Dynamics Model (LDM), Latent Dynamics Model with Multi-Layer Perceptron (LDM-MLP), and a baseline model.
These models are evaluated based on their ability to predict the environment’s dynamics over a 10-step prediction horizon. 
We use two metrics for comparison at the patch level: IoU, which measures the overlap between the predicted and true semantic (water) patches, and L1 loss, which captures the absolute differences between the predicted and true patches. 
Water semantic masks in Safe Riverine Environment are obtained by the Unity Perception package \cite{unity-perception2022} as the ground truth.
Here we briefly introduce the core components of the four vision dynamics model, whose general comparison can be found in Table \ref{tab:model_comparison}.

\begin{sidewaystable}[htbp]
\centering
\caption{Comparison of SDM, SDM-MLP, LDM, LDM-MLP, and Baseline models for vision dynamics. MLP denotes multi-layer perceptron, RNN denotes recurrent neural network, GRU is gated recurrent network, VAE means variational autoencoder. Model sizes for LDM and LDM-MLP do not include the VAE model. Inference is on Intel 14700K CPU.}
\footnotesize
\begin{tabular}{|P{2cm}|P{3.5cm}|P{1.5cm}|P{2cm}|P{2cm}|P{2.4cm}|P{2.2cm}|P{2cm}|P{1.8cm}|}
\hline
\textbf{Model} & \textbf{Input Representation} & \textbf{Model Type} & \textbf{Recurrent Layer} & \textbf{VAE Embedding} & \textbf{Geometrical Constraint} & \textbf{Prediction} & \textbf{Inference Time (ms)} & \textbf{Model Size (\# Parameters)} \\
\hline
\textbf{SDM (Proposed)} & $128\times128$ water mask, patchified to $16\times16$ binary patches & MLP & No & No & Yes \newline (Homography) & Predicts corner patches deviations & $4.54 \pm 0.88$ & 42,184 \\
\hline
\textbf{SDM-MLP} & $128\times128$ water mask, patchified to $16\times16$ binary patches & MLP & No & No & No & Predicts next patchified mask & $1.40 \pm 0.47$ & 58,304 \\
\hline
\textbf{LDM} & $4\times128\times128$ (RGB + mask) input, passed through VAE to 64-dim latent vector & RNN & Yes (GRU) & Yes (pretrained VAE) & No & Predicts next latent vector & $16.83 \pm 1.64$ & 84,288 \\
\hline
\textbf{LDM-MLP} & $4\times128\times128$ (RGB + mask) input, passed through VAE to 64-dim latent vector & MLP & No & Yes (pretrained VAE) & No & Predicts next latent vector & $15.58 \pm 3.50$ & 21,248 \\
\hline
\textbf{Baseline} & $16\times16$ patchified water mask & - & No & No & No & Simply uses current mask as the next one & - & - \\
\hline
\end{tabular}

\label{tab:model_comparison}
\end{sidewaystable}

\begin{itemize}
    \item The \textbf{SDM} takes a $128 \times 128$ binary water mask, which is patchified into $16 \times 16$ patches, with each patch being $8 \times 8$ pixels. Each patch is then thresholded to produce a single binary semantic value, Figure  \ref{fig:patchified}. The input to the model is this patchified representation, along with the agent’s action. 
    %The model predicts the next semantic mask by estimating the homography transformation based on the corner pixel deviations, with the agent’s action influencing the predicted homography. 
    This method incorporates geometric constraints by modeling patch-to-patch transitions as homography transformations.
    \item Like SDM, the \textbf{SDM-MLP} model uses the same patchified water mask as input. However, instead of using a homography transformation, this model replaces the geometric transformation with a fully connected multi-layer perceptron (MLP) that learns to predict the next semantic mask based purely on learned representations. 
    SDM-MLP does not incorporate explicit geometric constraints and instead relies entirely on its learned representations from the input patches.
    \item The \textbf{LDM} uses a Recurrent State Space Model (RSSM, \cite{hafner2019learning}) framework  where the dynamics of the environment are learned in a latent space rather than directly from pixel-level or patch-level observations.
    The LDM uses a $4 \times 128 \times 128$ RGB+water mask as input. This input is encoded using a Variational Autoencoder (VAE, \cite{kingma2013auto}) to yield a 64-dimensional latent vector. The VAE is pre-trained and fixed during the dynamics training phase, ensuring stable latent space representations. 
    The Gated Recurrent Unit (GRU, \cite{chung2014empirical}) is used as the core recurrent layer to capture temporal dependencies between latent states. 
    The GRU predicts the next latent state, which is then decoded to produce the next observation.
    \item Similar to LDM, the \textbf{LDM-MLP} model uses the 64-dimensional latent vector obtained from the pre-trained VAE as input. However, instead of using a recurrent layer (GRU), the dynamics are modeled using an MLP that predicts the next latent state directly from the current latent state and action. 
    This non-recurrent architecture simplifies the dynamics learning and tests whether a non-temporal model can capture the environment’s dynamics as effectively as a recurrent model.
    To ensure a fair comparison, the predicted latent state of future observations from the LDM and LDM-MLP models are first reconstructed through the VAE, then patchified and thresholded in the same manner as the SDM models. 
    This ensures that the metrics reflect performance on the same semantic level across all models, with lower L1 loss and higher IoU indicating better performance.
    \item The \textbf{baseline} model assumes no temporal dynamics and uses the current observation as the next observation for all prediction steps. This serves as a simple comparison point to evaluate the predictive power of the dynamics models over a 10-step horizon.
\end{itemize}

The riverine dataset, manually collected in the medium-level safe riverine environment, consists of 1720 transitions for the training set and 492 transitions for the test set (a single episode). 
The dataset ensures that all agent actions are triggered during data collection, providing diverse transition dynamics for the models to learn.
\revise{
SDM and SDM-MLP with MLP network type have two linear layers both with dimension 64, and \textit{tanh} activation function.
GRU network in LDM has a hidden dimension of 128, with 1 hidden layer.
MLP network in LDM-MLP has two linear layers with dimensions $[128, 64]$, and \textit{relu} activation function.
The VAE model used by the latent dynamics models was pre-trained on the training dataset for 100 epochs with a batch size of 128. The VAE’s architecture consists of hidden dimensions [16, 32, 64, 128], resulting in a final latent vector dimension of 64. The VAE-encoded mean embedding (without variations) serves as the latent state for LDM and LDM-MLP.
}

\revise{
The static analysis of different vision dynamics models is also presented in Table \ref{tab:model_comparison}.
The proposed SDM has about half the model parameters of LDM, while being three times faster than it.
Both SDM and SDM-MLP have smaller model sizes and faster inference times than LDM and LDM-MLP, especially the latter models need to be combined with VAE models to work.
Capable of running at 20 Hz with a moderate model size, SDM shows its feasibility in being deployed for real-time state planning, on onboard devices with limited power and memory.
}

All models, except the baseline, were trained in one-step prediction for 30 epochs with a batch size of 64, using the Adam optimizer \cite{kingma2014adam} with a learning rate of 0.001. 
During testing, at each iteration, the prediction start index is shifted one step forward, and the models recursively predict the next 10 steps based on their own predictions. 
The testing statistics, including the mean and standard deviation of the IoU and L1 metrics, are summarized at each prediction step to assess the model's ability to predict future visual observations given the actions.

%While in field tests, semantic segmentation model is trained using the Aerial Fluvial Image Dataset (AFID) \cite{PURR4105} that has finely labeled pixels of water, and non-water objects like bridges, sand islands and etc. 

\subsection{Constrained Actor Dynamics Estimator Evaluation}\label{sec:experiments-cade}
By combining the MGAE method for reward advantage, the SDM for state planning and the cost estimator for cost advantage estimation, we construct the overall Constrained Actor Dynamics Estimator (CADE) architecture, as shown in Figure \ref{fig:cade}.
%This section details how safe policy learning progresses by incorporating "imaginary" rollouts based on the SDM and the cost estimator. 
In CADE, the Lagrangian-based method regulates policy learning by dynamically balancing reward and cost advantages, softly steering the agent toward safer behaviors during training. 
However, this approach does not enforce strict action constraints during execution. 
To explore the effectiveness of hard safety interventions, we \revisefinal{evaluate an alternative safety regulator}: \textit{cost-based planning}\revisefinal{, which is a safety layer method (more in Section \ref{sec:related-mbsrl} and \ref{sec:methodologies-cade})}.
\revisefinal{This safety layer is implemented within the CADE framework, leverages the SDM and the cost estimator to evaluate multiple candidate trajectories before selecting an action. It complements the soft Lagrangian training with a hard cost-based action filter that overrides the actor only at selection. }

% Details of safety layer methods
\revisefinal{As a safety-regulation strategy evaluated on the same backbone of CADE, for} cost-based planning, the agent generates a finite set of imagined trajectories, evaluates their cumulative estimated costs, and selects the first action of the trajectory with the lowest cost, attempting to enforce safety constraints during training. 
\revisefinal{This safety layer does not modify the actor’s training objective, it only overrides actions at selection time using SDM-predicted rollouts and cost estimates, providing a hard alternative to the soft Lagrangian training used in CADE.}
%The reward-cost trade-off planning method extends this by incorporating the reward estimator, balancing cumulative rewards and costs within each trajectory. The agent selects the first action on the trajectory with the highest net trade-off score, defined as the cumulative reward minus the cumulative cost, to ensure a balance between task performance and safety.
The safety layer is activated only after the first third of training to preserve the agent’s exploration ability and are engaged selectively—intervening only if all sampled trajectories starting from the policy-selected action exceed a predefined cost threshold. 
If no severe failures (high-cost events) are predicted, the safety layer does not override the policy's decision. 
\revise{
The safety layer has not been activated since the beginning of training because the SDM and the cost estimator have not yet converged.
} 
The log probability of the selected action is used for importance sampling in policy updates (Equation \ref{eqn:focops-grad}). 
This method provides direct \revisefinal{comparative} against the Lagrangian-based approach, highlighting the differences between hard safety constraints that intervene at execution time and soft constraints that shape policy updates over time.

% Experimental details
The prediction horizon $H$ for the Lagrangian-based cost advantage (Equation \ref{eqn:cost_adv_bar}) and the safety layer method are set to 1, to take advantage of the higher short-term prediction accuracy of SDM over other vision dynamics models. 
The cost budget $d$ for the Lagrangian-based method is set to 1, the maximum actual immediate cost, which is a strict restriction where a single-step violation would trigger the increase of the Lagrange multiplier. 
Similarly, the cost threshold over the prediction horizon $H$ for the safety layer method is set to 1 for the CliffCircular-v1 environment, and 0.5 for SRE, representing the minimum single-step cost that triggers episode termination (Table \ref{tab:env_comparison}). 
The safety layer will engage to override the original policy-selected action if the predicted incurred cost after this action exceeds this threshold.
Ten trajectories will be sampled at each step when safety layer with cost-based planning is activated.
The code is built upon infrastructural framework for SafeRL, OmniSafe \cite{ji2024omnisafe}.

\section{RESULTS AND ANALYSIS}\label{sec:results}
In this section, we provide qualitative and quantitative results and analysis of the proposed MGAE method (Section \ref{sec:results-mgae}), SDM (Section \ref{sec:results-sdm}), and the CADE architecture (Section \ref{sec:results-cade}).

\subsection{Marginal Gain Advantage Estimator}\label{sec:results-mgae}
\begin{figure}[H]
    \centering
    \includegraphics[width=0.49\linewidth]{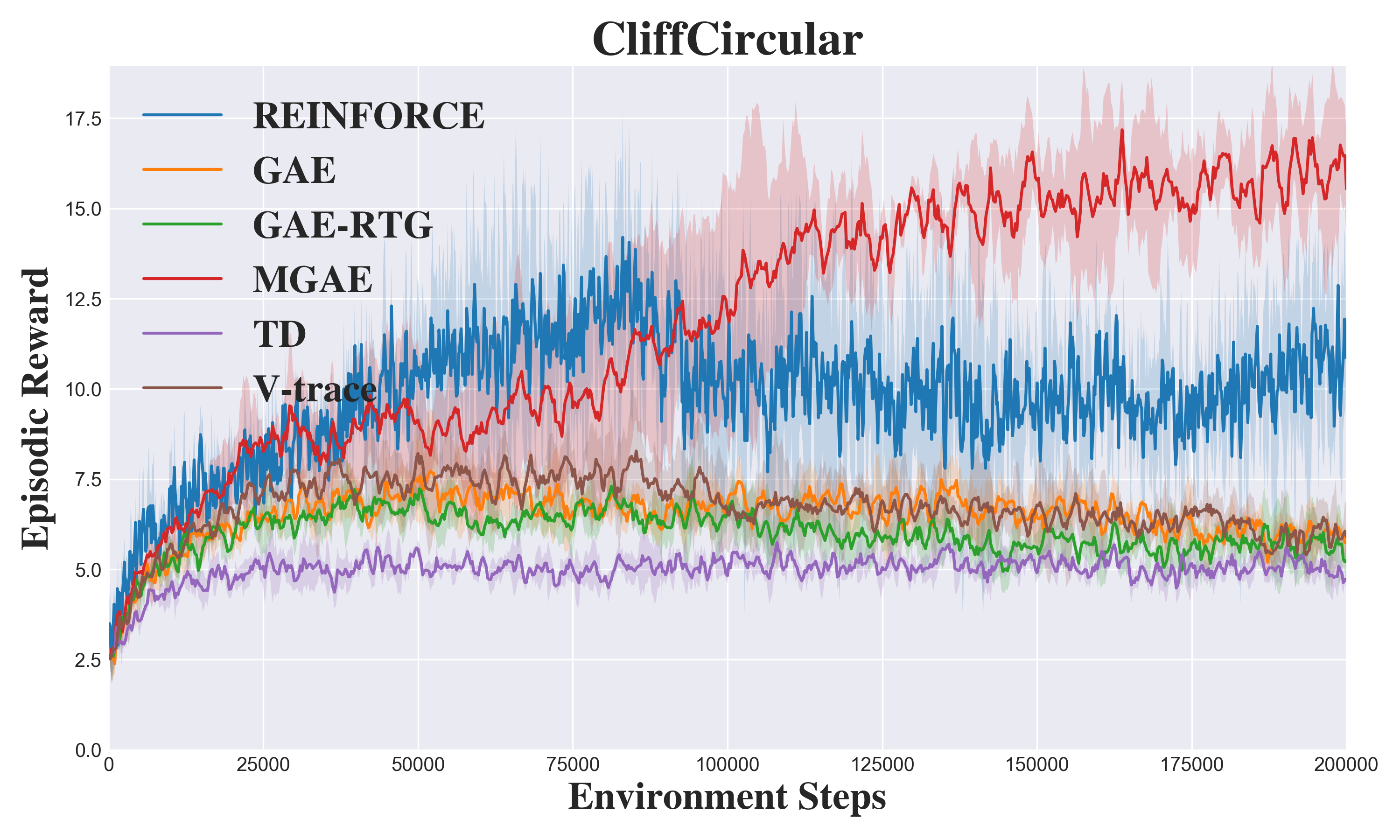}
    \includegraphics[width=0.49\linewidth]{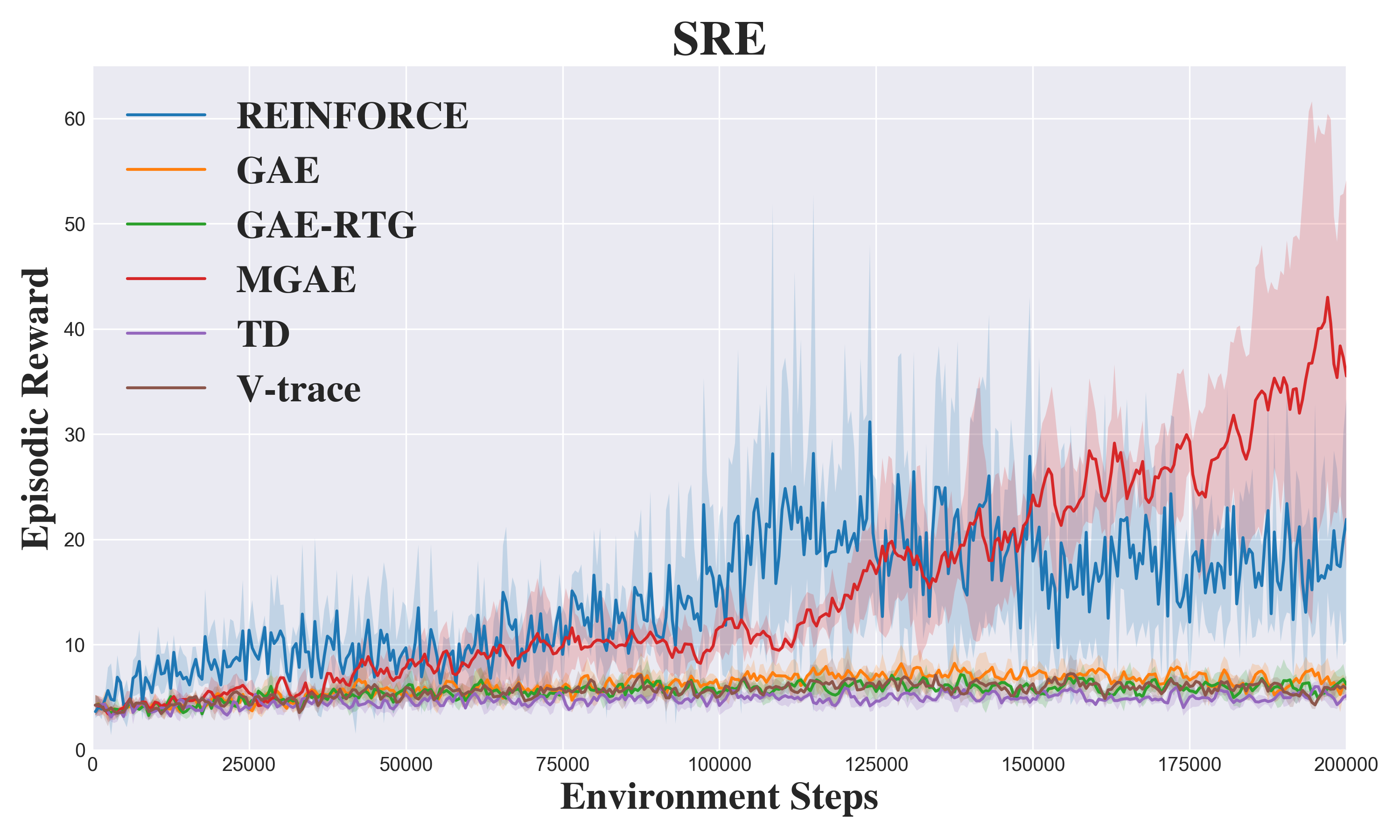}
    \caption{\revisenew{Episodic rewards during training in \textit{Medium} level of CliffCircular-v1 environment and Safe Riverine Environment for all reward advantage estimation methods with 3 seeds.}}
    \label{fig:adv-comp-training}
\end{figure}

To show the efficacy of the proposed marginal gain advantage estimation (MGAE) method with only reward signals, we first plot its learning curve of episodic rewards, as well as the curves of other reward advantage estimation methods, in the medium levels of the CliffCircular environment and Safe Riverine Environment, as shown in Figure \ref{fig:adv-comp-training}.
It can be observed that the proposed MGAE method's episodic reward increases gradually, and is consistently higher than the other methods, in both environments.
MGAE reaches a higher episodic reward at a faster rate compared to other methods, particularly in the CliffCircular environment. 
This accelerated convergence suggests that MGAE provides a more informative advantage signal for optimizing the policy in non-Markovian tasks, where reward dependencies extend over longer trajectories.
In the SRE environment, where the agent’s reward is sparse and depends on visiting new river segments, MGAE significantly outperforms state-wise methods like GAE and TD. This is likely due to MGAE’s trajectory-wise marginal gain calculation, which effectively captures long-term dependencies essential for accumulating rewards in sparse, history-dependent tasks.
While V-trace attempts to correct off-policy data by reweighting advantage estimates, it underperforms in both environments due to the sparse, non-Markovian nature of rewards. In contrast, GAE, which smooths TD errors across time steps, provides better stability but fails to achieve the same level of episodic reward as MGAE. 

\revisenew{
The most comparable method to MGAE is another Monte Carlo inspired one: REINFORCE with state value baseline, which also uses trajectory returns, but relies on bootstrapped value estimates as a baseline.
However, in a SMDP with non-Markovian rewards, a value function conditioned only on partial observation (common for egocentric vehicles) struggles to predict future rewards reliably.
MGAE instead builds its baseline from a sliding window average of past episodic returns and the cumulative estimated rewards looking back at the current trajectory.
This incorporates both the agent’s recent performance and its own reward estimator’s evolving predictions, yielding a more relevant reference point and further variance reduction.
}

%This discrepancy highlights the advantage of MGAE’s backward-looking cumulative approach in handling trajectory-level dependencies.

The superior performance of the MGAE over traditional reward advantage estimation methods in non-Markovian environments like CliffCircular and SRE can be attributed to several factors. 
First, MGAE’s trajectory-wise estimation approach inherently aligns with the non-Markovian reward structure, as it accounts for historical dependencies essential in tasks where rewards depend on visiting new, unvisited regions. 
In contrast, value-based methods such as GAE and TD struggle to capture these dependencies, often resulting in higher reward critic loss due to their reliance on an immediate state-value function, which is less suited to non-Markovian \revisenew{partially observable} settings. 
Second, MGAE provides more dynamic feedback to the policy by capturing prompt changes in reward across the trajectory, leading to sharper fluctuations in advantage estimation that enable more adaptive policy updates. 
Finally, the episode-wise policy update strategy used in training further enhances MGAE’s effectiveness, as it aligns with MGAE’s trajectory-based estimation while potentially limiting the granularity of state-wise adjustments available to traditional advantage estimators. 
Together, these factors underscore MGAE’s suitability for environments with complex reward dependencies, offering an approach that is both responsive to immediate changes and robust to sparse rewards.

\begin{figure}[htb]
    \centering
    \includegraphics[width=0.49\textwidth]{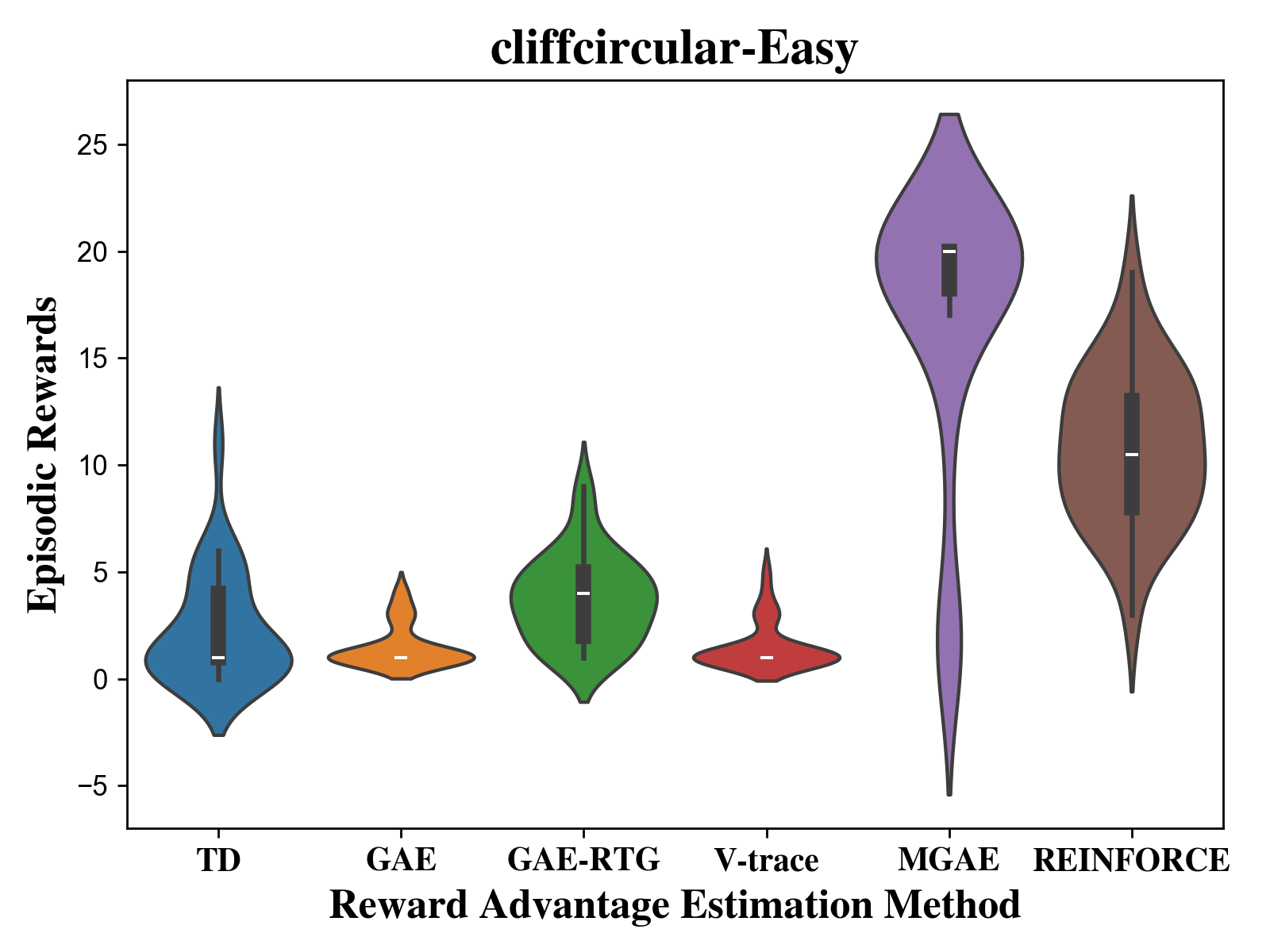}
    \includegraphics[width=0.49\textwidth]{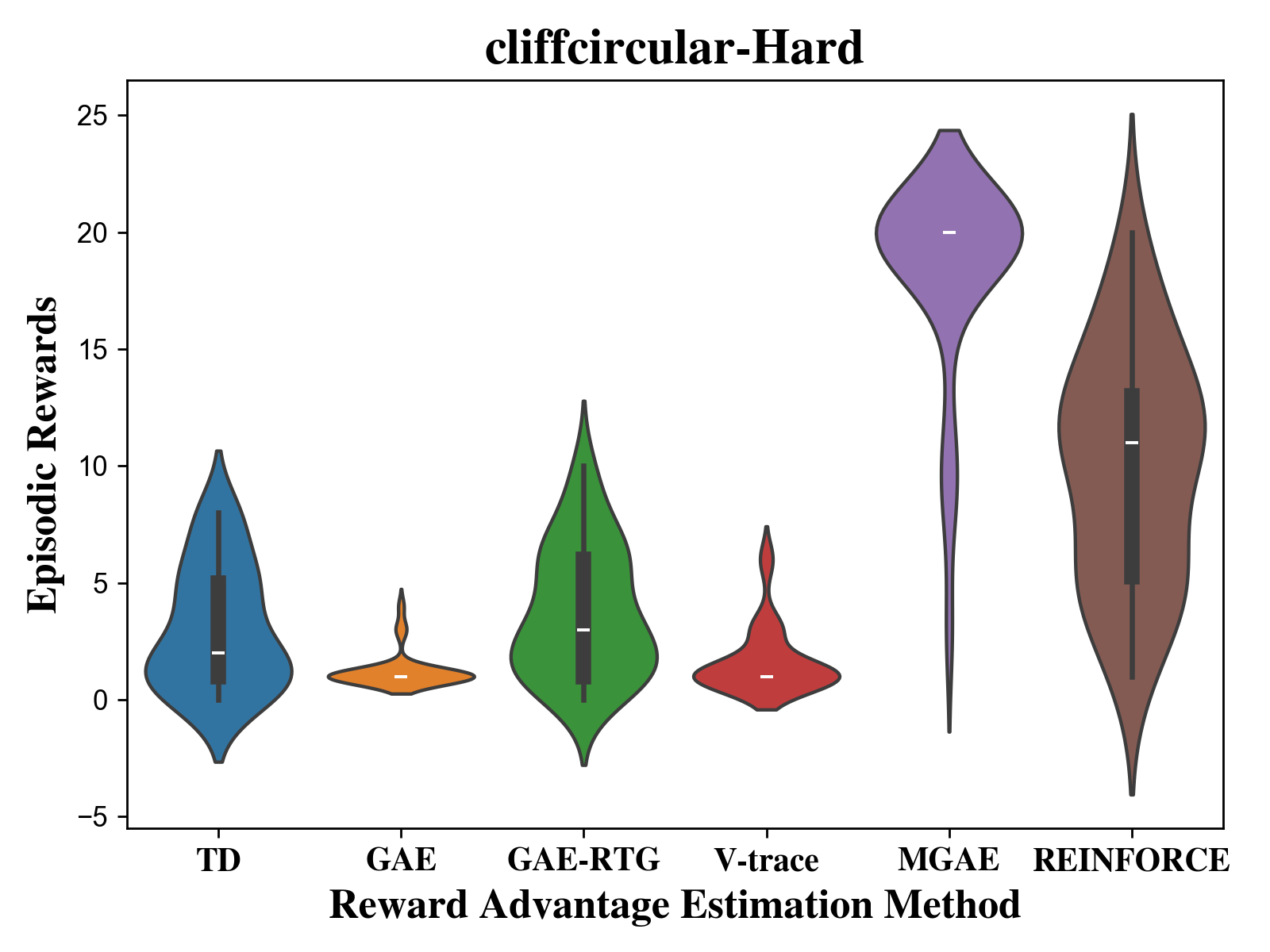}
    \includegraphics[width=0.49\textwidth]{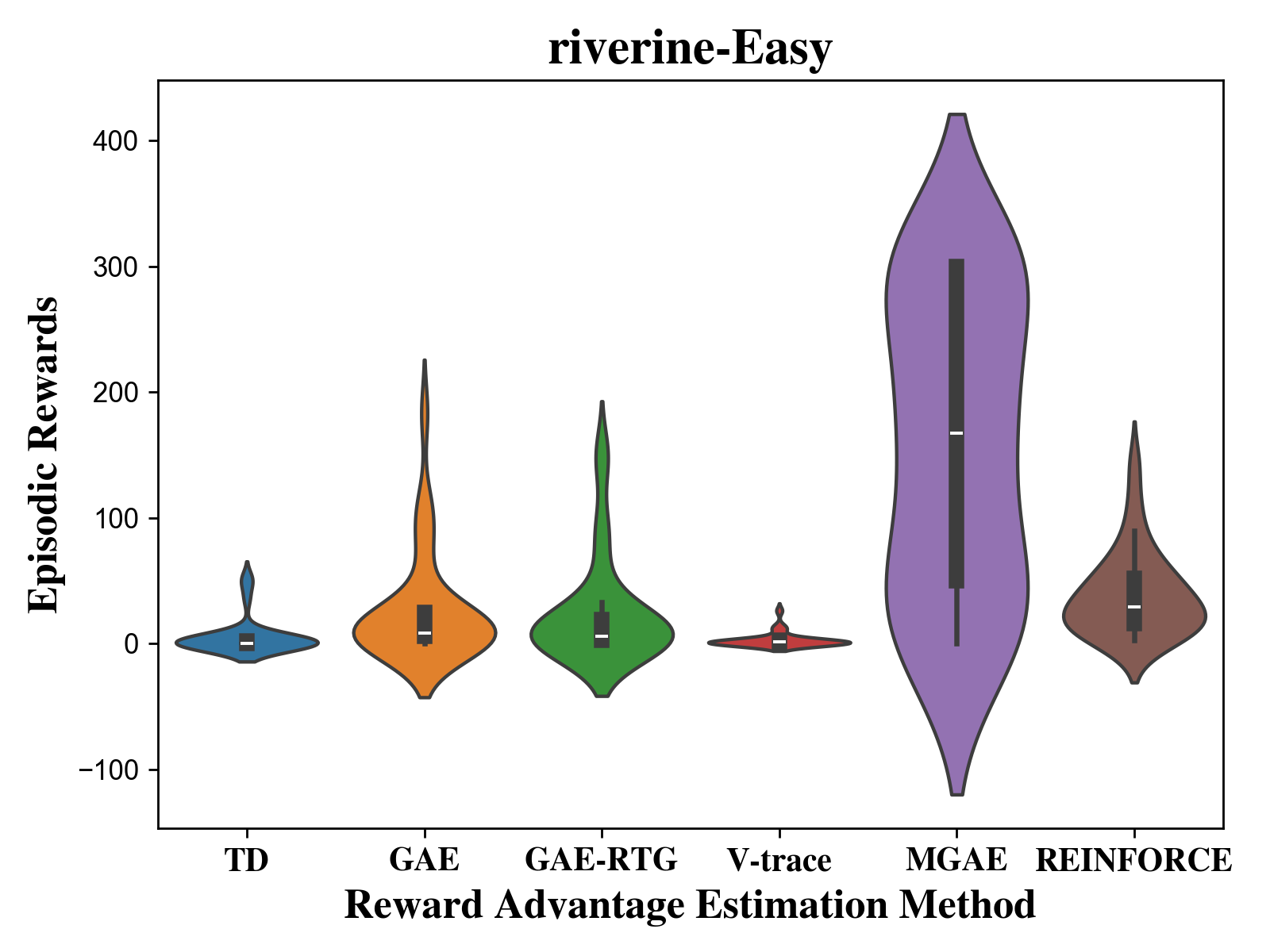}
    \includegraphics[width=0.49\textwidth]{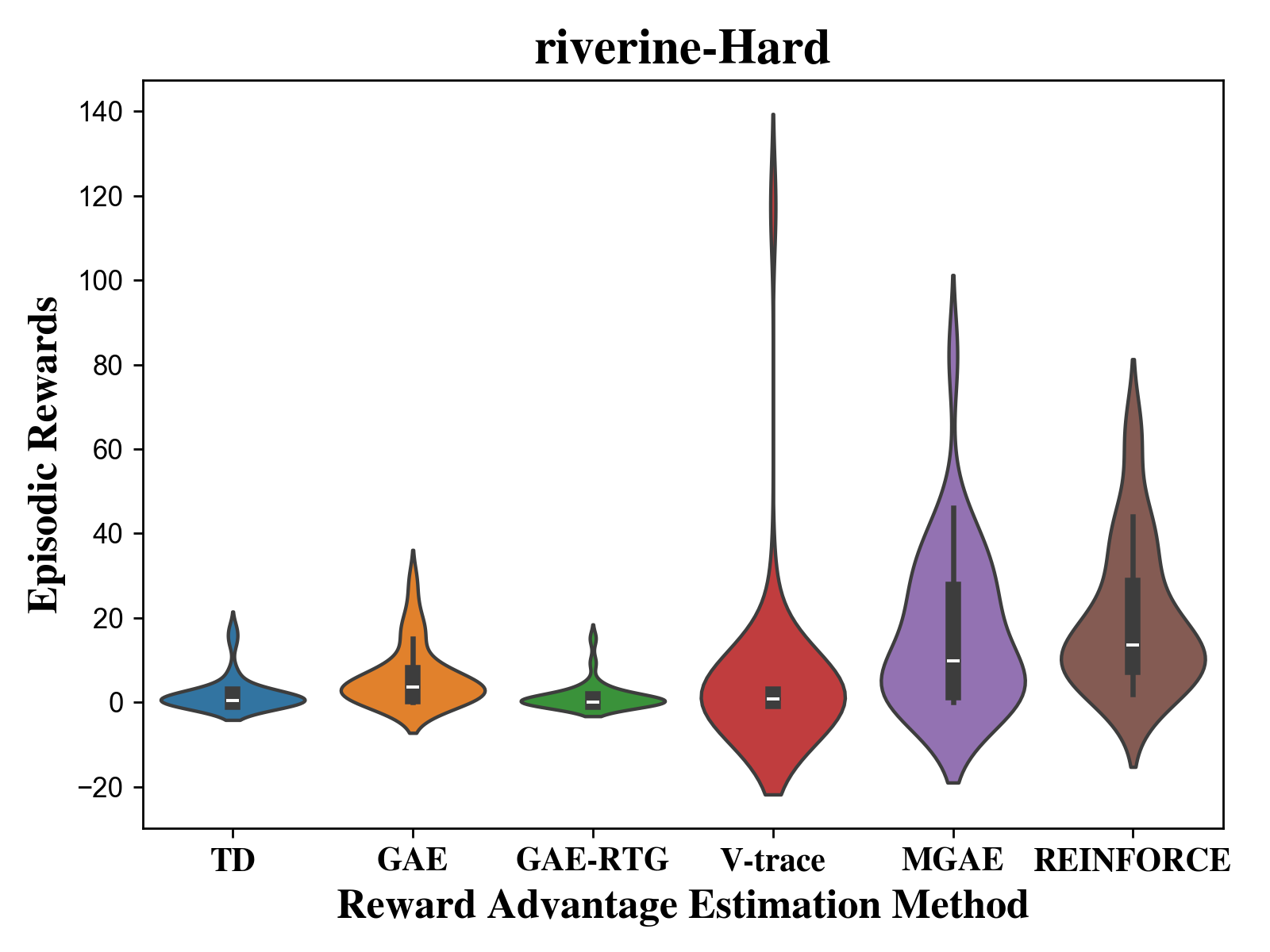}
    \caption{\revisenew{Evaluation distributions of episodic rewards of different reward advantage estimation methods for agents trained in the \textit{Medium}, evaluated in the \textit{Easy} and \textit{Hard} levels of the CliffCircular-v1 environment and the Safe Riverine Environment, each for 30 episodes.}}
    \label{fig:adv-comp-testing}
\end{figure}

To further evaluate the generalization abilities of different reward advantage estimation methods, agents trained in the \textit{Medium} level were tested in both \textit{Easy} and \textit{Hard} levels of the CliffCircular and Safe Riverine environments. 
The violin plots in Figure \ref{fig:adv-comp-testing} illustrate the distribution of episodic rewards across these conditions. 
MGAE consistently yields higher reward distributions, particularly in harder levels, suggesting its robustness in adapting to increased task complexity. While other methods, such as GAE, GAE-RTG, and TD, exhibit relatively low reward values. 
\revisenew{
The only exception is REINFORCE with state critic baseline, which has a tiny better mean evaluation performance in the \textit{Hard} level of SRE, but has shorter peak compared to MGAE.
}
MGAE maintains a higher median reward across episodes, indicating more effective learning of task-relevant behaviors. This advantage is likely due to MGAE’s trajectory-sensitive approach, which better captures historical dependencies essential for environments with submodular, non-Markovian rewards. 
Consequently, MGAE’s performance demonstrates that it not only accelerates learning but also enhances the agent’s ability to generalize effectively across varying task difficulties, leading to higher reward attainment even in previously unseen conditions.

\subsection{Semantic Dynamics Model}\label{sec:results-sdm}
%In the context of model-based value expansion (MVE, \cite{ebert2018visual}), single-step model rollouts are surprisingly competitive in policy learning \cite{janner2019trust}.

\begin{figure}[htb]
    \centering
    \includegraphics[width=0.9\textwidth]{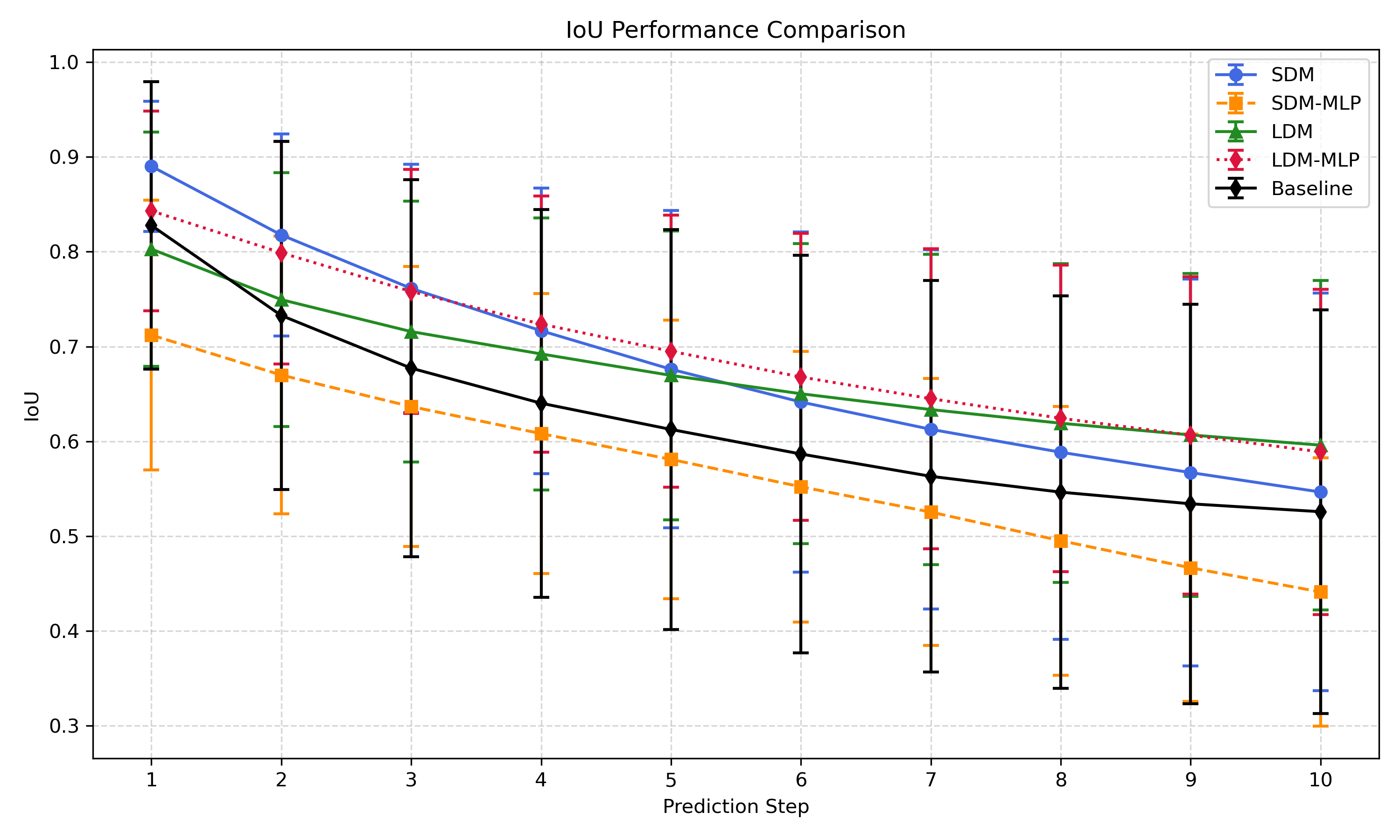}
    \caption{Comparison of vision dynamics models in terms of IoU, for 10-step prediction.}
    \label{fig:sdm-comp-iou}
\end{figure}

\begin{figure}[htb]
    \centering
    \includegraphics[width=0.9\textwidth]{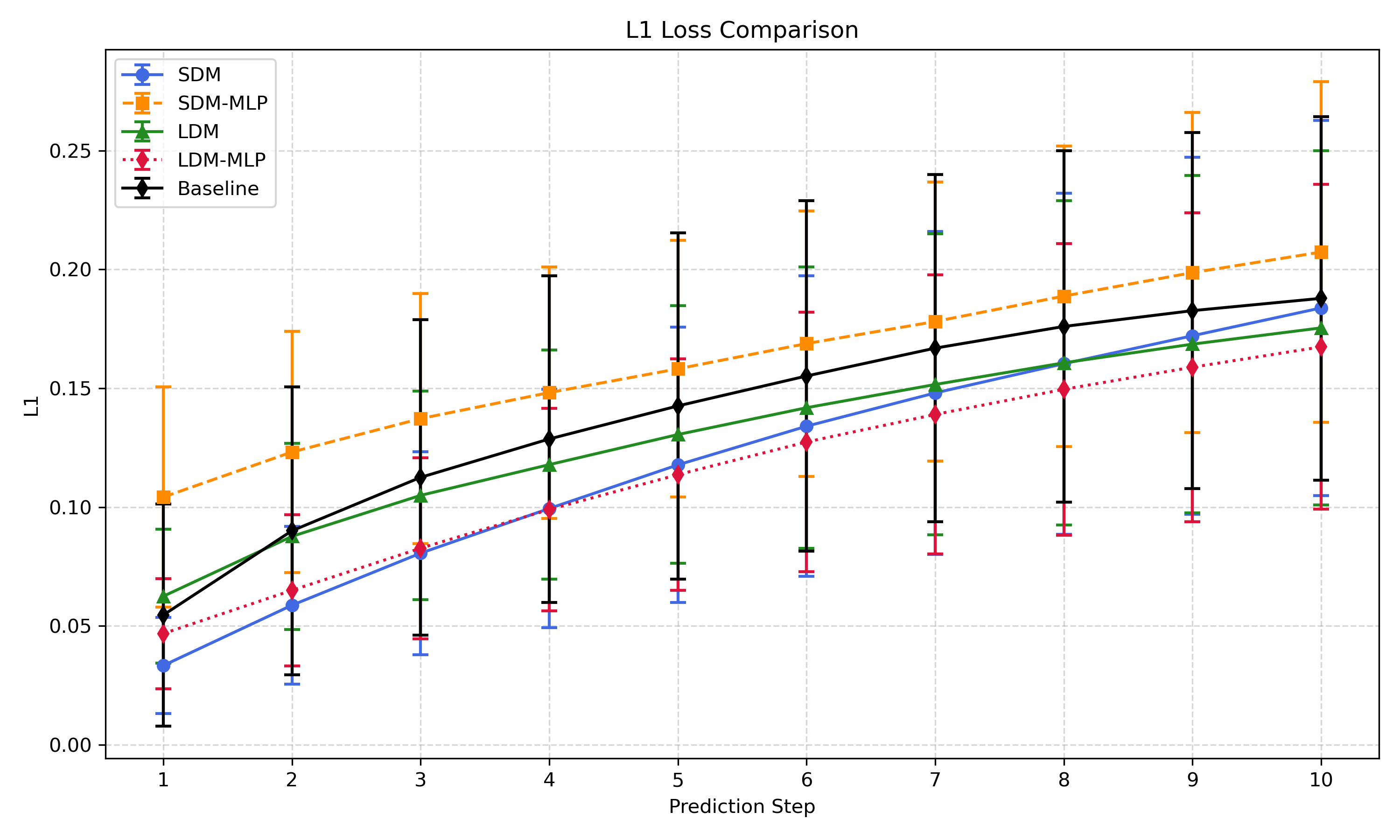}
    \caption{Comparison of vision dynamics models in terms of L1 loss for 10-step prediction.}
    \label{fig:sdm-comp-l1}
\end{figure}

The performance of the proposed Semantic Dynamics Model (SDM) was evaluated against alternative vision dynamics models, including SDM-MLP, LDM, LDM-MLP, and a baseline method, using L1 loss and Intersection over Union (IoU) metrics over a 10-step prediction horizon in the Safe Riverine Environment (Figure \ref{fig:sdm-comp-iou} and \ref{fig:sdm-comp-l1}).
\revise{
The predictive accuracy of SDM degrades as the planning horizon increases, reaching baseline performance at approximately 10 steps. 
Beyond this point, the SDM's predictions provide no additional benefit over simply assuming the current observation persists, indicating that accumulated errors from homography transformations and large localization shifts render long-horizon predictions unreliable. 
This empirically determined horizon aligns with the needs of short-term planning in Safe RL, where immediate future observations are critical for avoiding obstacles and maintaining trajectory alignment.
}

The results demonstrate that SDM consistently achieves the highest prediction accuracy in short-term predictions, as reflected by its lower L1 loss and higher IoU compared to other models. 
Notably, the performance gap between SDM and latent dynamics models (LDM and LDM-MLP) is most pronounced within the first three prediction steps, highlighting SDM's ability to capture precise spatial transformations in the immediate future. 
This is particularly critical for safety in reinforcement learning tasks, as accurate short-term predictions directly influence the agent’s ability to foresee and avoid hazardous states during navigation. 
Unlike LDM-based methods, which rely on latent vector representations that may obscure critical geometric details, SDM explicitly models spatial dynamics using homography constraints, enabling it to provide geometrically consistent predictions. 
Moreover, the SDM-MLP, which lacks these geometric constraints, underperforms compared to SDM and even the baseline, further underscoring the importance of incorporating structured priors. 
While all models experience performance degradation as the prediction horizon increases, SDM's superior short-term accuracy ensures it is better suited for real-time safety-critical decision-making as part of the Constrained Actor Dynamics Estimator architecture.

\begin{figure*}[h]
    \centering
    \includegraphics[width=\textwidth]{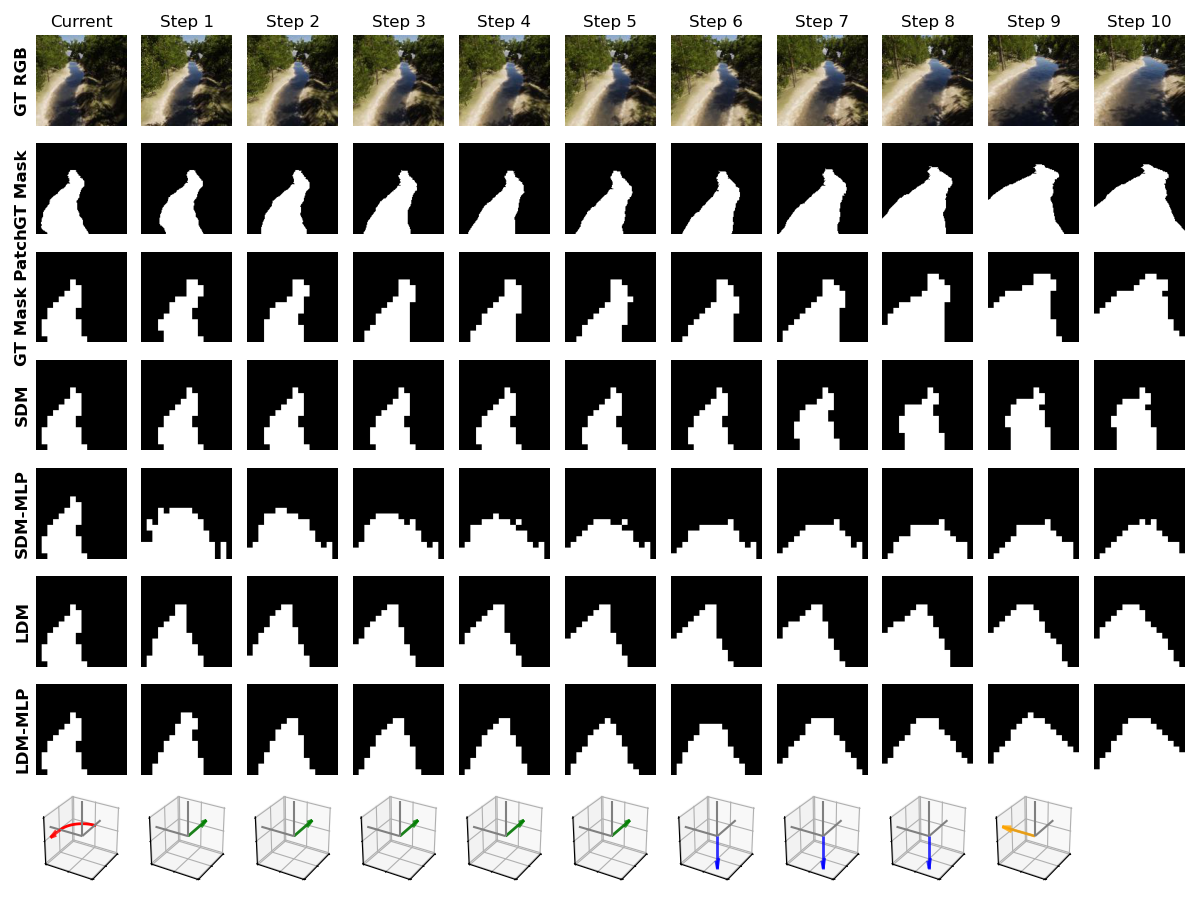}
    \caption{Qualitative comparison of different vision dynamics models for 10 step predictions of future (patchified) observations. The last row represents the actual action taken at each prediction step. Red arrow represents horizontal rotation, green arrow is longitudinal translation, orange arrow is latitudinal translation and blue arrow is vertical translation. Best viewed in color.}
    \label{fig:dynamics-comp}
\end{figure*}

Figure \ref{fig:dynamics-comp} provides a qualitative comparison of the predicted observations from different vision dynamics models over a 10-step prediction horizon, juxtaposed against the ground truth RGB and patchified water semantic masks. 
SDM demonstrates superior consistency and alignment with the ground truth, especially in the early prediction steps, where its patchified masks closely resemble the semantic structure of the environment. 
This can be attributed to SDM's explicit incorporation of geometric constraints through homography, enabling accurate spatial transformations.
Conversely, SDM-MLP, which lacks these geometric priors, exhibits noticeable drift and degradation in mask fidelity, even in short-term predictions. 
The latent dynamics models (LDM and LDM-MLP) maintain relatively smooth predictions but often fail to preserve fine semantic details, as the latent embeddings abstract away critical spatial features. 
Over longer prediction horizons, all models experience varying degrees of degradation, but SDM consistently retains more structural integrity compared to others. 
The final row, depicting the actions taken at each prediction step, further illustrates how SDM's predictions align with the agent's executed actions, ensuring spatial coherence between the predicted and real observations. 
This qualitative analysis reinforces SDM's efficacy in providing interpretable and geometrically consistent predictions, crucial for vision-driven safety-critical tasks.

\subsection{Constrained Actor Dynamics Estimator}\label{sec:results-cade}
The proposed Constrained Actor Dynamics Estimator (CADE) structure stands as a model-based safe reinforcement learning framework for partially observable Markov Decision Process with non-Markovian reward and Markovian cost, which is a subset of the CSMDP. 

We compare the episodic rewards and episodic costs of two methods of safe learning: Lagrangian-based advantage estimation and safety layer with cost-based planning. The baseline is the MGAE method based on rewards without any cost feedback.
All algorithms are trained in the medium level of CliffCircular-v1 and SRE, and evaluated in all three difficulty levels to test the generalizability of each method.

% Figure of training in medium CliffCircular-v1
\begin{figure}[h]
    \centering
    \includegraphics[width=0.49\textwidth]{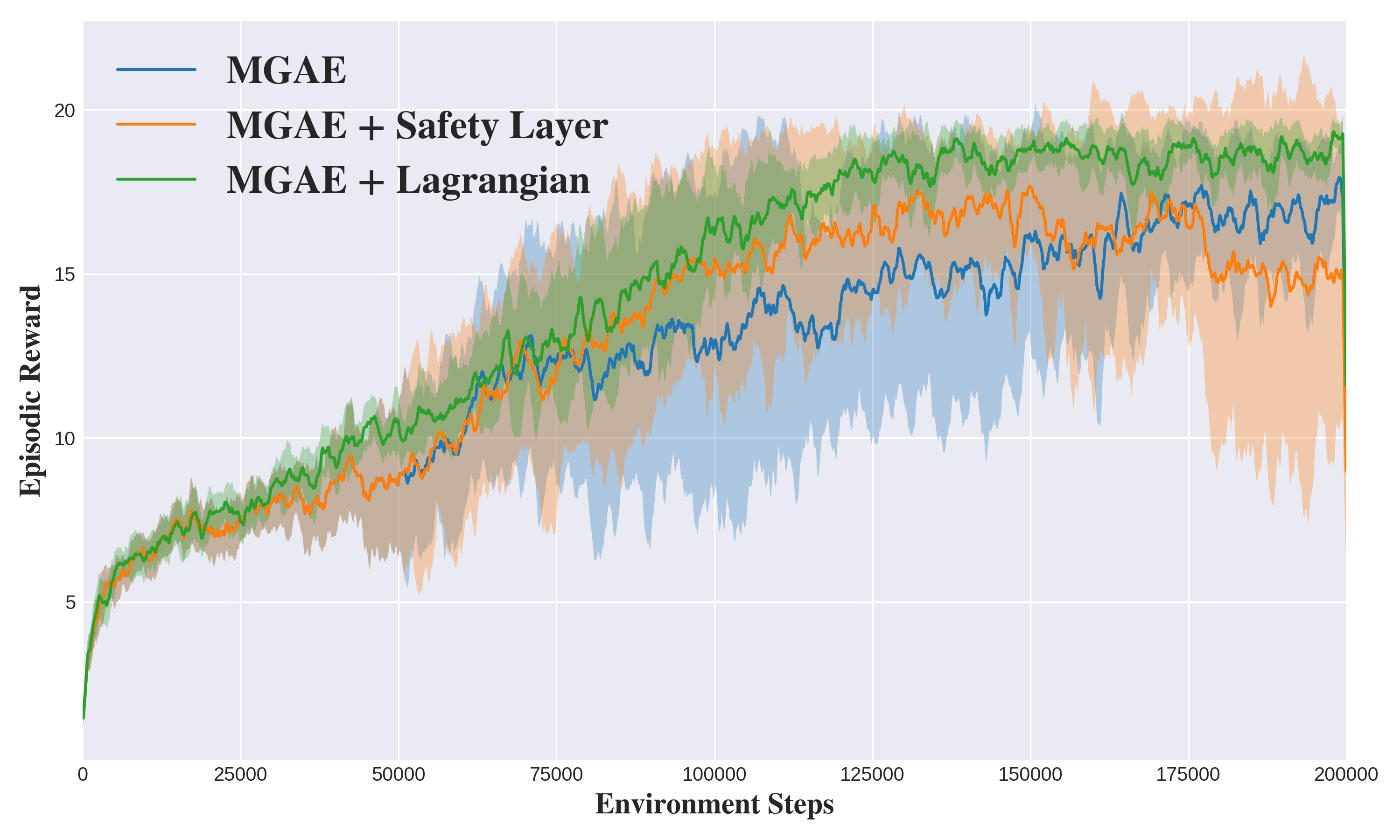}
    \includegraphics[width=0.49\textwidth]{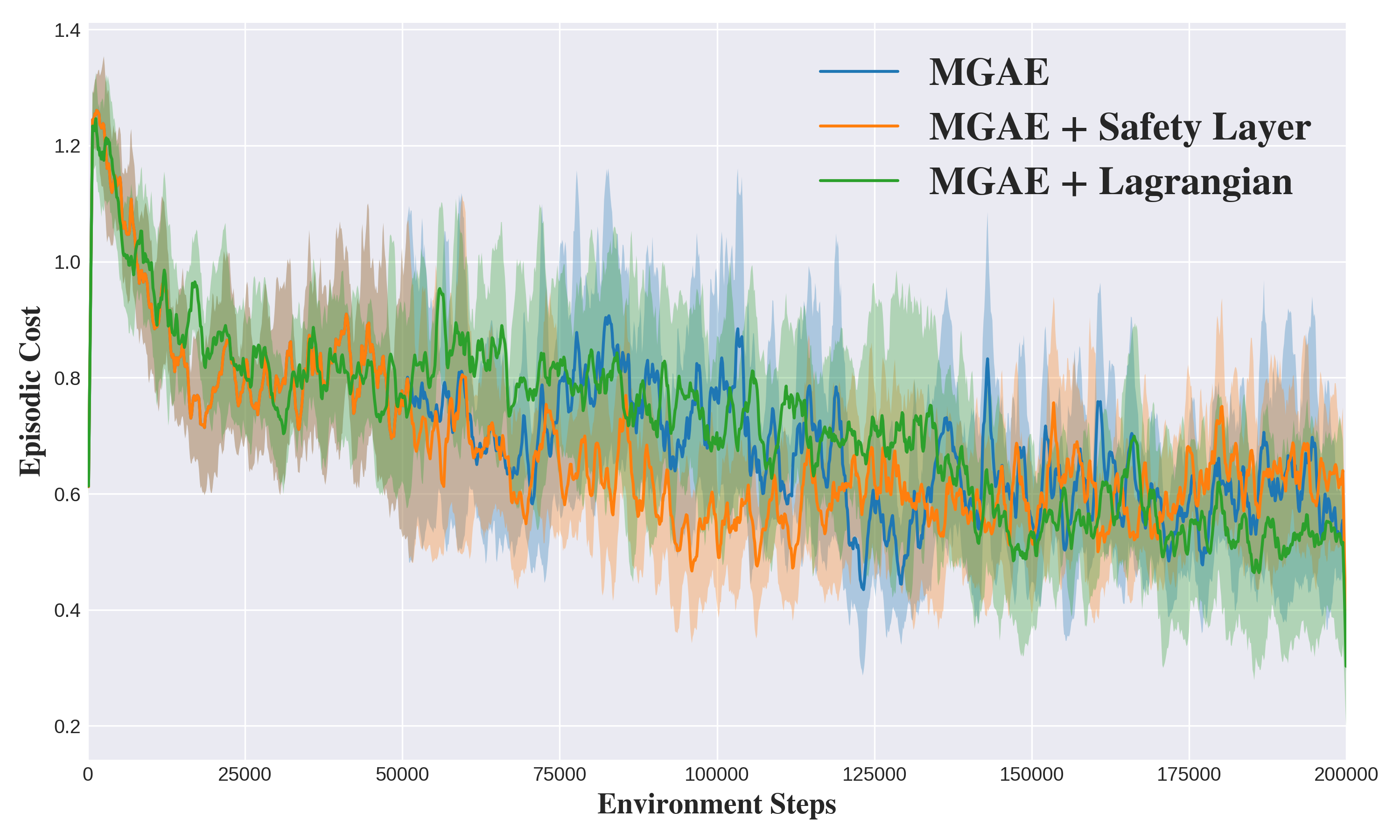}
    \caption{Comparison of safety regulation methods in terms of episodic reward and episodic cost during training in the medium level of CliffCircurlar-v1 environment.}
    \label{fig:cade-cliffcircular-training-comp}
\end{figure}

% Figure of training in medium SRE
\begin{figure}[h]
    \centering
    \includegraphics[width=0.49\textwidth]{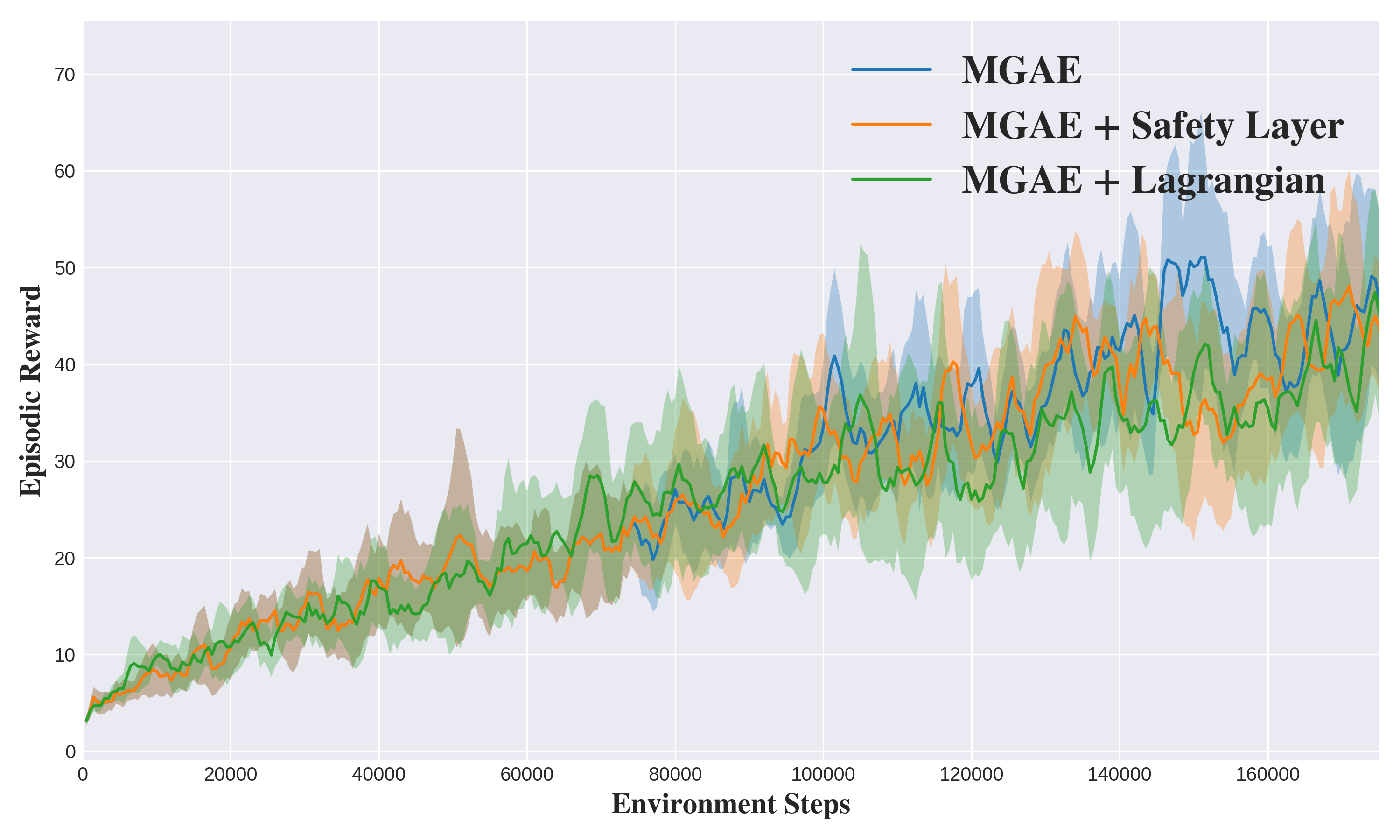}
    \includegraphics[width=0.49\textwidth]{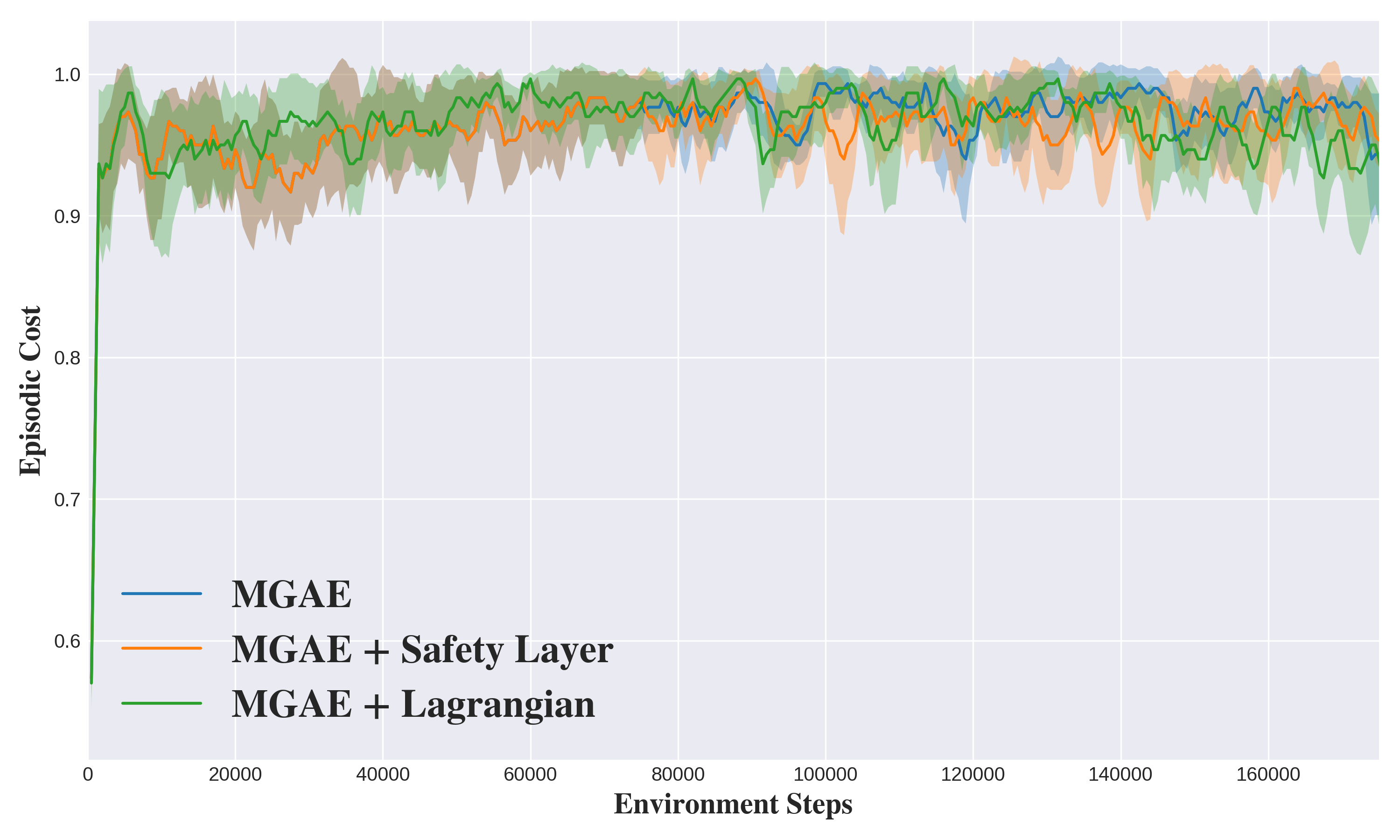}
    \caption{Comparison of safety regulation methods in terms of episodic reward and episodic cost during training in the medium level of Safe Riverine Environment.}
    \label{fig:cade-sre-training-comp}
\end{figure}

\begin{table}[t]
\centering
\caption{\revisefinal{Evaluation results of safety regulation methods in all three difficulty levels of the \textbf{CliffCircular-v1} environment, with all methods trained only in the medium level. Each method is evaluated over 30 runs per seed, using 3 different seeds, for a total of 90 evaluations per method per difficulty level. EpR means episodic reward, EpC means episodic cost. Best values are marked as bold.}}
\label{tab:cliffcircular_eval_cade}
\footnotesize
\begin{tabular}{lccc}
\hline
 & MGAE & MGAE+Lagrangian & MGAE+SafetyLayer \\
\hline
Easy--EpR   & 17.40 $\pm$ 4.43 & \textbf{19.76 $\pm$ 1.12} & 15.20 $\pm$ 7.36 \\
Easy--EpC   & 0.42 $\pm$ 0.46  & \textbf{0.33 $\pm$ 0.26}  & 0.53 $\pm$ 0.53  \\ \hline
Medium--EpR & 16.30 $\pm$ 5.87 & \textbf{19.24 $\pm$ 2.42} & 14.68 $\pm$ 7.44 \\
Medium--EpC & 0.65 $\pm$ 0.60  & \textbf{0.42 $\pm$ 0.30}  & 0.72 $\pm$ 0.61  \\ \hline
Hard--EpR   & 15.80 $\pm$ 6.12 & \textbf{18.86 $\pm$ 2.66} & 14.66 $\pm$ 6.85 \\
Hard--EpC   & 0.75 $\pm$ 0.63  & \textbf{0.60 $\pm$ 0.33}  & 0.70 $\pm$ 0.54  \\
\hline
\end{tabular}
\end{table}

\begin{table}[t]
\centering
\caption{
\revisefinal{Evaluation results of safety regulation methods in all three difficulty levels of the \textbf{Safe Riverine Environment}, with all methods trained only in the medium level. Each method is evaluated over 30 runs per seed, using 3 different seeds, for a total of 90 evaluations per method per difficulty level. EpR means episodic reward, EpC means episodic cost. Best values are marked as bold.}
}
\label{tab:sre_eval_cade}
\footnotesize
\begin{tabular}{lccc}
\hline
 & \textbf{MGAE} & MGAE+Lagrangian & MGAE+SafetyLayer \\
\hline
Easy--EpR  & \(\mathbf{85.62 \pm 71.48}\) & \(36.86 \pm 44.66\) & \(70.00 \pm 58.09\) \\
Easy--EpC  & \(\mathbf{0.95 \pm 0.18}\)   & \(1.00 \pm 0.00\) & \(0.98 \pm 0.09\) \\ \hline
Medium--EpR& \(\mathbf{48.26 \pm 38.10}\) & \(41.15 \pm 33.05\) & \(43.07 \pm 33.30\) \\
Medium--EpC& \(0.98 \pm 0.10\)   & \(0.99 \pm 0.07\)   & \(\mathbf{0.96 \pm 0.14}\) \\ \hline
Hard--EpR  & \(24.75 \pm 26.35\) & \(24.14 \pm 30.94\) & \(\mathbf{33.67 \pm 35.03}\) \\
Hard--EpC  & \(\mathbf{0.91 \pm 0.19}\) & \(0.95 \pm 0.14\) & \(0.93 \pm 0.18\) \\
\hline
\end{tabular}
\end{table}

Figure \ref{fig:cade-cliffcircular-training-comp} illustrates the training curves in the CliffCircular-v1 environment. 
Here, the MGAE+Lagrangian method consistently achieves higher episodic rewards compared to plain MGAE and MGAE+SafetyLayer. 
Notably, the episodic reward curve for MGAE+Lagrangian not only reaches higher average values but also exhibits significantly lower variance. 
The episodic cost trajectory similarly indicates that MGAE+Lagrangian exhibits safer behavior at the end of training, with lower average costs. 
These results suggest that Lagrangian-based integration effectively incorporates safety constraints into the policy optimization process, guiding the agent toward both high performance and safety.
However, the training results in SRE do not show significant differences in episodic rewards or costs between the three methods, Figure \ref{fig:cade-sre-training-comp}. 
To investigate the causes of this discrepancy and evaluate how well the safety regulation methods generalize and perform in both simpler and more challenging settings, we evaluate the trained models in all three difficulty levels of both environments.

% Evaluation Results in CliffCircular and SRE
Table \ref{tab:cliffcircular_eval_cade} summarizes the evaluation metrics for the CliffCircular environment across three difficulty levels. 
MGAE+Lagrangian exhibits the highest episodic rewards and the lowest episodic costs consistently, with, for example, mean episodic rewards around 19.7 and costs around 0.33 for the Easy level. 
In contrast, MGAE+SafetyLayer lags behind in both metrics. 
Conversely, Table \ref{tab:sre_eval_cade} shows the evaluation results for SRE: while plain MGAE outperforms on the Easy and Medium levels, MGAE+SafetyLayer delivers the best performance on the Hard level in terms of episodic reward (e.g., 33.67 versus 24.14 for MGAE+Lagrangian) and maintains comparable costs. 
%These differences suggest that the effectiveness of safety regulation strategies is environment-dependent, likely due to the inherent differences in the cost signal and action space structure between CliffCircular and SRE.
\revise{
The cost-planning safety layer exhibits performance degradation in basically all levels of two environments, which potentially leads to overly conservative behavior, limiting agent's exploration during training. }

The differing results between the two environments could be attributed to several interacting factors. 
Unlike CliffCircular-v1, in SRE, the cost signal is provided only at terminal states (with severe or light violations) and is not directly tied to the patchified semantic observation, Table \ref{tab:env_comparison}. 
As a result, the cost estimator has difficulty converging to accurate predictions, especially when similar observations—occurring several steps before and after a violation—are associated with different cost values. 
Consequently, the 1-step cost advantage estimates in SRE are generally very small (mostly less than 0.2), which diminishes the influence of cost-based corrections in the MGAE+Lagrangian method, even degrades its performance compared to MGAE method. 
In contrast, the CliffCircular-v1 environment uses a cost function directly derived from the geometric layout of the semantic (cliff or non-cliff) mask, leading to more informative cost advantages (with several near 1 during training) that effectively guide policy updates.

Additionally, the performance differences between the safety layer and Lagrangian methods may stem from how each method utilizes the cost estimator. 
In the safety layer approach, the relative differences in cost estimates might matter more, allowing it to override actions when the immediate cost is relatively high. Conversely, the Lagrangian method depends on the absolute values of the cost estimator for shaping the entire policy optimization. 
On the other hand, in the SRE, where the action space is higher-dimensional and the policy entropy remains relatively high during training, the planner can more effectively sample safe actions, which may explain why MGAE+SafetyLayer achieves better performance in the hard level and even outperforms MGAE in episodic reward. 
Conversely, in the easier CliffCircular-v1 environment, the policy converges faster, reducing entropy and thereby diminishing the corrective opportunity for the safety layer. 
These observations suggest that both the Lagrangian method and safety layer with cost planning could play important roles in safety regulation during training, provided that the cost estimator is well-learned and a better trajectory sampling strategy is employed.

% Scatter Plot Analysis with safety layer Enabled
\revise{Since the safety layer with cost planning can be decoupled from the learning of the actor, to} investigate how it could help safety regulation during the inference or deployment phase, we evaluated all three methods with the safety layer enabled.
This means that CADE components like the SDM and the cost estimator can still be active post-training to regulate actions selected by the actor component.
% Figure of enabling safety layer in CliffCircular env
\begin{figure}[h]
    \centering
    \includegraphics[width=0.98\textwidth]{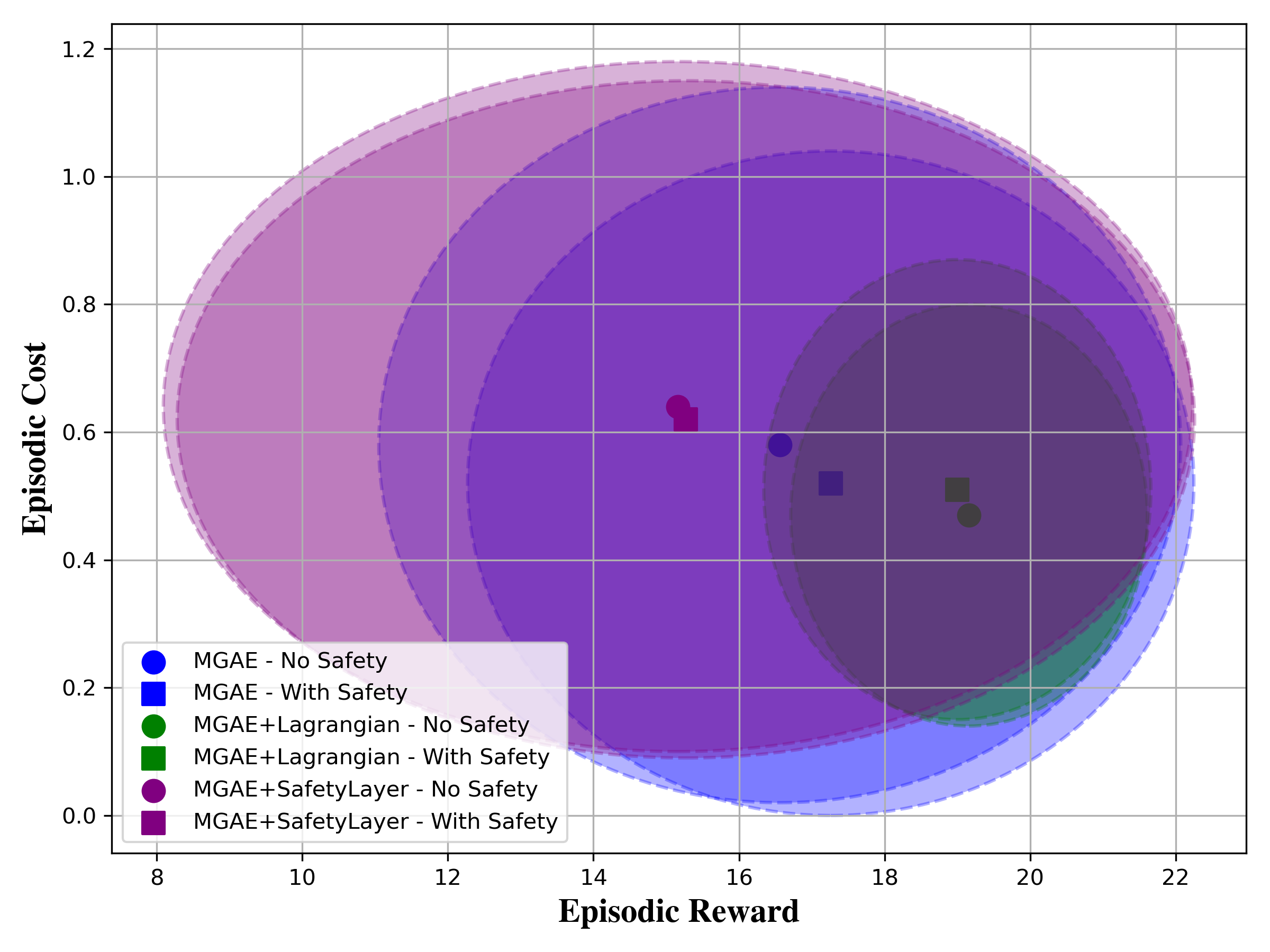}
    \caption{Comparison of episodic reward and cost during evaluation for MGAE, MGAE+Lagrangian, and MGAE+SafetyLayer, with and without the safety layer, in \textbf{CliffCircular-v1} environment. Each marker represents method's mean reward and cost, while ellipses indicate ±1 standard deviation.}
    \label{fig:cliffcircular-safety-layer-eval}
\end{figure}
% Figure of enabling safety layer in SRE
\begin{figure}[h]
    \centering
    \includegraphics[width=0.98\textwidth]{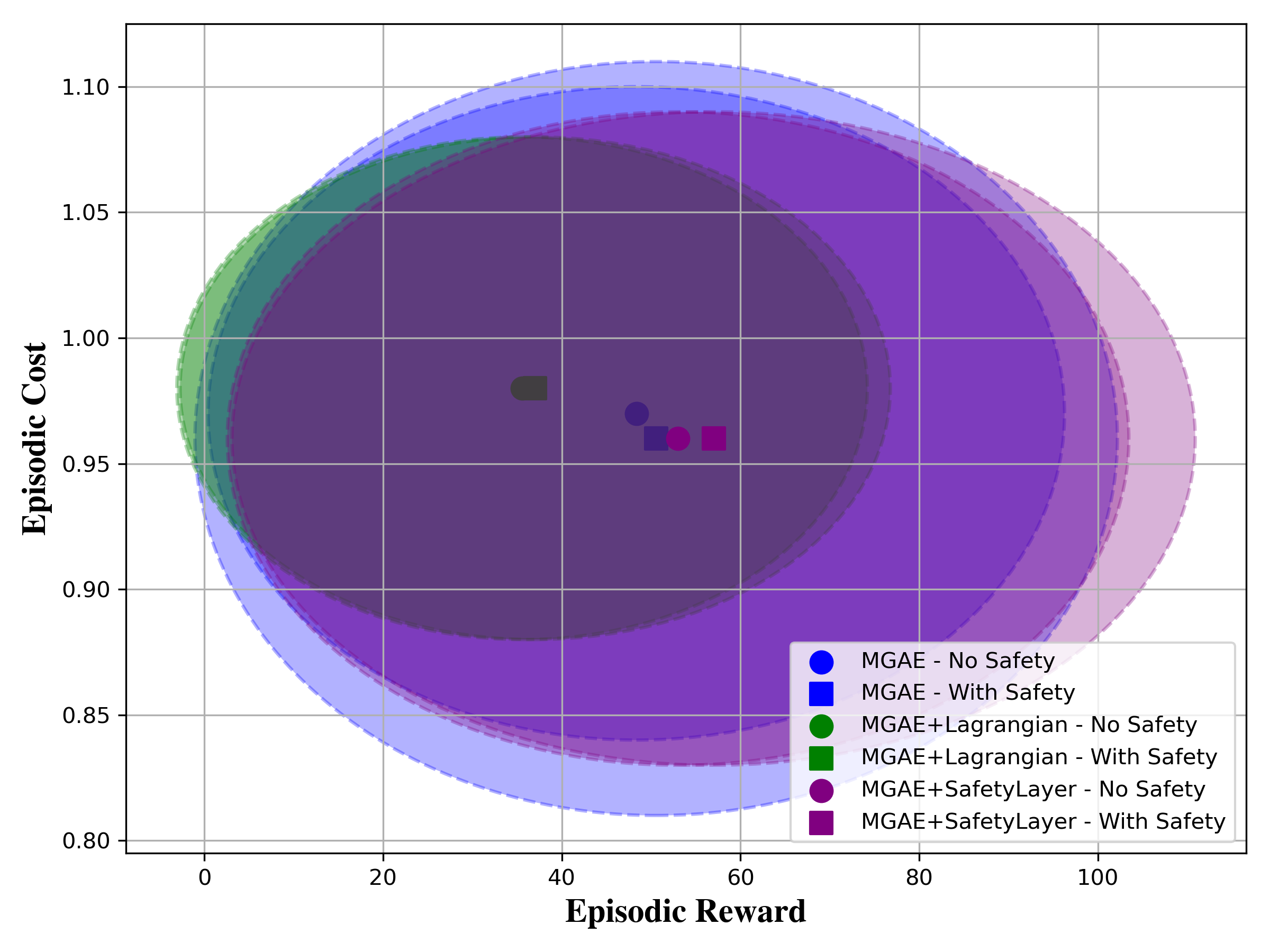}
    \caption{Comparison of episodic reward and cost during evaluation for MGAE, MGAE+Lagrangian, and MGAE+SafetyLayer, with and without the safety layer, in \textbf{Safe Riverine Environment}. Each marker represents method's mean reward and cost, while ellipses indicate ±1 standard deviation.}
    \label{fig:sre-safety-layer-eval}
\end{figure}
The scatter plots in Figures \ref{fig:cliffcircular-safety-layer-eval} and \ref{fig:sre-safety-layer-eval} compare the evaluation results with and without the safety layer enabled.
In the CliffCircular environment, safety activation increased episodic rewards and reduced episodic costs for MGAE and also provided a modest benefit for MGAE+SafetyLayer, while MGAE+Lagrangian experienced a slight performance degradation. 
In contrast, in the SRE, enabling safety layer resulted in higher episodic rewards for all three methods.
This improvement indicates that the safety layer is effective at refining action selection, as it helps the policy avoid actions that would otherwise incur high terminal costs. 
The consistent boost in performance suggests that the safety layer’s corrective intervention is particularly beneficial in more complex SRE, where the agent may not have learned safety constraints as effectively in the CliffCircular-v1 environment, and can still benefit from safety layer's corrective feedback.
%In the CliffCircular-v1 environment, the scatter of episodic rewards and costs is notably tighter for MGAE+Lagrangian, indicating both high performance and consistent safety regulation. 
%In SRE, however, while MGAE+Lagrangian still shows low variance, the overall performance is more mixed—plain MGAE and MGAE+SafetyLayer sometimes achieve higher rewards, particularly in harder scenarios.
%This could be attributed to the fact that, in SRE, the cost estimator is less accurate given the sparse and terminal nature of cost signals, and the relatively higher action space dimensionality allows the safety layer more opportunity to sample safe actions. 
%Moreover, the relative impact of the cost estimator appears to play a larger role in the safety layer method when compared to the Lagrangian method, which depends on the absolute cost values.

% Summary
The experimental results of CADE for safety regulation in partially observable CSMDPs lead to several conclusions. 
First, the Lagrangian-based safety regulation via dynamic soft balance between reward advantage and cost advantage requires an accurate cost estimator for Markovian cost.
Second, the direct action override at execution in training phase by safety layer with cost planning could harm the exploration ability of the agent. Thus, the trajectory sampling strategy that attends to current policy entropy, and the trajectory ranking mechanism that considers both estimated reward and cost could be further investigated.
Third, applying a safety layer during inference universally improves episodic rewards across all safety regulation methods, underscoring the importance of SDM and cost estimator in CADE.

\begin{figure}[h]
    \centering
    \includegraphics[width=\textwidth]{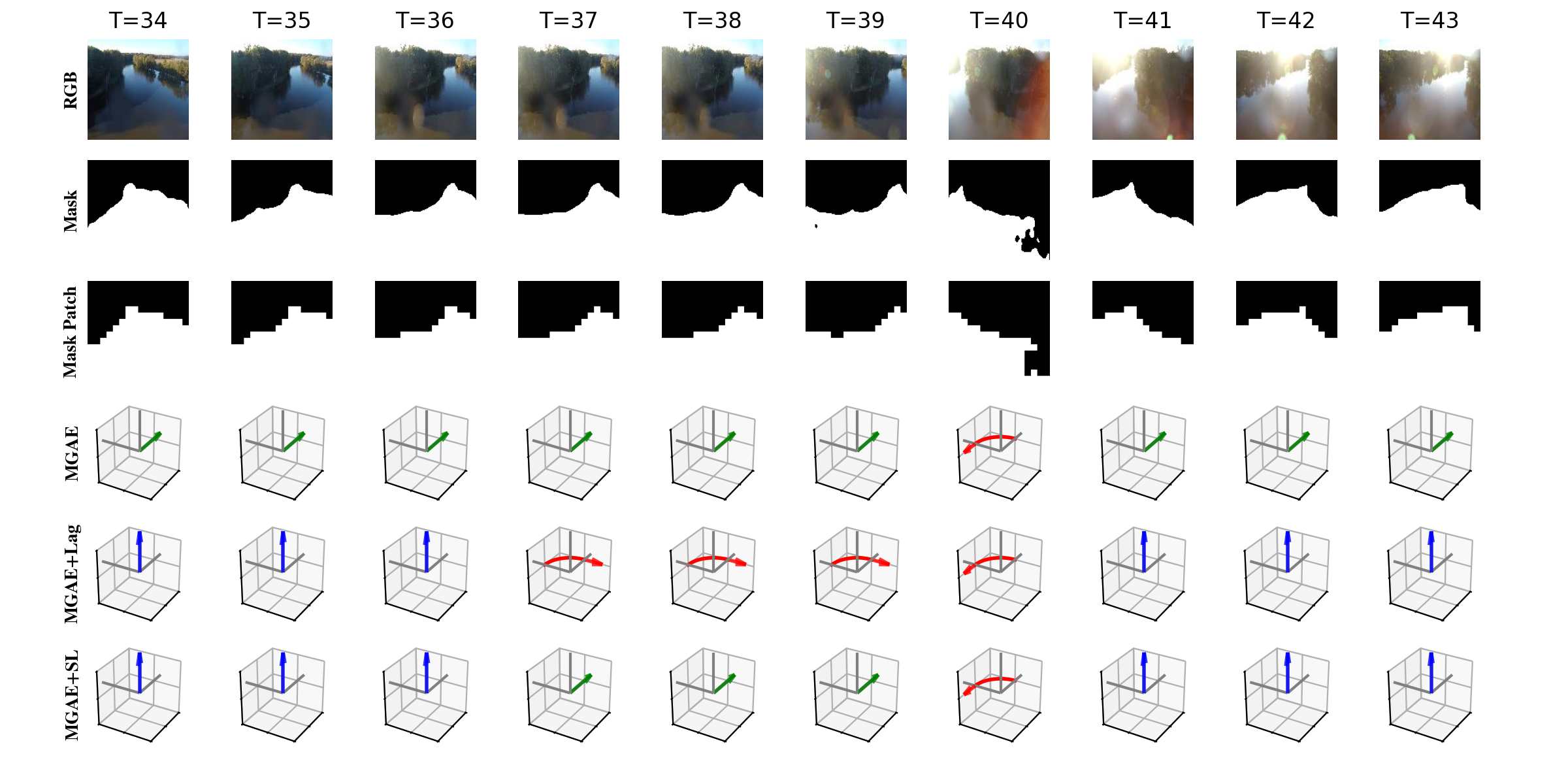}
    \caption{\revisenew{Open-loop policy evaluation of MGAE, MGAE+Lagrangian, and MGAE+SafetyLayer on \textbf{Wabash River}. T is the frame index in the video of 2 frames per second.}}
    \label{fig:model-eval-wabash}
\end{figure}

\begin{figure}[h]
    \centering
    \includegraphics[width=\textwidth]{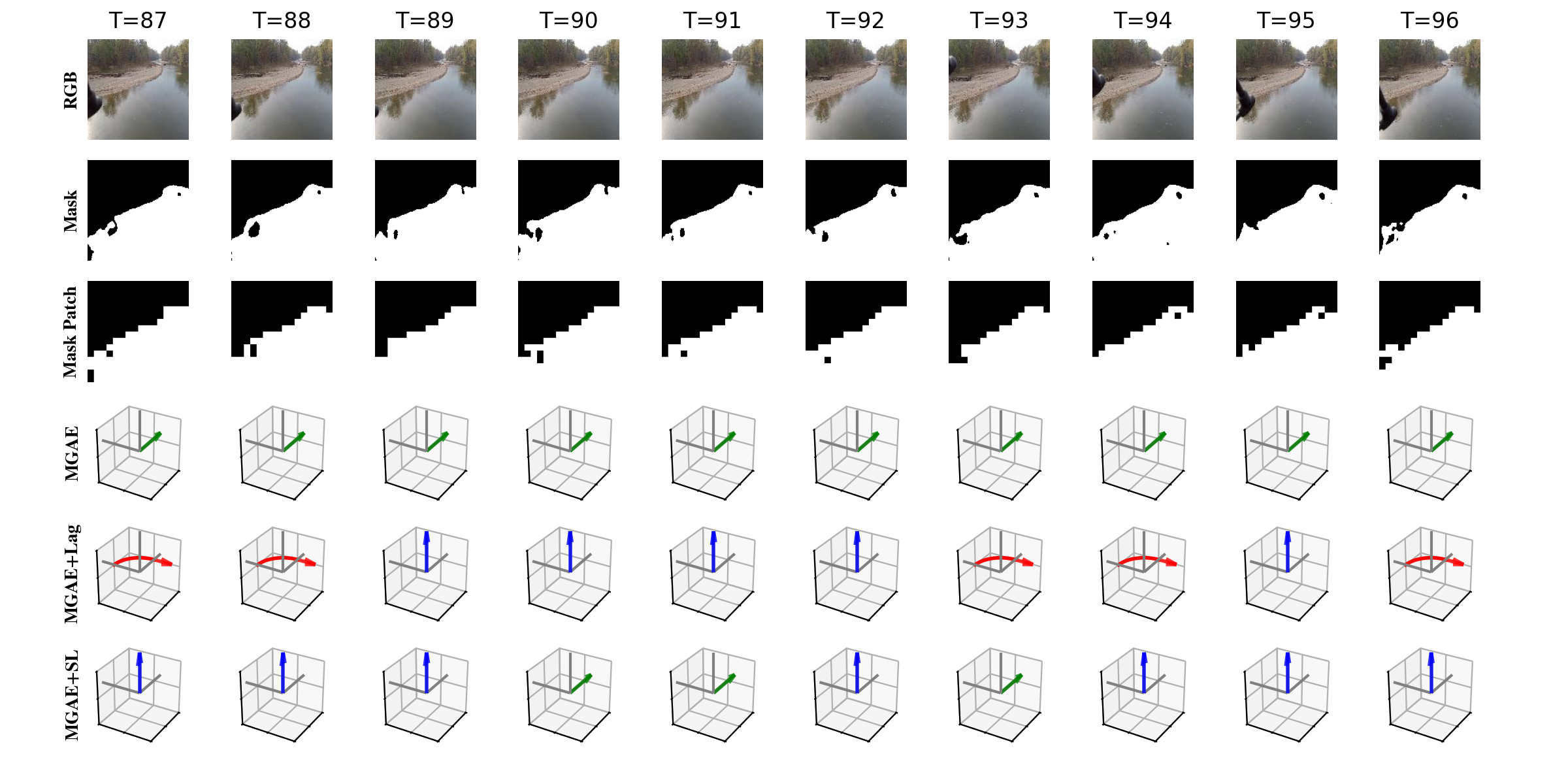}
    \caption{\revisenew{Open-loop policy evaluation of MGAE, MGAE+Lagrangian, and MGAE+SafetyLayer on \textbf{Wildcat Creek}. T is the frame index in the video of 2 frames per second.}}
    \label{fig:model-eval-wildcat}
\end{figure}

\revisenew{
We did the open-loop policy evaluation on the real-world riverine aerial images, which are the video frames of Wabash River and Wildcat Creek in Indiana, USA. 
The RGB images are passed through the trained semantic segmentation network (Unet \cite{ronneberger2015u} + Resnet34 \cite{he2016deep}, implementation adapted from \cite{Yakubovskiy:2019} and illustrated in \cite{wang2023aerial}) to obtain the binary water mask, then patchified to form the observation input to the navigation policies, Figure \ref{fig:patchified}.
Since the drone-view video was manually collected by safely flying the UAV above the creek, the estimated costs from the cost estimator of all policies exhibit very small values, thus we mainly compare the MGAE, MGAE+Lagrangian and MGAE+Safety Layer, without post-training safety layer enabled.
All trained policies are from the last saved checkpoints (Figure \ref{fig:cade-sre-training-comp}), and use deterministic actions for evaluation on the real-world video.
Besides, the latent vector of GRU is reset at each policy inference step since sequential observations in the video are not caused by the policy (open-loop), thus it is valid to test the non-recursive "first response" of all policies.
The sequence of the actions selected by these three policies, as well as the observations (RGB, Mask, Patchified Mask), is shown in Figure \ref{fig:model-eval-wabash} and \ref{fig:model-eval-wildcat}.
}

\revisenew{
By observing the actions of different policies on the Wabash River video frames (Figure \ref{fig:model-eval-wabash}), we found that policies trained with either Lagrangian method or cost-planning safety layer tend to be more conservative. 
From $T=37$ to $T=40$, we can see MGAE+Lagrangian method prefers horizontal rotation to make UAV heading align with the river, instead of moving forward as MGAE policy does.
For both MGAE+Lagrangian and MGAE+SafetyLayer, they prefer increasing the altitude of the UAV ($T=[34, 36]$, $T=[41, 43]$) due to the large proportion of water patches in the observation, whereas MGAE chooses to move forward at these steps.
This reflects that both methods considering the safety cost during training learn the trade-off between navigation and potential hazards, because a higher percentage of water patch usually means lower flying altitude and less perception of the whole water region or river banks, which are the main clue of vision-driven river follow.
}

\revisenew{
Similar phenomenon can be observed from the policy evaluation on the Wildcat Creek (Figure \ref{fig:model-eval-wildcat}).
The priority of MGAE+Lagrangian policy is to adjust the heading angle and height of the UAV before making the longitudinal navigation action, whereas MGAE policy basically thinks that the current observation is still suitable to proceed forward.
MGAE+SafetyLayer policy lies in between the aggressive MGAE policy and the conservative MGAE+Lagrangian policy, with intermittent forward moving actions ($T=\{90, 91, 93\}$). 
The evaluation videos for all frames can be found on our Github page.
Overall, the open-loop policy evaluation validates the efficacy of the proposed MGAE method for learning navigation policy in partially observable SMDPs, while the two safety-injection learning methods (MGAE+Lagrangian and MGAE+SafetyLayer) show their capabilities in constraining potentially unsafe actions. 
The results are coherent with the numerical comparison in Table \ref{tab:sre_eval_cade}, where MGAE method is prone to achieve higher episodic reward due to its aggressiveness, while the MGAE+SafetyLayer method achieves the best trade-off between obtaining marginal gains while reducing safety violations. 
The semantic segmentation on real-world video frames brings another source of uncertainty for inference of vision-driven policies, but is out of the scope of this paper.
}

\section{DISCUSSION}\label{sec:discussion} 
% Nuances of implementation details, implications and limitations and suggestions.
In developing the CADE framework, several design choices were made to balance performance, safety, and practicality. This section discusses these decisions, their implications based on experimental results, and potential directions for future research.

% Magnitudes of rewards (reward scaling) matter for MGAE, especially when the actor and the reward estimator share the same recurrent network (multi-task learning).
\textit{Reward Scaling and Multi-Objective Learning}:  
The magnitude of rewards significantly influences the MGAE, especially when the actor and reward estimator share a recurrent network. Empirical evidence suggests that larger reward scales (for example, $\max(R) = 1$) enhance policy learning, potentially due to more substantial gradient signals during backpropagation. This observation aligns with findings that appropriate reward scaling can balance exploration and exploitation, leading to improved learning efficiency and performance. Conversely, smaller rewards may result in weaker gradients, hindering effective policy updates. Future work could explore adaptive reward scaling techniques to dynamically adjust reward magnitudes, optimizing learning stability and convergence in this multi-objective learning scenario.

% The reason behind reward-cost separation, reward is task-dependent but cost is environment-dependent. Separation benefits multi-task learning and safe exploration without any task on mind.
\textit{Separation of Reward and Cost Functions}: 
The separation of reward and cost functions in CSMDP allows for independent learning of task-specific objectives and environment-induced risks. 
Since the cost estimator and SDM operate solely based on the environment’s structure and immediate observations, they remain functional even when the task changes within the same environment. 
This design facilitates seamless task transitions without requiring extensive retraining, making it highly adaptable for multi-task learning scenarios. 
Additionally, this separation enables safe exploration in reinforcement learning settings where multiple objectives must be optimized sequentially or simultaneously. 
For instance, an agent trained for river following can be repurposed for search-and-rescue missions in the same riverine environment, leveraging the existing safety mechanisms. 
Furthermore, this approach enhances transferability across different reinforcement learning paradigms, including curriculum learning and lifelong learning, by ensuring that safety constraints persist while allowing flexible adaptation to new reward structures.

% Episode-by-episode training is in part due to the hidden unit in GRU needs sequential data input, but from hindsight, this approach suits more for online RL and real-time training, which is critical in real world deployment of RL policy.
\textit{Episode-by-Episode Updates}:
The decision to update networks on an episode-by-episode basis, rather than through random sampling, ensures that the GRU's hidden units accurately capture sequential dependencies. This approach is particularly advantageous for online reinforcement learning and real-time applications, where maintaining temporal coherence is critical.

% GRU, as a recurrent network that captures historical information for both decision-making and non-Markovian reward prediction, can be replaced by LSTM or Transformer structures. However, for simplicity of deployment and compactness of model size, GRU was chosen.
\textit{Choice of Recurrent Networks}: 
The GRU was selected for its ability to capture historical information pertinent to decision-making and non-Markovian reward prediction, while maintaining a relatively simple architecture. 
Alternatives like LSTMs \cite{hochreiter1997long} or Transformers \cite{waswani2017attention} offer potential benefits in modeling long-range dependencies. However, GRUs provide a favorable trade-off between performance and computational efficiency, making them suitable for deployment in resource-constrained environments. Future work could evaluate the performance of these alternative architectures within the CADE framework, particularly in tasks requiring complex temporal dependencies.

% Cost is designed to be Markovian in track/river following environments as the semantic observation at the current step contains enough information about the danger at this frame. However, in cases where cost is history-dependent, the cost estimator can also be appended to the recurrent network, like the reward estimator.
\textit{Design of Cost Function}: 
In track or river following tasks, the cost function is designed to be Markovian, relying solely on the current observation, which sufficiently encapsulates immediate environmental risks. 
This design simplifies the cost estimation process. 
In scenarios where costs depend on historical context, integrating the cost estimator with a recurrent network could enhance predictive accuracy. 
Future research could develop adaptive mechanisms that determine the necessity of historical information for cost estimation, adjusting the model architecture accordingly.
\revisenew{
If using Markovian cost, as we observed the performance discrepancies of safety regulation methods in CliffCircular (Table \ref{tab:cliffcircular_eval_cade}) and SRE (Table \ref{tab:sre_eval_cade}), correctly relating the visual observation of the agent to the corresponding cost is the key to improving the performance of safety regulation methods during learning.
}

% Observation space
\textit{Observation Space Considerations}
The design of the observation space leverages patchified semantic observations, represented as multi-binary maps, to achieve several critical benefits. 
First, by reducing the dimensionality compared to raw RGB images or continuous VAE encodings, the approach improves sample efficiency, lowers computational costs, and accelerates training convergence. 
Second, the semantic segmentation process abstracts away incidental variations such as lighting and textures, enhancing the robustness of the policy against environmental variability. 
Third, the modular separation of vision interpretation from policy learning facilitates transferability and enables independent fine-tuning for different tasks or environments. 
Fourth, the inherent interpretability of binary masks aids in debugging and provides transparency into the agent’s decision-making process. 
Finally, by focusing on invariant environmental features through pre-trained segmentation, the approach mitigates task bias often introduced by end-to-end learning, resulting in more robust and adaptable policies.

\revise{
\textit{Patchification Granularity:}
The choice of $16 \times 16$ patches from a $128 \times 128$ semantic mask in SDM balances spatial resolution, computational efficiency, and predictive stability. 
Finer patchification (e.g., $32 \times 32$) retains more local details but makes the model highly sensitive to minor misalignments and introduces higher computational overhead in planning. 
Coarser patchification (e.g., $8 \times 8$) risks losing meaningful spatial relationships, leading to homogenized inputs and degraded predictive accuracy. 
Additionally, in real-world riverine environments, semantic segmentation models introduce noise, particularly along water boundaries. 
Using a moderately coarse patch size helps smooth out segmentation noise, improving robustness to perception errors. 
The current choice is a trade-off between model validity and efficiency, but future work may explore the impact of different patch sizes on SDM’s generalization across diverse riverine environments and real-world deployment scenarios.}

% Continuous vs Discrete action space. This is parallel to the selection between waypoint control and velocity control of a vehicle. Waypoint control decouples the task from the vehicle dynamics and environmental disturbances, thus is applicable to different vehicles without fine-tuning of either policy or SDM.
\textit{Action Space Considerations}: 
The choice between continuous and discrete action spaces parallels the decision between waypoint control and velocity control. 
Waypoint control abstracts the task from specific vehicle dynamics and environmental disturbances, promoting transferability across different platforms without extensive re-tuning. 
In contrast, continuous action spaces, often used in velocity-based control, offer finer maneuverability but require vehicle-specific tuning.

% The incorporation of no operation action in the action space complicates the training, increase the chance of stagnation and conservativeness. The reason why it is kept is that for future research like drone-boat collaborative navigation task, two vehicles need to maintain a safe communication distance. In that situation, halting is a good action for the other vehicle to catch up. On the other hand, MGAE method shows its ability to learn a decent track/river following policy even with the presense of no operation action.
\textit{Inclusion of No-operation Action}:
Within the discrete action space, the inclusion of a no-operation action introduces further complexity to policy learning. 
While it increases the likelihood of stagnation and overly conservative behaviors, it plays a crucial role in multi-agent scenarios. 
For example, in a collaborative drone-boat navigation task, maintaining a safe communication distance between agents is essential. 
In such cases, the ability to remain stationary allows one agent to pause and wait for the other to adjust its trajectory, improving coordination. 
Despite the challenges, the MGAE method demonstrates robust policy learning even in the presence of the no-operation action, successfully guiding the agent toward effective track and river following behaviors.

% Baseline selection in MGAE can be in Monte-Carlo manner (past mean episodic return) or boot-strapping state-dependent manner (reward-to-go critic), as long as it provides a update-to-date reasonable estimate of what the agent expects to gain starting from current state onward.
\revise{\textit{Baseline Selection in MGAE}:
The MGAE framework allows for flexibility in baseline selection, utilizing either a Monte Carlo approach, based on past mean episodic returns, or a bootstrapped, state-dependent method, such as a reward-to-go critic. 
The critical requirement is that the baseline provides an up-to-date and reasonable estimate of expected returns from the current state onward. 
Future work could explore adaptive baseline strategies that dynamically select the most appropriate method based on the agent's learning phase and environmental dynamics.}

% SDM requires the sequential observations to have enough overlap for it to work, thus the step size (discrete) or the sampling interval (continuous) should not be too large.
\textit{SDM and Observation Overlap}:
The effectiveness of the SDM hinges on sufficient overlap in sequential observations; excessive step sizes or sampling intervals can degrade its performance. Ensuring appropriate temporal resolution in data collection is essential for accurate modeling of environmental dynamics. Future research could investigate adaptive sampling techniques that balance the need for observational overlap with computational efficiency, optimizing the performance of the SDM across various tasks, outdoor or indoor.

\revise{
\textit{SDM and Dynamic Objects}:
SDM relies on the assumption of a predominantly static environment, using planar homography to predict future water surface observations. 
However, dynamic obstacles such as boats, floating debris, or wildlife may occasionally appear on the water surface, potentially introducing noise into the learned homography transformation. 
Rather than modifying SDM to explicitly model dynamic objects—an approach that would require transitioning to a 3D environmental representation—handling such obstacles is more appropriately addressed at the semantic segmentation stage. 
A preprocessing step can be introduced to detect and mask dynamic obstacles before passing the segmented water surface to SDM. 
Motion detection techniques, optical flow estimation, or temporal consistency checks in segmentation outputs can help identify transient objects, ensuring that only the stable water surface boundary information is retained. 
This approach maintains SDM’s effectiveness in capturing the essential river structure while preventing dynamic obstacles from corrupting its learned geometric constraints.
}

% safety layer with reward-cost tradeoff planning
\textit{Safety Layer with Reward-Cost Tradeoff Planning}:
Compared to safety layer with cost planning, safety layer with reward-cost tradeoff planning could prevent excessive focus on safety from hindering task completion. 
Instead of purely ranking costs among sampled trajectories, integrating planned reward-cost evaluations would enable more balanced decision-making, especially during deployment. 
This raises a key research direction: how to effectively balance rewards and costs to maintain safety without overly conservative behavior. 
%Developing adaptive trade-off mechanisms could enhance both performance and safety in real-world applications.

\revise{
\textit{Real-World Feasibility:}
While the proposed framework demonstrates strong performance in simulation, deploying it in real-world riverine environments presents several challenges.
Robustness to real-world water reflections and seasonal variations remains a key concern, as these factors could introduce noise into the semantic segmentation model, affecting both SDM and actor. 
Future work could incorporate domain adaptation techniques or uncertainty-aware segmentation models to enhance resilience against such variations. 
Additionally, the safety layer’s intervention mechanism could be extended beyond autonomous overrides—when triggered, the UAV could notify a human operator about the intended safe action and await confirmation before proceeding. 
This human-in-the-loop approach could ensure safer real-world deployment while retaining automation benefits. 
Furthermore, leveraging human feedback for online re-training of the policy, SDM, and cost estimator could help the agent adapt to evolving environmental conditions and mitigate discrepancies between simulation and reality.
}

\section{CONCLUSION}\label{sec:conclusions}
We present a vision-driven SafeRL framework for UAV-based river following, addressing the challenges of navigating dense riverine environments with unreliable GPS signals. 
We frame the task as a Constrained Submodular Markov Decision Process (CSMDP) and introduce three key contributions: Marginal Gain Advantage Estimation (MGAE) for improved reward advantage estimation by an immediate reward estimator in submodular reward settings, Semantic Dynamics Model (SDM) for interpretable and data-efficient future first-person-view observation prediction, and the Constrained Actor Dynamics Estimator (CADE) architecture, which integrates SDM and an immediate cost estimator into safety-aware model-based RL for solving partially observable CSMDPs.

Simulations demonstrate that MGAE improves training efficiency and policy performance, outperforming traditional critic-based advantage estimation methods. 
SDM enhances short-term state prediction, allowing for more accurate cost estimation, which is critical for effective safety regulation. The CADE framework balances task performance and safety, with the Lagrangian-based approach proving effective in structured environments where cost signals are well-defined, while the safety layer plays a crucial role in inference by mitigating risk in complex settings with sparse cost signals.

Future work will explore enhancing cost estimation accuracy, particularly in environments where costs are terminal and weakly correlated with observations. 
Additionally, improving trajectory planning and sampling strategies could further refine safety regulation, ensuring robust decision-making across diverse operational scenarios. 
Finally, integrating real-world flight tests will validate the practical applicability of CADE for autonomous UAV navigation in real riverine environments.

% Future work
%% Non-Markovian cost
%For Constrained Submodular Markov Decision Processes whose cost function is also non-Markovian, different strategies need to tackle the cost critic to better approximate the cost value, e.g., using an autoregressive function approximator for the cost function.
%% Action prior for SDM

\section*{Declaration of generative AI and AI-assisted technologies in the writing process
}
During the preparation of this work the author(s) used ChatGPT in order to improve the language and readability. After using this tool/service, the author(s) reviewed and edited the content as needed and take(s) full responsibility for the content of the publication.

\bibliographystyle{elsarticle-num} 
\bibliography{reference}




% %% For numbered reference style

%   \textit{\LaTeX: a document preparation system},
%   Addison Wesley, Massachusetts,

% \end{thebibliography}

\end{document}